\documentclass[11pt,a4paper]{memoir}

\setlrmarginsandblock{3.5cm}{*}{*}
\setulmarginsandblock{3.5cm}{*}{*}
\checkandfixthelayout 

\usepackage[OT1]{fontenc}

\usepackage[english]{babel}

\usepackage[utf8]{inputenc}

\usepackage[sc]{mathpazo}

\usepackage{amsmath,amssymb,amsfonts,mathrsfs}

\usepackage[amsmath,thmmarks]{ntheorem}

\usepackage{graphicx}

\usepackage{soul}

\usepackage{pdfpages}

\usepackage{varioref}

\usepackage{datetime}

\usepackage{mathtools}

\usepackage[h]{esvect}

\usepackage{array}

\usepackage{listings}
\usepackage[activate]{pdfcprot}

\usepackage{booktabs}
\usepackage{multirow, multicol}
\usepackage{colortbl}

\usepackage{tikz}

\usepackage{algorithm}
\usepackage{algorithmic}

\usepackage{pifont}

\usepackage{color, array}
\usepackage{todonotes}

\usepackage{caption}
\usepackage{subcaption}

\nonzeroparskip
\defaultlists

\makeatletter

\if@twoside
  \copypagestyle{chapter}{Ruled}
\else
  \copypagestyle{chapter}{ruled}
\fi
\makeoddhead{chapter}{}{}{}
\makeevenhead{chapter}{}{}{}
\makeheadrule{chapter}{\textwidth}{0pt}
\copypagestyle{abstract}{empty}

\makechapterstyle{bianchimod}{%
  \chapterstyle{default}
  \renewcommand*{\chapnamefont}{\normalfont\Large\sffamily}
  
  \renewcommand*{\printchaptername}{%
    \chapnamefont\centering\@chapapp}

  }

\chapterstyle{bianchimod}

\setsecheadstyle{\Large\bfseries\sffamily}
\setsubsecheadstyle{\large\bfseries\sffamily}
\setsubsubsecheadstyle{\bfseries\sffamily}
\setparaheadstyle{\normalsize\bfseries\sffamily}
\setsubparaheadstyle{\normalsize\itshape\sffamily}
\setsubparaindent{0pt}

\captionnamefont{\sffamily\bfseries\footnotesize}
\captiontitlefont{\sffamily\footnotesize}
\setsecnumdepth{subsection}
\settocdepth{subsection}

\pretitle{\vspace{0pt plus 0.7fill}\begin{center}\HUGE\sffamily\bfseries}
\posttitle{\end{center}\par}
\preauthor{\par\begin{center}\let\and\\\Large\sffamily}
\postauthor{\end{center}}
\predate{\par\begin{center}\Large\sffamily}
\postdate{\end{center}}

\def\@advisors{}
\newcommand{\advisors}[1]{\def\@advisors{#1}}
\def\@department{}
\newcommand{\department}[1]{\def\@department{#1}}
\def\@thesistype{}
\newcommand{\thesistype}[1]{\def\@thesistype{#1}}

\renewcommand{\maketitlehookb}{\vspace{1in}%
  \par\begin{center}\Large\sffamily\@thesistype\end{center}}

\renewcommand{\maketitlehookd}{%
  \vfill\par
  \begin{flushright}
    \sffamily
    \@advisors\par
    \@department, ETH Z\"urich
  \end{flushright}
}

\checkandfixthelayout

\makeatother

\theoremstyle{plain}
\numberwithin{equation}{chapter}

\theoremstyle{nonumberplain}
\theorembodyfont{\normalfont}
\theoremsymbol{\ensuremath{\square}}

\renewcommand{\epsilon}{\ensuremath\varepsilon}

\renewcommand{\phi}{\ensuremath{\varphi}}

\def\zapcolorreset{\let\reset@color\relax\ignorespaces}
\def\colorrows#1{\noalign{\aftergroup\zapcolorreset#1}\ignorespaces}

\usepackage[linkcolor=black,colorlinks=true,citecolor=black,filecolor=black,pdfencoding=auto,psdextra]{hyperref}

\usepackage{cleveref}

\usepackage{adjustbox}

\title{Learning Representations For Images With Hierarchical Labels}
\author{Ankit Dhall}
\advisors{Advisors: Anastasia Makarova, Dr. Octavian-Eugen Ganea, Dario Pavllo \\Prof.\ Dr.\ Andreas Krause}
\thesistype{Master's Thesis}
\department{Department of Computer Science}
\date{2019}

\begin{document}

\frontmatter

\begin{titlingpage}
  \calccentering{\unitlength}
  \begin{adjustwidth*}{\unitlength-24pt}{-\unitlength-24pt}
    \maketitle
  \end{adjustwidth*}
\end{titlingpage}

\begin{abstract}
    Image classification has been studied extensively but there has been limited work in the direction of using non-conventional, external guidance other than traditional image-label pairs to train such models. In this thesis we present a set of methods to leverage information about the semantic hierarchy induced by class labels. In the first part of the thesis, we inject label-hierarchy knowledge to an arbitrary classifier and empirically show that availability of such external semantic information in conjunction with the visual semantics from images boosts overall performance. Taking a step further in this direction, we model more explicitly the label-label and label-image interactions by using order-preserving embedding-based models, prevalent in natural language, and tailor them to the domain of computer vision to perform image classification. Although, contrasting in nature, both the CNN-classifiers injected with hierarchical information, and the embedding-based models outperform a hierarchy-agnostic model on the newly presented, real-world ETH Entomological Collection image dataset \cite{dhall_20.500.11850/365379}.
\end{abstract}
\clearpage
\renewcommand{\abstractname}{Acknowledgements}
\begin{abstract}
    I would like to thank Prof. Dr. Andreas Krause and Anastasia Makarova for believing in me and granting me the opportunity to work on this thesis in collaboration with the Institute of Machine Learning at ETH Zurich. I am grateful to Dr. Octavian-Eugen Ganea and Dario Pavllo for coming on-board the project and sharing their ideas and insight. It was great collaborating and brainstorming with all my supervisors who made this an extremely enriching experience.
    
    I would extend my gratitude to Dr. Michael Greeff from the ETH Entomological Collection for allowing access to their collection and Maximiliane Okonnek from the ETH Library Lab for ensuring that this project has a meaningful and significant impact for the scientific community long after its completion.
    
    I would like to thank my friends for their support. I am grateful to my family for their understanding, support and the unconditional freedom to pursue my goals.
\end{abstract}

\cleartorecto
\tableofcontents
\mainmatter

\chapter{Introduction}
\label{ch:intro}
\section{Motivation}
\label{sec:motivation}

In machine learning, the task of classification is traditionally performed using softmax and one compares class scores and returns the highest scoring label as the prediction. Such an approach safely assumes that categories might not be correlated among each other. Contrary to this assumption, in many commonly used datasets, labels are correlated and can be agglomerated to create more abstract concepts which are made up of a collection of relatively specific concepts. For instance \texttt{jeans}, \texttt{t-shirt}, \texttt{rain-jacket} and \texttt{ball-gown} are all \texttt{dresses}. Only a handful of previous works have used hierarchical information in the context of computer vision. Among them, in \cite{RedmonYOLO9000} the label-hierarchy from WordNet \cite{miller1995wordnet} is used to consolidate data across various datasets. On another occasion, \cite{deng2012hedging} show how to optimize the trade-off between accuracy and fine-grained-ness of the predicted class, but their proposed method only considers the label-hierarchy (=semantic similarity) and therefore disregards the visual similarity when performing this optimization.

Even though a classifier might not be able to distinguish between two breeds of dogs, it can still predict a more abstract yet correct label, \texttt{dog}. Predicting labels at different levels of abstractions can help catch errors when predicting more fine-grained labels and hence provide more meaningful predictions. Labels with varying levels of abstraction may also be beneficial for further downstream tasks that involve both natural language and computer vision such as image captioning, scene graph generation and visual-question answering (VQA). This work tries to exploit semantic information available in the form of hierarchical labels. We show that visual models when provided such guidance outperform a hierarchy-agnostic model. We also show how these models can be made more interpretable by using more explicity representation models such as embeddings for the task of image classification.

\subsection{Leveraging label-label interactions} Image classification models are usually designed as flat N-way classifiers. Originally, these models relied on hand-crafted features but nowadays use learnable convolutional filters to extract image features. These convolutional layers are tuned during the training procedure to maximize classification performance. Initial convolution layers contain simpler, more generic feature extractors for edges and blobs and as one moves through the cascade of filters, these meld together to extract more complex visual features such as textures and patters and eventually parts of objects and finally whole objects themselves. Such models perform classification solely on the basis of visual signals. These models only capture the label-image interactions and do not use additional information available about the inter-label interaction that could boost performance and additionally make the model more understandable.

\subsection{Long-tailed data distributions} Data imbalance is a common sight in the real-world machine learning setting. It is often the case that only a handful of images are available for some of the classes. A plausible explanation could be when the object of interest occurs infrequently in the domain from which the data is collected. In life science it could be a rarely occurring anomaly while in image-based datasets it could be an object that is seen less often than others.

If one were to arrange the labels in the form of a hierarchy or a directed acyclic graph (DAG), classes that represent more abstract data would usually occupy the upper levels while more specific classes would be their descendants, forming the lower levels of the hierarchy. The data distribution is such that there are fewer classes in the upper levels but, on average, have a larger number of data points for a given label. The distribution gradually change trends as one traverses down to the lower levels in the label hierarchy. At the bottom level, the complete opposite holds, the levels have a large number of labels but with least amount of data per label. This leads to the formation of a long tailed distribution with the classes in the upper level contributing to a large number of samples while classes in the lower level forming the long tail.

Such long tail distributions are not best-suited for machine learning models whose generalization capabilities rely largely on the availability of large amounts of data for each label. Leveraging auxiliary information could be of help in the presence of long tail distributions. Using this, coupled with visual features the model is able to relate classes across different levels and can exploit information from the data-rich upper levels in the hierarchy.

For instance, if a particular label lower in the hierarchy has only a handful of data, it can still share visual information about labels via its siblings (i.e. which share the same parent) if information about the hierarchy is injected into the model. Usually concepts that are siblings or belong to the same sub-tree of the hierarchy have commonalities among them this can be exploited by the model if information about label-label interactions is used.

\subsection{Visual similarity does not imply semantic similarity} 
\begin{figure*}[!htbp]
\centering
\begin{subfigure}[b]{0.2\textwidth}
    \centering
    \includegraphics[width=\textwidth]{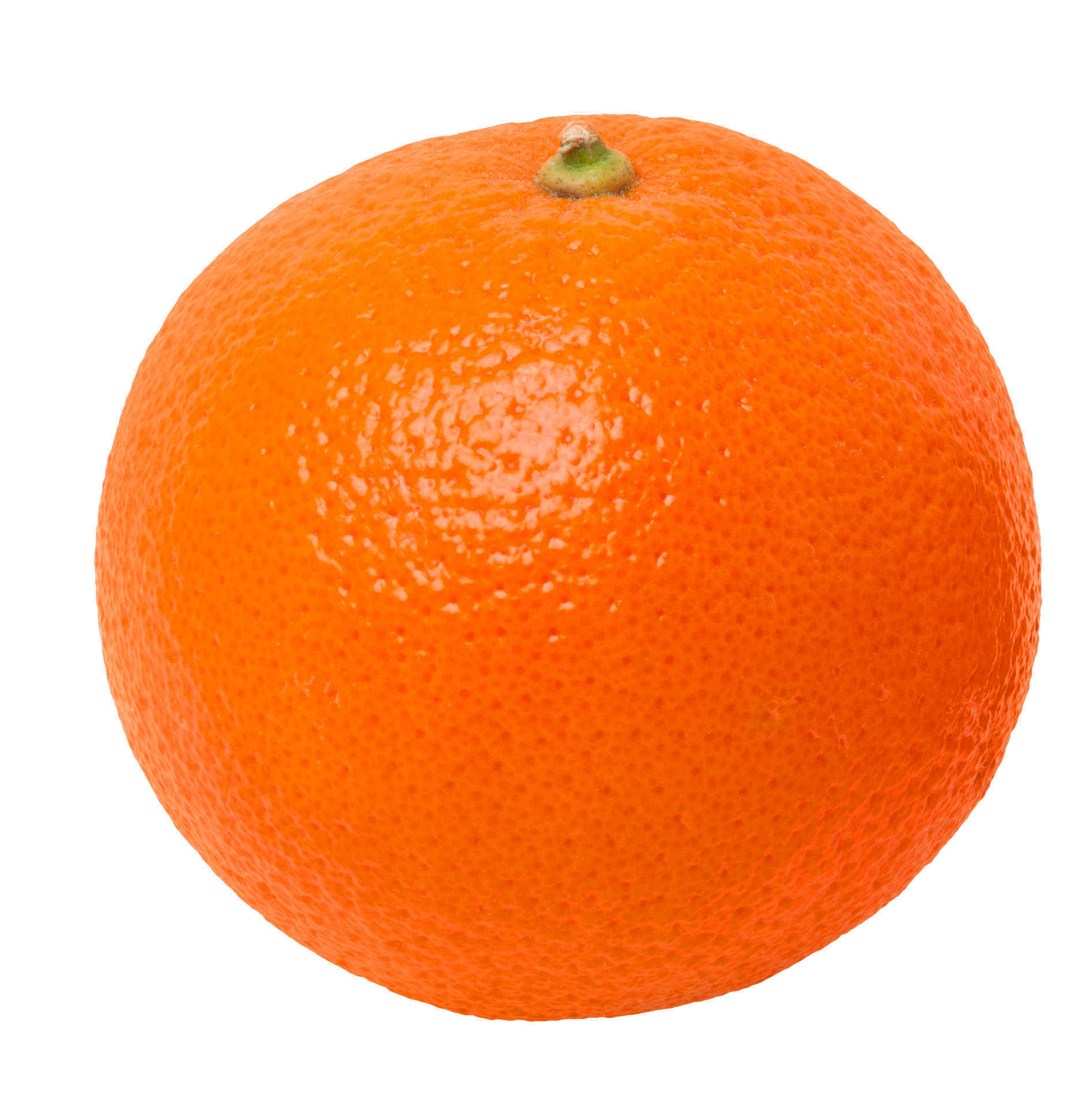}
    \caption[Network2]%
    {{\small \texttt{orange}}}    
    \label{fig:orange}
\end{subfigure}
\hfill
\begin{subfigure}[b]{0.25\textwidth}  
    \centering 
    \includegraphics[width=\textwidth]{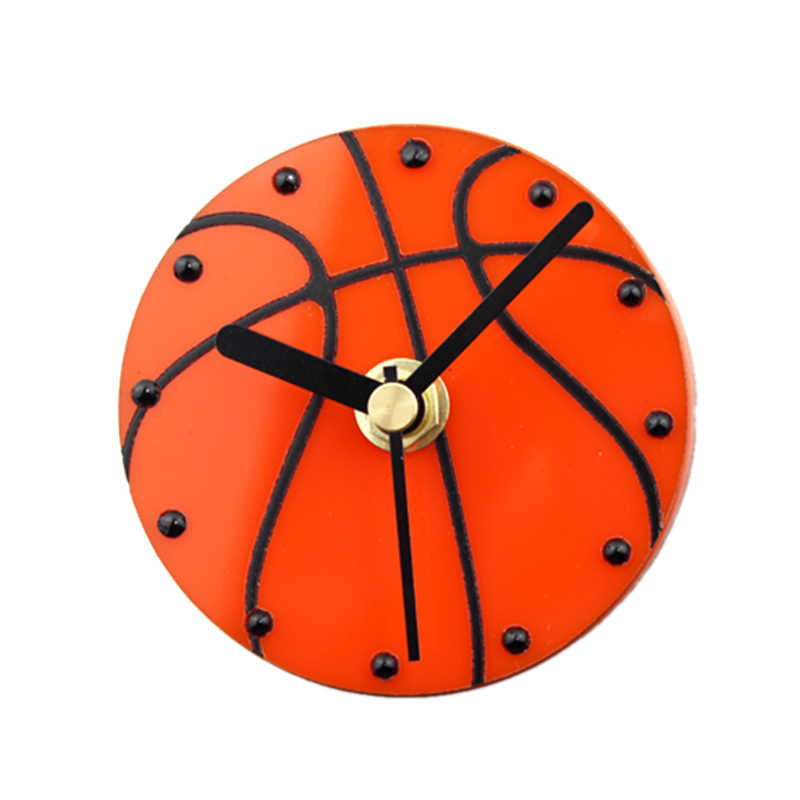}
    \caption[]%
    {{\small \texttt{clock}}}    
    \label{fig:basketvall-clock}
\end{subfigure}
\hfill
\begin{subfigure}[b]{0.2\textwidth}   
    \centering 
    \includegraphics[width=\textwidth]{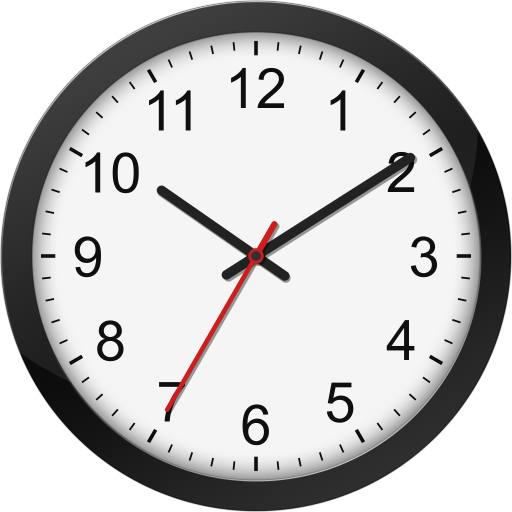}
    \caption[]%
    {{\small \texttt{clock}}}    
    \label{fig:analog-clock}
\end{subfigure}
\hfill
\begin{subfigure}[b]{0.2\textwidth}   
    \centering 
    \includegraphics[width=\textwidth]{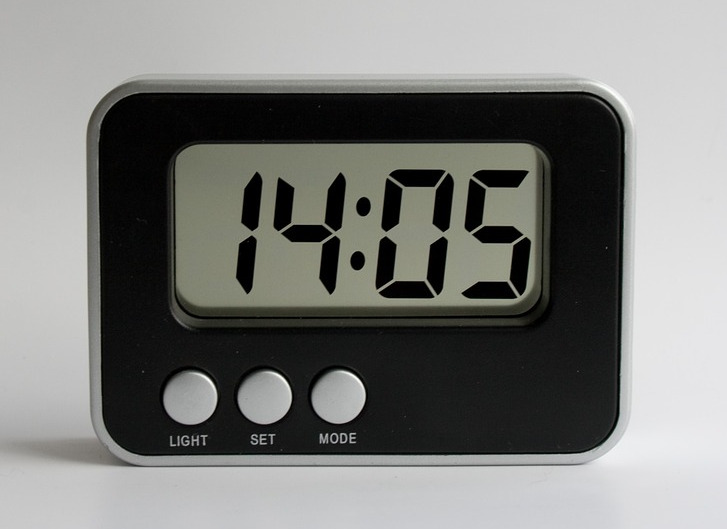}
    \caption[]%
    {{\small \texttt{clock}}}    
    \label{fig:digital-clock}
\end{subfigure}
\caption[]
{Although an \texttt{orange} and a basketball-themed clock have visual similarity, they are semantically unrelated. On the other hand, the digital, analog and basketball-themed clock are all visually distinct from each other but semantically similar as all of them are instances of \texttt{clock}. By introducing auxiliary information in the form of the label hierarchy such confusion could be avoided by models that only pay attention to visual features. {\footnotesize Image credits: Wikimedia, Lucky retail, Amazon, Pixabay}} 
\label{fig:semantic-vs-visual-sim}
\end{figure*}

Visual models rely on image based features to distinguish between different objects. But more often than not semantically related classes might exhibit marked visual dissimilarity. Sometimes it might even be the case that the intra-class variance of visual features for a single label is larger than the inter-class variance. In such scenarios learned representations for two instance with different visual appearance would be coerced away from each other, indirectly affecting the image understanding capability of the model.
In \cref{fig:semantic-vs-visual-sim} one can notice how semantic similarity and visual similarity are different concepts but are both essential to achieve better visual understanding.

\subsection{Uncovering the black-box model} If a human is tasked with classifying an image, the natural way to proceed is to identify the membership of the image to abstract concepts or labels and then move to increasingly detailed labels that provide more fine-grained understanding of the object in question. Even if an untrained eye cannot tell apart an \texttt{Alaskan Malamute} from a \texttt{Siberian Husky}, it is more likely to at least get the concept of \texttt{mammal} and its sub-concept \texttt{dog} correct.

\begin{figure}[!htbp]
\centering
\includegraphics[width=0.6\textwidth]{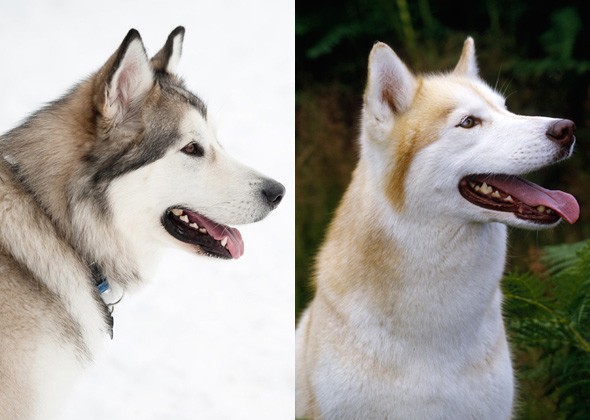}
\caption{The stark resemblance between an \texttt{Alaskan Malamute} and a \texttt{Siberian Husky} would make the life of an image classification model tough as it relies solely on visual features. {\footnotesize Image credits: Karin Newstrom, Animal Photography; Sally Anne Thompson, Animal Photography}}
\label{fig:malmute-vs-husky}
\end{figure}

Similarly, using the label hierarchy to guide the classification models we are able to bridge the gap in the way machines and humans deal with visual understanding. Incorporating such auxiliary information positively affects the explainability and interpretability of image understanding models.

\section{Predicting Taxonomy for Scientific Collections}

One of the main goals of this work is to assist natural collections, museums and other similar organizations that maintain a large library of biodiversity including both the flora and fauna. A lot of amateur collectors maintain their personal collections of insects and butterflies over their lifetime. Eventually most of these are donated and end up at collections and museums. With more than 2,000,000 specimens, the ETH Zürich Entomological Collection is one of the largest insect collections in Central Europe. 

The collection needs to sort these specimens according to their taxonomies. The process involves hiring of an external specialist who specializes in particular families of these organisms. The process of sorting these is not only expensive but is also constrained by the number of available specialists. If this resource intensive task could be preceded by a pre-sorting procedure where these specimens are categorized based on their \texttt{family}, \texttt{sub-family}, \texttt{genus} and \texttt{species} in that order, it would make the complete process more economical.



With the help of data and machine learning, such a repetitive can be facilitated by non-specialists, largely cutting the costs. For example, in Switzerland, from 120 CHF per hour to 28 CHF per hour.

Annually, 40,000 specimens are donated to the ETHEC by the public. If this technology is accessible to the general public, the collection will already receive pre-sorted specimen, making their task simpler. A 100 million euros initiative beginning in 2019 will develop standards to integrate digitization across European institutions (DiSSCo \cite{addink2018dissco}). In Switzerland a similar initiative is underway (SwissCollNet \cite{frick2019swisscollnet}).

In this work, we particularly focus on the entomology of insects, more specifically the butterflies. A digitized version from the ETH's collection was used to create a dataset \cite{dhall_20.500.11850/365379} to perform empirical analysis using the methods investigated in this work.

\subsection{ETHEC dataset: a new entomological image dataset with label-hierarchy} We present a new dataset with images and the corresponding inter-label relationships in addition to the generally provided image-label relationships. The challenging dataset provides a good foundation to build upon for the rest of the work by using it to evaluate experiments.

\begin{figure}[!htbp]
\centering
\includegraphics[width=1.0\textwidth]{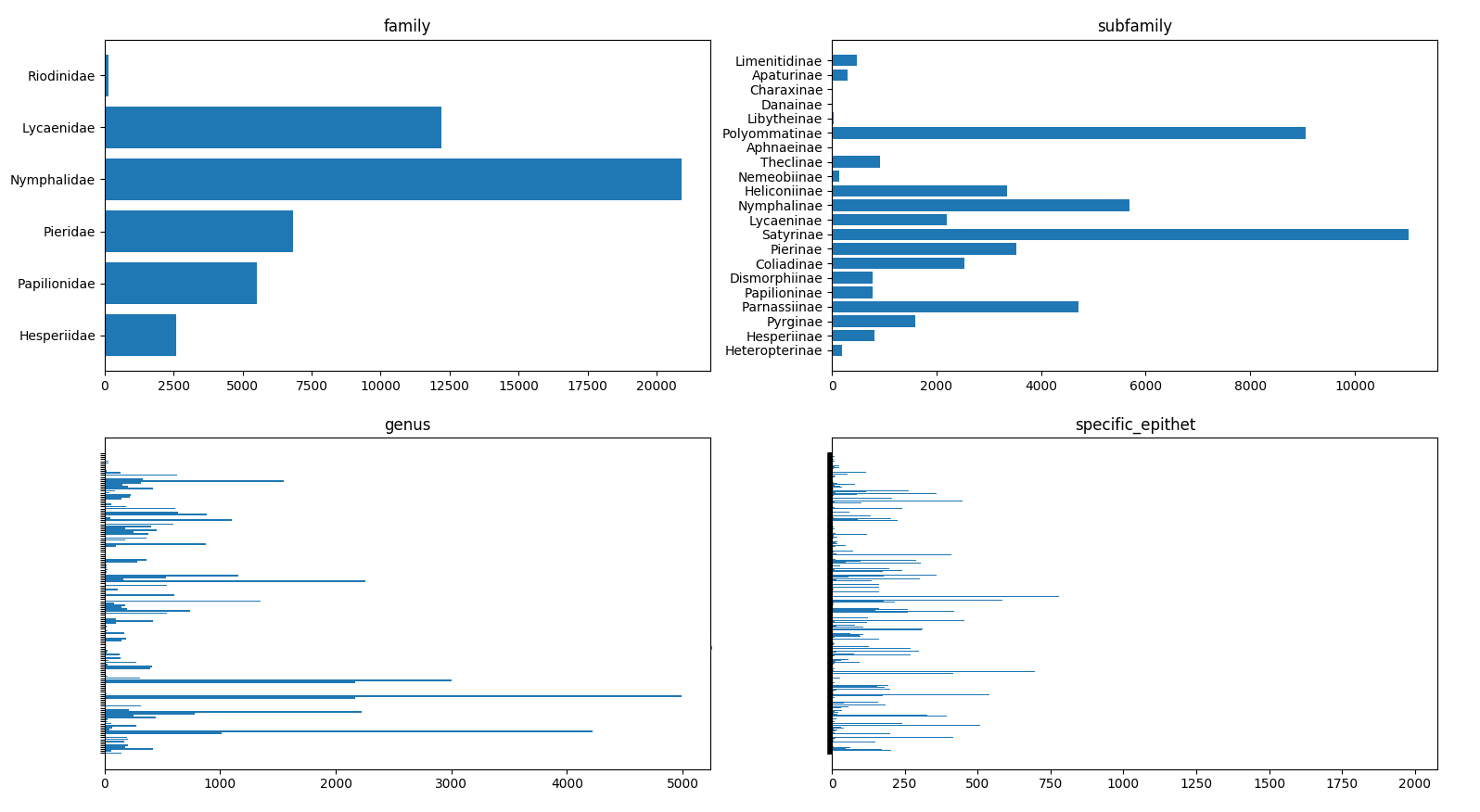}
\caption{The diagram shows the image distribution across each labels from the 4 levels of the hierarchy: 6 \texttt{family}, 21 \texttt{sub-family}, 135 \texttt{genus} and 550 \texttt{species}. The x-axis represents the number of images for a particular label and the ticks on the y-axis represent each label. For clarity, we have omitted the labels for genus and species.}
\label{fig:ethec_distribution}
\end{figure}

The ETH Entomological Collection (ETHEC) dataset \cite{dhall_20.500.11850/365379} has been directly taken from the field and is representative of a real-world dataset with imbalance not only in terms of the images per sample but also there is a significant disparity between classes and their descendant sub-classes as some of these sub-trees are disproportionately sized in terms of nodes. In \cref{fig:ethec_distribution} we illustrate the data distribution for each label in the ETHEC hierarchy.

With these peculiarities the dataset is representative of real-world scenarios and is more realistic as compared to the optimistic CIFAR \cite{krizhevsky2010convolutional_cifar} or ImageNet \cite{deng2009imagenet} that have balanced classes. The proposed dataset provides a challenging addition to the multi-label image classification task.

The ETHEC dataset is the digitized form of a subset of the collection which contains 47,978 butterfly specimens with 723 labels spread across 4 levels of hierarchical labels: 6 \texttt{family}, 21 \texttt{sub-family}, 135 \texttt{genus} and 550 \texttt{species} (561 \texttt{genus + species} combinations). They are interesting to look at from a computer vision perspective as they are the most visual specimens in the collection and these cues can be used as features to make label prediction and distinguishing specimens.

\begin{figure}[!htbp]
\centering
\includegraphics[width=0.9\textwidth]{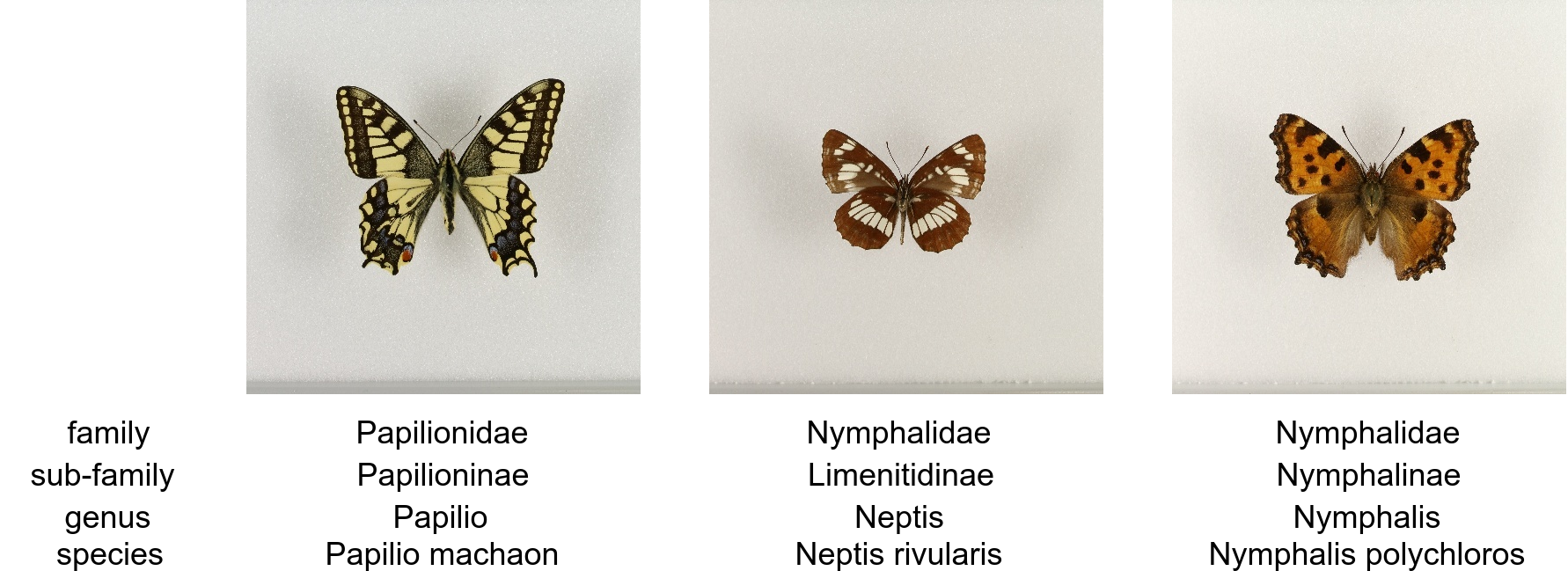}
\caption{Sample images and their 4-level labels from the ETHEC dataset. The dataset consists of 47,978 butterfly specimens with 723 labels spread across 4 levels of hierarchical labels: 6 \texttt{family}, 21 \texttt{sub-family}, 135 \texttt{genus} and 550 \texttt{species}.}
\label{fig:ethec_samples}
\end{figure}

The images are taken from the digitized collection at the ETH Entomological Collection. We pre-process them to remove any visual signals (barcodes, text labels, markings) that might leak label information about the specimen to a visual model. We also crop the images to lie at the center of the image and resize them to $448 \times 448$. We also provide metadata and labels for each of the 4 levels in the label-hierarchy. The dataset is split into \texttt{train}, \texttt{val} and \texttt{test} as 80-10-10. For labels with fewer than 10 images, we split the images equally between the three sets.

The dataset is been made publicly available, and can be found at the open-access link: \href{https://www.research-collection.ethz.ch/handle/20.500.11850/365379}{https://www.research-collection.ethz.ch/handle/20.500.11850/365379}.
\section{Contributions}
\label{sec:contributions}

\subsection{Injecting label-hierarchy information to improve CNN classifiers}

The work proposes, in addition to a hierarchy-agnostic baseline, 4 different methods of passing on knowledge about the label-hierarchy to a classifier to boost performance over a hierarchy agnostic classifier. Each method differs in the way they make this information available to the classifier and also the kind of information injected. So on top of the traditionally available image-label pairs during training, the methods provide additional label-label information as well.

The proposed methods are agnostic to the kind of features used or in general the feature extractor and can be easily extended to any classifier whose labels are arranged in a hierarchy. Since, the work tackles image classification, we use well-known visual feature extractor convolutional neural networks (CNNs) \cite{he2016resnet, krizhevsky2012imagenet, simonyan2014VGG} in our experiments. Although, there are works that propose modifications \cite{hu2019see} directly to the CNN architecture, we refrain from doing so such that these methods are model-agnostic can be used with any general classifier.
  
\subsection{Performing image classification by jointly embedding labels and images}

Order-preserving embeddings have shown great promise for capturing relations between concepts and tokens in the field of natural language processing \cite{ganea2018entailment_cones, vendrov2015order, suzuki2019hyperbolic_disk, le2019hearst_cones}. This work explores them in the context of computer vision to solve the task of image classification. We embed the labels (which arrange themselves as a hierarchy) and the images in a joint embedding space. Relations between labels and a given image can be used to predict labels and classify a given image.

In contrast to state-of-the-art approaches that use classical cross-entropy inspired classification loss function on CNN-based feature extraction backbones for images, we use embedding models to more explicitly represent label-label and label-image interactions. The idea is to allow the CNN to benefit from label-hierarchy information. In our experiments we show that a model trained with a classical cross-entropy inspired loss function performs worse than embedding-based classifiers that exploit the label-hierarchy.

Depending on the geometry that the parameters use and the space in which the embeddings live, these models can be categorized into Euclidean and non-Euclidean models.

\subsubsection{Euclidean models}
The field of natural language usually deals with modeling concepts as hierarchical structures and learning embeddings from unstructured text. Recent works \cite{vendrov2015order, ganea2018entailment_cones} model them as DAGs and suggest to embed them in order to preserve their asymmetric entailment relations. This information is usually lost if symmetric distance functions are used. Order-embeddings \cite{vendrov2015order} propose propose an asymmetric distance function that arranges the embedded concepts in an order-preserving manner. A more recent approach, entailment cones \cite{ganea2018entailment_cones}, use a more generalized version of the order-embeddings that are more space efficient and perform better. In contrast to the above approaches that have been proposed in the context of natural language we propose to jointly embed images and their labels and use their interactions to predict labels for unseen images.

\subsubsection{Non-Euclidean models}
Unlike the Euclidean models, non-euclidean models exploit non-zero curvature of their geometries. Hyperbolic geometry has negative curvature and can accommodate tree like structures (such as DAGs) with ease in comparison to Euclidean geometry. In hyperbolic space the volume of a ball grows exponentially with the radius \cite{nickel2017poincare} unlike the polynomial growth that we are aware of in Euclidean space. A set of works have \cite{nickel2017poincare, ganea2018entailment_cones, suzuki2019hyperbolic_disk} proposed to exploit spaces of negative curvature to better embed concepts and create state-of-the-art models to embed hierarchies. We use a model similar to the hyperbolic entailment cones \cite{ganea2018entailment_cones} where in addition to the labels we embed the images as well, treating the problem in a joint manner.

Generally embedding models and CNN-based classifiers are hard to compare because of the vastly different use-case and domain they are generally applied to. We use the embedding models as image classifiers and are able to make a fair performance comparison between different model categories. In addition to the image classification and joint embedding of labels and images, for the embedding based models, we also look at the quality of the embedding of the label-hierarchy itself. We report the performance on the ETH Entomological Collection (ETHEC) dataset \cite{dhall_20.500.11850/365379}.

\subsection{Contributions Summary}
\begin{itemize}
    \item We show how order-preserving embedding models, which are generally used for NLP tasks, can be extended for computer vision tasks such as image classification. We compare embedding-based classifiers with the label-hierarchy injected CNN-based classifiers. Both the Euclidean and non-Euclidean variants of embedding models are implemented and outperform the hierarchy-agnostic baseline. This shows promise for modeling and tackling downstream tasks that lie at the intersection of computer vision and natural language in a joint fashion.
    \item We compare a multi-label hierarchy-agnostic classifier as the baseline and 4 different methods detailed in the thesis to inject label-hierarchy knowledge into a classifier. Each of these methods takes into account hierarchical information at different levels of abstraction such as: the depth of hierarchy, edge connections and sub-tree relations.
    \item Although CNNs and embedding-based models are based on distinct paradigms we investigate the performance boost obtained by incorporating label-hierarchy information. Using the ETHEC dataset presented with this work, our experiments show that irrespective of the type of model being used, exploiting label-hierarchy leads to better image classification performance for both CNN-based and embedding-based classifier.
\end{itemize}

\section{Outline}
The remainder of this thesis is structured as follows:
\begin{itemize}
    \item In \cref{ch:intro} the motivation behind the methods and the need to exploit information from hierarchically organized labels is outlined.
    \item We skim over the relevant work in a similar direction as the one proposed in this manuscript in \cref{ch:background}. It provides mathematical background for methods that this work extends for joint label-image embedding for image classification. It also contains information regarding datasets and CNN-based feature extractors (CNN-backbones).
    \item In \cref{ch:cnn_models} we discuss in detail label-hierarchy injection into CNN-based models, probability distributions computation over the labels and finally how the predictions are made. We first discuss the baseline that disregards any external information than the image-label interactions. For the rest of the models, with each model, more information regarding the hierarchy is made available to the classifier. For clarity we separate out and compile the empirical analysis for the CNN-based models in \cref{ch:cnn_analysis}
    \item In \cref{ch:emb_models} we sketch the details for embeddings based models both Euclidean and non-Euclidean variants. The chapter also discusses label-embeddings before jointly embeddings labels together with images. We present the empirical results of the embeddings-based models in \cref{ch:emb_analysis}.
    \item Concluding remarks and possible directions for future work are discussed in \cref{ch:conclusion}.
\end{itemize}

\chapter{Problem Statement \& Background}
\label{ch:background}
\section{Related Work}
\subsection{Embedding based models for text and language}
An embedding is a mapping that maps discrete objects such as images, words or concepts to a relatively compact representation in the form of a vector living in low-dimensional embedding space.

For instance, words in a particular language can be represented using one-hot encoding in an V-dimensional space where V would be the vocabulary size for that particular language. However, this representation would contain little information or semantic meaning due to the inherent sparsity of the one-hot encoding. In addition to the embeddings lying in low-dimensional space, ideally, one would want these embeddings to arrange themselves in a manner such that the objects embedded close together represent high semantic similarity among themselves.

Traditionally, embeddings use a symmetric distance function to measure similarity between two objects. When one tries to embed concepts that have an asymmetric relation between them then using symmetric distance functions this detail is lost. One needs to use an asymmetric distance function to capture this relationship.

\paragraph{Order-embeddings.}

In \cite{vendrov2015order}, the authors tackle embedding of a semantic hierarchy as a partial ordering. Their work embeds a visual-semantic hierarchy that is anti-symmetric in nature. Instead of considering Euclidean or Manhattan distance, between two concepts as a measure of similarity the authors propose to use an asymmetric distance function when representing a hierarchy over images and text via embeddings. The work proposes a function that measures the presence of a parent-concept child-concept relation if the child-concept lives within a part of the subspace that is owned by the parent-concept. The distance metric is designed such that it defines a sub-space where it is valid for a child-concept to lie. This valid space is the positive orthant translated such that its origin is at the location (coordinates) of the embedding of the particular concept.

As opposed to the distance-preserving nature (which is generally the case), the order-preserving nature of order-embeddings ensures that anti-symmetric and transitive relations can be captured well without having to rely on physical closeness between points. Instead, the embeddings are learned by minimizing a loss that penalizes order violations. In \cite{vendrov2015order} the authors tackle two tasks: hypernymy prediction and image-caption retrieval. A hypernym is a pair of concepts where the first concept is more generic or abstract than the second. For instance, \texttt{(fruit, mango)} or \texttt{(emotion, happiness)}. The hypernymy prediction task has a natural hierarchy to the concepts, however, for the image-caption they create a two-level hierarchy where the captions form the more abstract, upper level while the images being more detailed form the lower level.

\paragraph{Euclidean cones.}
One major restriction of the representation and indirectly the distance function proposed in \cite{vendrov2015order} is that each concept occupies a large volume in the embeddings space (the coordinates of each embedding own a translated orthant irrespective of the number of descendants they have) and also suffers from heavy orthant intersections. This ill-effect is amplified especially in extremely low dimensions such as $\mathbb{R}^2$. To ameliorate such affects, the authors in \cite{ganea2018entailment_cones} propose a generalized version of order-embeddings called the entailment cones. These are more flexible and the region owned by a concept is not restricted to be a translated orthant but a convex cone. The cone that is owned by a concept originates at the location of the concept's embedding with its apex lying at these coordinates. Any concept that falls within the cone is considered as a sub-concept in context of hypernymy prediction.

\paragraph{Hyperbolic cones.}
In addition to the Euclidean cones, \cite{ganea2018entailment_cones} takes advantage of non-Euclidean geometry by learning embeddings in the hyperbolic space where the volume of a ball grows exponentially with the radius as compared to polynomially in Euclidean space. This property allows one to embed directed-acyclic graphs (DAGs); especially trees that grow exponentially with the height of the tree ($\text{height}=\text{log}_{\text{branchingFactor}}(N_{\text{nodes}})$), quite well even in very low-dimensional space \cite{ganea2018entailment_cones}.

The authors use a version of the Stochastic Gradient Descent (SGD) \cite{bottou2010sgd} that is for optimizing parameters on the Riemannian manifold, the Riemannian SGD to optimize embeddings in non-Euclidean manifolds. In their work, they propose the non-Euclidean entailment cones living in the hyperbolic space as well as their Euclidean variant. They focus on the task of hypernymy prediction on the WordNet hierarchy \cite{miller1995wordnet} by embedding a directed-acyclic graph using hyperbolic entailment cones and use it to classify whether a pair of concepts is a hypernym pair.

\paragraph{Hyperbolic Neural Networks.} In a more recent work \cite{ganea2018hyperbolicNN} the authors propose to have feed-forward neural networks to be parameterized in hyperbolic space. This allows downstream tasks to use hyperbolic embeddings for natural language processing (NLP) tasks in a more principled and natural fashion. They derive hyperbolic variants of logistic regression, feed-forward neural networks and recurrent neural networks. These are then used to take as input hyperbolic embeddings and are seen to perform at par or better than their Euclidean counterparts.

\paragraph{Disk embeddings.} \cite{suzuki2019hyperbolic_disk} proposes a generalization of order-embeddings \cite{vendrov2015order} and entailment cones \cite{ganea2018entailment_cones} for embedding DAGs with exponentially increasing nodes. The work focuses on the task of hypernymy prediction on the WordNet hierarchy \cite{miller1995wordnet} given a pair of concepts.

\paragraph{Other embedding methods.} The work proposed in \cite{barz2018hierarchy} maps images onto class embeddings where pairwise dot product is used as a measure of similarity. To embed the class labels they use a deterministic algorithm to compute class centroids by using hierarchical information from WordNet \cite{miller1995wordnet} to guide the embeddings semantically. They conjecture that semantics are complicated and are hard to learn only from visual cues. The class embeddings are pre-computed using the hierarchy. The image embeddings are mapped to the fixed class embeddings using a CNN with a combination of image classification and embedding loss. Their work focuses on the image retrieval task. A drawback of such an approach is that the label embeddings are fixed when training on the image embeddings. The labels might be embedded properly however they might not be arranged in a way that puts visually similar labels together. Fixing them when learning image embeddings prevents the combination of visual and semantic similarity to re-arrange the label embeddings in a manner that is better suited.

\cite{le2019hearst_cones} combines the idea of Hearst patterns and hyperbolic embeddings to infer \texttt{is-a} relationship from text such as \texttt{is-a}(car, object) or \texttt{is-a}(Paris, city). They propose to create a graph with the help of Hearst patterns and consequently embed it in low-dimensional hyperbolic space. They focus on different hypernymy tasks for text given a pair of concepts $(u, v)$: (1) if $u$ is a hypernym of $v$, (2) is $u$ more general than $v$, and (3) to what degree $u$ is a $v$.

\subsection{Embedding based models for images}
Visual-semantic embeddings, proposed in \cite{faghri2017vse++}, defines a similarity measure instead of a function that classifies a given pair as positive or negative. They calculate similarity scores and return the closest concept in the embedding space for a given query. They map features for language via an LSTM (Long short-term memory) and images via a CNN and map to the joint embeddings space through a linear mapping and measure similarity in this space using the inner product. They minimize a hinge-based triplet loss term and emphasize on hard-negatives by computing the loss for the closest negative (=hard-negative) instead of summing over all negatives. They focus on the task of cross-modal retrieval: caption retrieval given an images and image retrieval given a caption.

In one particular work, embeddings have been used for image classification \cite{tasho2018thesis}. The work uses order-embeddings to embed labels and images together for classification of the hierarchy. The work proposes to embed the labels first and then uses the transitivity of the embeddings across levels in the hierarchy to implicitly predict the upper levels after explicitly predicting the lower-most level. We extend this and use non-Euclidean models and also propose CNN-based models that exploit the hierarchy in varying degrees. 

In contrast to general CNNs for image classification, the work done in \cite{frome2013devise} extracts and exploits external information in the form of unannotated text in addition to the labeled images. They use a single unified models with embeddings and transfer knowledge from the text-domain to a model for visual recognition. They additionally perform zero-shot classification on classes extended on top of the ones in the ImageNet dataset \cite{deng2009imagenet}. The proposed work uses a combination of inner product to measure similarity and the hinge loss. With this approach they generalize well to unseen labels and are able to make relevant prediction even if the model classifies an image incorrectly (compared to the ground truth) for unseen classes from ImageNet 21K.

\subsection{Convolutional Neural Networks based models}
Kumar et al. \cite{kumar2017hierarchical} predict labels based on a tree formed from types of clothing. They create the hierarchy on the basis of detection errors, more specifically from the commonly confused classes by using a matrix of false positives. They use their methodology to classify clothing types by creating a 2-level hierarchy. To account for the hierarchy they predict conditional probability $P(\text{child}|\text{parent})$ as outputs from their classifier and multiply probabilities together to make predict labels for both the levels. They perform experiments with a 2-level hierarchy with a handful of labels in total and we observe the issues arise when the hierarchy is extensive and the data is scarce. A drawback of their method is the fact that the hierarchy is formed on the basis of confusion while predicting solely based on visual cues. This implies that the constructed hierarchy might not really have semantic similarity as the guiding principle but rather visual similarity. Our methods, on the other hand jointly incorporate both visual and semantic similarity to the model via injecting information about both image-label and label-label interactions.

In work done by Chen et al. \cite{chen2018finegrained} they propose to predict labels for different levels in a hierarchy. Their work is closest to ours in the sense that it tries to predict labels for each level in the hierarchy to which the images belong. They develop a sophisticated CNN architecture that uses a common feature extractor which then uses separate neural networks where each specializes to predict labels for each level. The fact that they use completely separate networks to predict labels for each level makes the model prone to over-fitting when the dataset is small and computationally intensive as well. They present a dataset with a 4-level hierarchy with images of butterflies across 200 species similar to the ETHEC dataset and construct hierarchies for existing Caltech UCSD Birds dataset \cite{WelinderEtal2010CUBdataset}. They compare performance for the final level in the hierarchy with many baseline methods but these methods only predict labels for the the most fine-grained label category (=the final level in the hierarchy) and not the others.

In tasks relating to fine-grained image classification, it is common to have class labels with only a handful of images. Instead of fine-tuning models pre-trained on all classes of large dataset like the ImageNet, the work done in \cite{cui2018large} proposes to select a subset of top-K labels to be used for pre-training based on domain similarity between the source and target domains. After pre-training on a subset of the large dataset that is visually similar, the transfer-learning then yields better performance than models which are fine-tuned after pre-training on the entirety of the large dataset. They propose a better method that is more efficient when performing pre-training. In a similar direction, \cite{srivastava2013discriminativeTL} use hierarchy information to perform transfer learning.

Hu et al.\cite{hu2019see} work on the task of classifying fine-grained visual classification where intra-class variance is large and inter-class variance is small due to the visual similarity of objects in images. They propose a CNN architecture which tries to learn discriminative regions in the image via attention maps. This is then used to refine the prediction of the model by looking at it closely with the help of the learned attention maps and draw attention to discriminative parts of the object. In addition to this, they also propose using unsupervised image data augmentation strategy guided by the attention maps by zooming, cropping and erasing parts of the image in order to generalize better. The proposed model uses attention maps to help focus on smaller details however, unlike our proposed CNN-models, it does not use any information about the hierarchy in which the labels are arranged. One could consider it as a complementary approach to the ones proposed in the thesis, based only on visual cues.

There have been a set of other works \cite{liu2018crowdattn, liu2018crowdattn, wang2017multiattn, chen2018recurrentattn} that use attention based mechanisms to focus on discriminative regions in an image.

\cite{liu2017localizing, chen2018knowledge} also explore the idea of exploiting external information by using part-based attributes to help models during the learning process.

\section{Background}
\subsection{Order-embeddings}
Typically a symmetric distance is used to ascertain semantic similarity between concepts in the embedding space. Order-embeddings \cite{vendrov2015order} propose to learn a mapping that cares about preserving the \textit{order} between objects than distance and introduce the problem of partial order completion. From a set of known ordered-pairs $\mathcal{P}$ and unordered-pairs $\mathcal{N}$ the goal is to determine if an arbitrary, unseen pair is ordered or not.

They propose to use a reversed product order on $\mathbb{R}^{N}$ due to its desirable properties. This is defined in \cref{eq:order_embeddings_rev_prod_order}.

\begin{equation}
\label{eq:order_embeddings_rev_prod_order}
y \preceq x \text{ if and only if } \bigwedge_{i=1}^{N} y_{i} \geq x_{i}
\end{equation}

The reversed order means that smaller coordinates represent a "higher" or more abstract position in the partial ordering.

Instead of having a hard-constraint they introduce an \textit{approximate} order-embedding to violate them as less as possible.

\begin{equation}
\label{eq:order_embeddings_E}
E(x, y) = || \text{max}(0, x-y) ||
\end{equation}

\begin{equation}
\label{eq:order_embeddings_loss}
\mathscr{L} = \sum_{(u, v) \in \mathcal{P}} E(f(u), f(v)) + \sum_{(u', v') \in \mathcal{N}} \text{max}(0, \alpha - E(f(u'), f(v')))
\end{equation}

where, $\mathcal{P}$ and $\mathcal{N}$ represent positive and negative edges respectively in the dataset $\mathcal{X}$. $\alpha \in \mathbb{R}_{+}$ is the margin. $f$ is a function that maps a concept to it's embedding. $E(f(u), f(v))$ is the energy that defines the severity of the order-violation for a given pair $(u, v)$ and is given by \cref{eq:order_embeddings_E}.

According to the energy $E(x, y) = 0 \iff y \preceq x$. For positive pairs where $y$ \texttt{is-a} $x$, one would like embeddings such that $E(x, y)=0$. $a$ \texttt{is-a} $b$ implies that $a$ is a sub-concept of $b$ or equivalently $b$ is more abstract than $a$ and that is its generalization.

\subsection{Euclidean Cones}
Euclidean cones \cite{ganea2018entailment_cones} are a generalization of order-embeddings \cite{vendrov2015order}. For each vector $x$ in $\mathbb{R}^N$, the aperture of the cone (with its apex located at this point) is based solely on the Euclidean norm of the vector, $||x||$, \cite{ganea2018entailment_cones} and is given by $\psi(x)$ in \cref{eq:euc_cones_psi}. One of the properties of these cones is that a cone can have a maximum aperture of $\pi/2$ \cite{ganea2018entailment_cones}. In addition to this, to ensure continuity and transitivity, the aperture should be a smooth, non-increasing function. To satisfy properties mentioned in \cite{ganea2018entailment_cones}, the domain of the aperture function has to be restricted to $(\epsilon, 1]$ for some $\epsilon$. $\epsilon=f(K)$ where K is a hyper-parameter.

\begin{equation}
\label{eq:euc_cones_psi}
\psi(x) = \text{arcsin} \bigg( \frac{K}{||x||} \bigg)
\end{equation}

$\Xi(x, y)$ computes the minimum angle between the axis of the cone at $x$ and the vector $y$. $E(x, y)$ measures the cone-violation which is the minimum angle required to rotate the axis of the cone at $x$ to bring $y$ into the cone.

\begin{equation}
\label{eq:euc_cones_xi}
\Xi(x, y) = \text{arccos}\bigg( \frac{||y||^2 - ||x||^2 - ||x-y||^2}{2 \; ||x|| \; ||x-y||} \bigg)
\end{equation}

\begin{equation}
\label{eq:euc_cones_E}
E(x, y) = \text{max}(0, \; \Xi(x, y) - \psi(x))
\end{equation}

\subsection{Hyperbolic Cones}
The Poincar\'e ball is defined by the manifold $\mathbb{D}^n = \{x \in \mathbb{R}^n: ||x|| < 1\}$. The distance between two points $x, y \in \mathbb{D}^n$ is given by \cite{ganea2018entailment_cones}:

\begin{equation}
\label{eq:hyp_dist}
d_{\mathbb{D}}(x, y) = \text{arccosh} \bigg(1 + 2 \frac{||x - y||^2}{(1-||x||^2)(1-||y||^2)} \bigg)
\end{equation}

and the Poincar\'e norm is defined as \cite{ganea2018entailment_cones}:

\begin{equation}
\label{eq:hyp_norm}
||x||_{\mathbb{D}} = d_{\mathbb{D}}(0, x) = 2 \; \text{arctanh}(||x||)
\end{equation}

Angles in hyperbolic space is the angle between the initial tangents of the geodesics. The angle between two tangent vectors $u, v \in T_{x}\mathbb{D}^{n}$ is given by $\text{cos}(\angle (u, v)) = \frac{\langle u, v \rangle}{||u|| \; ||v||}$ \cite{ganea2018entailment_cones}.

One needs to replace the cosine law and the exponential map to obtain the hyperbolic formulation for the cones in hyperbolic space \cite{ganea2018entailment_cones}. $|| . ||$ represents the Euclidean norm, $\langle . , . \rangle$ represents the Euclidean scalar-product and the unit vector $\hat{x} = x/||x||$.

The aperture of the cone is given by $\psi(x)$.

\begin{equation}
\label{eq:hyp_cones_psi}
\psi(x) = \text{arcsin} \bigg(K \frac{1-||x||^2}{||x||} \bigg)
\end{equation}

$\Xi(x, y)$ computes the minimum angle between the axis of the cone at $x$ and the vector $y$. $E(x, y)$ measures the cone-violation which is the minimum angle required to rotate the axis of the cone at $x$ to bring $y$ into the cone.

\begin{equation}
\label{eq:hyp_cones_xi}
\Xi(x, y) = \text{arccos}\bigg( \frac{\langle x, y \rangle (1 + ||x||^2) - ||x||^{2}(1 + ||y||^2)}{||x|| \; ||x-y|| \sqrt{1 + ||x||^2||y||^2 - 2 \langle x, y \rangle}} \bigg)
\end{equation}

\begin{equation}
\label{eq:hyp_cones_E}
E(x, y) = \text{max}(0, \; \Xi(x, y) - \psi(x))
\end{equation}

\subsection{Optimization in Hyperbolic Space}
In order to perform optimization, one cannot simply use the Euclidean gradients. For a given parameter $u$ one generally performs the following usual Euclidean gradient update:

\begin{equation}
\label{eq:euc_grad_update}
u \leftarrow u - \eta \; \nabla_{u} \mathscr{L}
\end{equation}

Instead, for parameters living in hyperbolic space, one should compute the Riemannian gradient and update it using the gradient direction in the tangent space and move $u$ along the corresponding geodesic in the hyperbolic space with the following update rule \cite{ganea2018entailment_cones} (Riemannian Stochastic Gradient Descent):

\begin{equation}
\label{eq:riemannian_grad_update}
u \leftarrow \text{exp}_{u} (\eta \; \nabla_{u}^{R} \mathscr{L})
\end{equation}

where, $\nabla_{u}^{R} \mathscr{L}$ is the Riemannian gradient for parameter $u$ and is computed by re scaling the Euclidean gradient by the inverse of the metric tensor given by:

\begin{equation}
\label{eq:riemannian_grad}
\nabla_{u}^{R} \mathscr{L} = (1/\lambda_{u})^2 \nabla_{u} \mathscr{L}
\end{equation}

$\lambda_{u}$ is different for each parameter and is computed as $\lambda_{u} = 2/(1 - ||u||^2)$ \cite{ganea2018entailment_cones}.

The exponential-map at a point $x$, $\text{exp}_{x}(v): T_{x}\mathbb{D}^{n} \rightarrow \mathbb{D}^{n}$, maps a point $v$ in the tangent space to the hyperbolic space and is defined as \cite{ganea2018entailment_cones}:

\begin{multline}
\label{eq:exp_x_v}
\text{exp}_{x}(v) =  x \; \frac{\lambda_{x}(\text{cosh}(\lambda_{x}||v||) + \langle x, \hat{v} \rangle \; \text{sinh}(\lambda_{x}||v||))}{1 + (\lambda_{x} - 1)\text{cosh}(\lambda_{x}||v||) + \lambda_{x} \langle x, \hat{v} \rangle \; \text{sinh}(\lambda_{x}||v||)} \\[5pt]
+ v \frac{(1 / ||v||) \; \text{sinh}(\lambda_{x}||v||)}{1 + (\lambda_{x} - 1)\text{cosh}(\lambda_{x}||v||) + \lambda_{x} \langle x, \hat{v} \rangle \; \text{sinh}(\lambda_{x}||v||)}
\end{multline}

\section{Datasets}
\label{sec:datasets}
\subsection{Hierarchical CIFAR-10}
We also perform experiments with the CIFAR-10 dataset \cite{krizhevsky2010convolutional_cifar}. There are 10 classes with 6000 32x32 images per class. In total, the dataset has 50000 images for training and 10000 for testing. To be consistent with our experiments, we use a $80\%-10\%-10\%$ split for training, validation and testing respectively. All fine-tuning is performed on the validation set, the test set is only used to report the model's performance.

The original dataset does not have a label hierarchy associated. Instead each image has a single ground truth label. Additional labels are added to introduce a 3-level hierarchy. Each image now is associated with 3 labels. The original labels are the leaves of this hierarchy. The root of the hierarchy is \textit{entity}, the first level of hierarchy splits into \textit{(living, non-living)}. \textit{living} entities are divided between \textit{(mammal, non-mammal)}. \textit{non-living} entities are divided between \textit{(vehicle, craft)}. The original classes are \textit{\{airplane, automobile, bird, cat, deer, dog, frog, horse, ship, truck\}}. \textit{mammal} is a parent of \textit{(cat, deer, dog, horse)} and \textit{non-mammal} of \textit{(bird, frog)}. \textit{vehicle} is a parent of \textit{(automobile, truck)} while \textit{craft} is a parent of \textit{(airplane, ship)}.

\subsection{Hierarchical Fashion MNIST}
Fashion MNIST \cite{xiao2017_FMNIST} is a dataset similar to MNIST where instead of hand-written digits it consists of 10 classes of clothing images. The dataset has 60,0000 training and 10,000 test samples. We split the training set such that 50,000 are used for training while the remaining 10,000 are used for validation. Each image is a 28x28 gray-scale image.

As in the case of CIFAR-10, here too, a 2-level hierarchy is introduced. The root of the hierarchy is \textit{fashion-wear}. The first level consists of \textit{top-wear}, \textit{bottom-wear and accessories} and \textit{footwear}. \textit{top-wear} has \textit{t-shirt}, \textit{pullover}, \textit{dress}, \textit{coat} and \textit{shirt} as the descendants. \textit{bottom-wear and accessories} has \textit{trousers} and \textit{bag} as descendants. \textit{footwear} has \textit{sandal}, \textit{sneaker}, \textit{ankle-boot} as descendants.

\subsection{ETH Entomological Collection}

\begin{figure}[!htbp]
    \centering
    \includegraphics[width=1.0\textwidth]{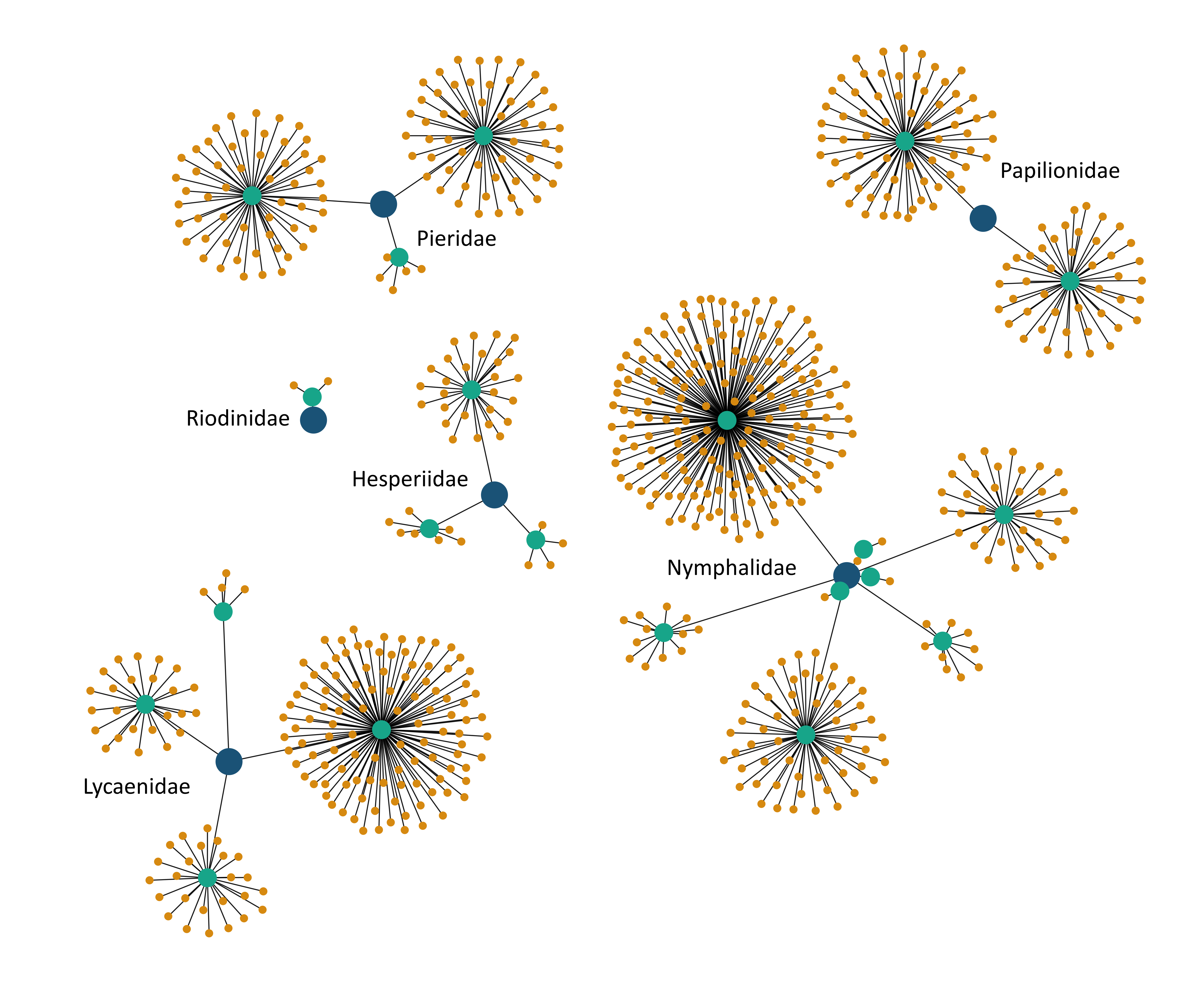}
    \caption{Hierarchy of labels from the ETHEC (Merged) dataset. It consists of labels arranged across 4 levels: family (blue), sub-family (aqua), genus (brown) and species. This visualisation depicts the first 3 levels. The name of the family is displayed next to its sub-tree.}
    \label{fig:d3_viz}
\end{figure}

\subsubsection{ETH Library's IMAGO project}
For experiments, we use data provided by ETH Entomological Collection abbreviated as ETHEC. The dataset, associated with ETH Library's IMAGO project, is an extensive collection of \textit{Lepidoptera} specimens that have been carefully curated and digitized with accurate metadata.

There exists metadata from 197052 specimen samples with all samples having labels spread across various hierarchical levels. For our experiments we make use of 4 such levels: 25 unique \textit{families}, 91 unique \textit{subfamilies}, 842 unique \textit{genera} and 2429 unique \textit{specific epithets} labels. The average branching factors are 25, 3.64, 9.25 and 2.88 for the respective levels. The label hierarchy has 3537 edges and 3387 nodes.

In the hierarchy, the maximum number of descendants belong to the \textit{family} \textit{Noctuidae} 19, to the \textit{subfamily} \textit{Noctuinae} 155 and to the \textit{genus} \textit{Eupithecia} 79. Maximum specimens belong to \textit{Geometridae} 48635 (\textit{family}), \textit{Noctuinae} 29555 (\textit{subfamily}), \textit{Zygaena} 17243 (\textit{genus}) and \textit{filipendulae} 2456 (\textit{specific epithet}).

Since this data is much larger as compared to other datasets discussed in this work, to better understand the data, we visualize the dataset using as an interactive graph using JavaScript. The visualization has basic functionality to view relations between nodes, the number of samples per label and the hierarchy level for a particular label. The nodes have size proportional to the order of magnitude of the number of samples for that label and are also color coded based on their hierarchy level. We visualize a subset of the complete hierarchy of the ETHEC dataset in \cref{fig:d3_viz}. More specifically, it is the subset that is used for our experiments.

\subsubsection{ETHEC Merged (ETHEC dataset)}
According to the way the nomenclature is defined, the \textit{specific epithet} (species) name associated with a specimen may not be unique. For instance, two samples with the following set of labels, (\textit{Pieridae}, \textit{Coliadinae}, \textit{Colias}, \textit{staudingeri}) and (\textit{Lycaenidae}, \textit{Polyommatinae}, \textit{Cupido}, \textit{staudingeri}) have the same \textit{specific epithet} but differ in all the other label levels - \textit{family}, \textit{subfamily} and \textit{genus}. However, the combination of the \textit{genus} and \textit{specific epithet} is unique. To ensure that the hierarchy is a tree structure and each node has a unique parent, we define a version of the database where there is a 4-level hierarchy - \textit{family} (6), \textit{subfamily} (21), \textit{genus} (135) and \textit{genus + specific epithet} (561) with a total of 723 labels. We call this version of the ETHEC dataset as ETHEC Merged dataset. We decide to keep the \textit{genus} level as according to experts in the field, information about genera helps distinguish among samples and result in a better performing model. For our experiments we use the merged version of the dataset to ensure that the hierarchy is a tree. The first 3 levels of the hierarchy are visualized in \cref{fig:d3_viz}.

\subsubsection{ETHEC Small dataset}
In order to allow for debugging and checking algorithms, we additionally use a smaller subset of the original ETHEC dataset, called the ETHEC Small dataset. \cref{table:ethecsmall_dataset} enumerates all labels across the 4 levels in the hierarchy for this the ETHEC Small dataset.

\begin{table}[htbp]
\centering
\begin{tabular}{| c || c |} 
 \hline
 Level & Label name  \\ [0.5ex] 
 \hline\hline
Family & Hesperiidae \\
       & Riodinidae \\
       & Lycaenidae \\
       & Papilionidae \\
       & Pieridae \\ \hline
Subfamily & Hesperiinae \\
          & Pyrginae \\
          & Nemeobiinae \\
          & Polyommatinae \\
          & Parnassiinae \\
          & Pierinae \\ \hline
Genus & Ochlodes \\
      & Hesperia \\
      & Pyrgus \\
      & Spialia \\
      & Hamearis \\
      & Polycaena \\
      & Agriades \\
      & Parnassius \\
      & Aporia \\ \hline
Genus + species & Ochlodes\_venata \\
                & Hesperia\_comma \\
                & Pyrgus\_alveus \\
                & Spialia\_sertorius \\
                & Hamearis\_lucina \\
                & Polycaena\_tamerlana \\
                & Agriades\_lehanus \\
                & Parnassius\_jacquemonti \\
                & Aporia\_crataegi \\
                & Aporia\_procris \\
                & Aporia\_potanini \\
                & Aporia\_nabellica \\
 [1ex] \hline
\end{tabular}
\caption{ETHEC Small dataset, a subset of the  ETHEC dataset.}
\label{table:ethecsmall_dataset}
\end{table}

\section{CNN-backbones}
We use convolutional neural networks to extract visual features from the images to perform classification. The CNN-based models are optimized using SGD \cite{bottou2010sgd} with a learning rate of 0.01 for 100 epochs and a batch-size of 64 unless specified otherwise.

\subsection{AlexNet}
AlexNet \cite{krizhevsky2012imagenet} proposed in 2012 shot to fame after exceptional performance on the ImageNet \cite{deng2009imagenet} challenge. It consists of 8 layers in total: the first 5 being convolutional layers and the remaining 3 being fully-connected layers. The original architecture outputs logits for 1000 class labels from the ImageNet challenge \cite{deng2009imagenet}.

\subsection{VGG}
VGGNet \cite{simonyan2014VGG} comprises of 16 convolutional layers and with 138 million parameters, is much larger than AlexNet \cite{krizhevsky2012imagenet}. They propose to use smaller filters (3 x 3) as opposed to larger filter size in previous CNNs. This reduces the effective number of parameters to achieve the same receptive field and in addition also incorporates more than one non-linearity.

\subsection{ResNet}
ResNet \cite{he2016resnet} showed that increasing depth improves network performance. They introduce skip-connections between groups of layers allowing the model to learn identity functions thus ensuring that the performance is as good as that of a shallower network. This facilitates better convergence rates than plain networks. The skip connections or shortcut connections do not increase the number of parameters in comparison to the original network (without skip connections). Even with the remarkable increase in depth ResNet-152 (152 layers) has fewer parameters than the VGG-16/19 \cite{simonyan2014VGG}. ResNets (pre-trained on the ImageNet dataset) are a popular choice as feature extractors for image related tasks.
\chapter{Methods: Injecting label-hierarchy into CNN classifiers}
\label{ch:cnn_models}
In this chapter we propose CNN-based models that use a convolutional layers to extract visual features and classify images. The architecture of the CNN in itself is not modified, rather the work focuses more on how different probability distribution formulations (across the labels) can be used to incrementally pass more information to the model about the label hierarchy. The chapter describes in detail 5 models where the first model is a baseline that is agnostic to any information from the label-hierarchy whatsoever. The remaining 4 models gradually make more information available to the model about such as the number of levels in the hierarchy and the edges between different labels.

\section{Hierarchy-agnostic classifier}

\begin{figure}[!htbp]
    \centering
    \includegraphics[width=0.9\textwidth]{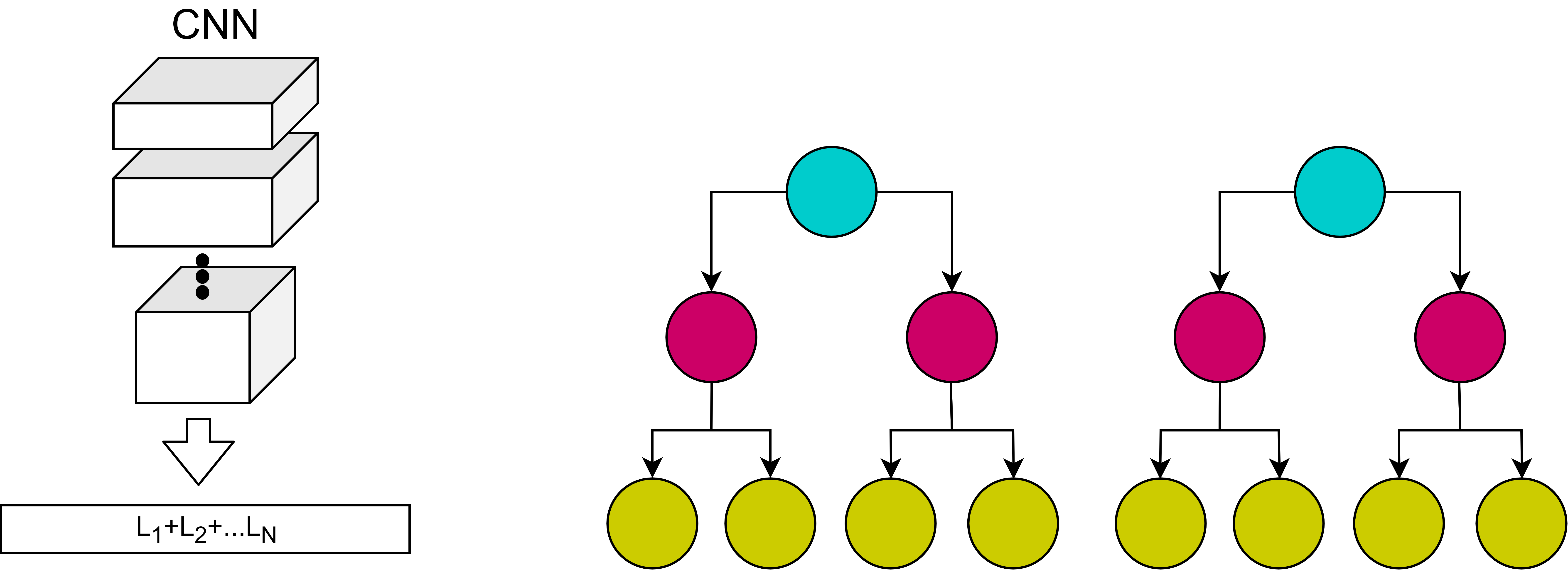}
    \caption{Model schematic for the hierarchy-agnostic classifier. The model is a multi-label classifier and does not utilize any information about the presence of an explicit hierarchy in the labels.}
    \label{fig:multi-label-model-schematic}
\end{figure}

As a baseline, we use a state-of-the-art convolutional neural network (CNN) image classification. For this, we use the residual network models proposed by \cite{he2016resnet}.

The baseline is agnostic to any information available in the form of the label hierarchy present in the dataset. In other words, it treats class labels from different levels in an unrelated manner with only the image being available for the model to predict a label for each level in the hierarchy. Labels across levels do not hold any special meaning and are treated indifferently.

The model performs $N_{total}$-way classification. $N_{total} = \sum_{i=1}^{L} N_{i}$ represents labels across all $L$ levels and $N_{i}$ are the number of distinct labels on the \textit{i-th} level. It uses the one-versus-rest strategy for each of the $N_{total}$ labels.

\begin{multline}
\label{eq:multi_label}
    \mathscr{L}(x, y) = - \frac{1}{N_{total}} * \sum_{j=1}^{N_{total}} y_{j} * \log\left(\frac{1}{(1 + \exp(-x_{j}))}\right) \\
                         + (1-y_{j}) * \log\left(\frac{\exp(-x_{j})}{(1 + \exp(-x_{j}))}\right)
\end{multline}

where, $x \in \mathbb{R}^{N_{total}}$, $y \in \left\{0, \; 1\right\}^{N_{total}}$ and $y^{T}y = L$.

$\mathcal{F}(\mathcal{I}) = x$, where $x$ are the logits (normalized and interpreted as a probability distribution) from the last layer of a model $\mathcal{F}$ which takes as input image $\mathcal{I}$.

\subsection{Per-class decision boundary (PCDB)}

Since, each image would be associated with more than one label we would apply a multi-label approach where multiple predictions for a single image are valid. For each class, the threshold is tuned based on the \textit{micro-F1} performance on the validation set \cite{yang2001study_scut}.

During evaluation time for the $L$ classes, label $L_j$ is assigned to an image if $x[j] > \theta_{j}, \forall j$. $\theta_{j}$ is tuned separately for label $j$ and is set to the value that maximizes F1-score performance on the validation set for that class.

\subsection{One-fits-all decision boundary (OFADB)}

In this variant, instead of having a different decision boundary for each class only a single decision boundary is used. To tune for a single threshold across classes, predicted scores and the ground truth labels across classes are used together to find the threshold that maximizes the \textit{micro-F1} score on the validation set.

During evaluation time for the $L$ classes, label $L_j$ is assigned to an image if $x[j] > \theta, \forall j$. Instead of having a per-class $\theta_{j}$ this version of the model uses a global threshold $\theta$. This is set to the value that maximizes F1-score performance on the entire validation set across all the $L$ classes.

\section{Per-level classifier}
For the next method more information from the label hierarchy would be made available to the network. Instead of a single $N_{total}$-way classifier as in the case of the hierarchy-agnostic model, we replace it with $L$ $N_{i}$-way classifiers where each of the $L$ classifiers handles all the $N_{i}$ labels present in level $L_{i}$. For this task we use a multi-label soft-margin loss.

\begin{figure}[!htbp]
    \centering
    \includegraphics[width=0.9\textwidth]{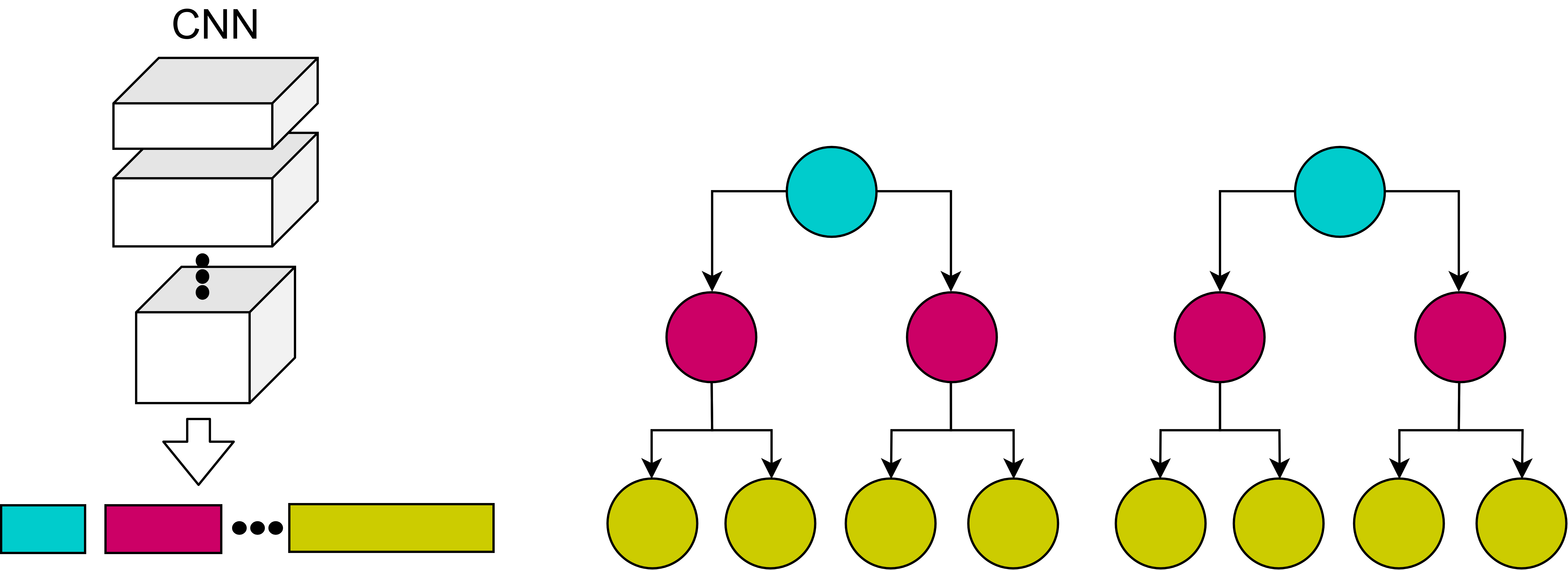}
    \caption{Model schematic for the per-level classifier (=$L$ $N_{i}$-way classifiers). The model use information about the label-hierarchy by explicitly predicting a single label per level for a given image.}
    \label{fig:multi-level-model-schematic}
\end{figure}

\begin{equation}
\label{eq:multi_level}
\mathscr{L}(x, \tau) = \sum_{i=1}^{L} \mathscr{L}_{i}(x_{i}, \tau_{i})
\end{equation}

\begin{equation}
\label{eq:multi_level_term}
\mathscr{L}_{i}(x_{i}, \tau_{i}) = -\log\left(\frac{\exp(x_{i}[\tau_{i}])}{\sum_{j=1}^{N_{i}} \exp(x_{i}[j])}\right)
                       = -x_{i}[\tau_{i}] + \log\left(\sum_{j=1}^{N_{i}} \exp(x_{i}[j])\right)  
\end{equation}




where, $\tau_{i}$ is the true label for the \textit{i-th} level. $x_{i} \in \mathbb{R}^{N_{i}}$, $\tau \in \mathbb{I}_{+}^L$.

$\mathcal{F}(\mathcal{I}) = x$ where, $x$ are the logits from the last layer of a model $\mathcal{F}$ which takes as input image $\mathcal{I}$. $x_{i}$ is a continuous sub-sequence of the predicted logits $x$, i.e. $x_{i} = (x_{i}[N_{i-1}+1], x_{i}[N_{i-1}+2], ..., x_{i}[N_{i-1}+N_{i}])$. 

\section{Marginalization (bottom up)}
In the baseline, the hierarchy-agnostic classifier, the model disregards the existence of a hierarchy in the labels. In the per-level classifier, the fact that each sample has exactly $L$ labels (due to the L-level hierarchy) is built into the design of the model by using $L$ such separate multi-class classifiers. The per-level classifier is still unaware of how these levels are ordered and is indifferent to the relation present between nodes from different levels. With the Marginalization method the information about the parents of each node is made available to the model.

The $L$-classifiers are replaced by a single classifier that outputs a probability distribution over the final level in the hierarchy. Instead of having classifiers for the remaining $L-1$ levels, we compute the probability distribution over each one of these by summing the probability of the children nodes. Although, the network does not explicitly predict these scores, the models is still penalized for incorrect predictions across the $L$-level hierarchy using the cross-entropy loss.

\begin{figure}[!htbp]
    \centering
    \includegraphics[width=0.9\textwidth]{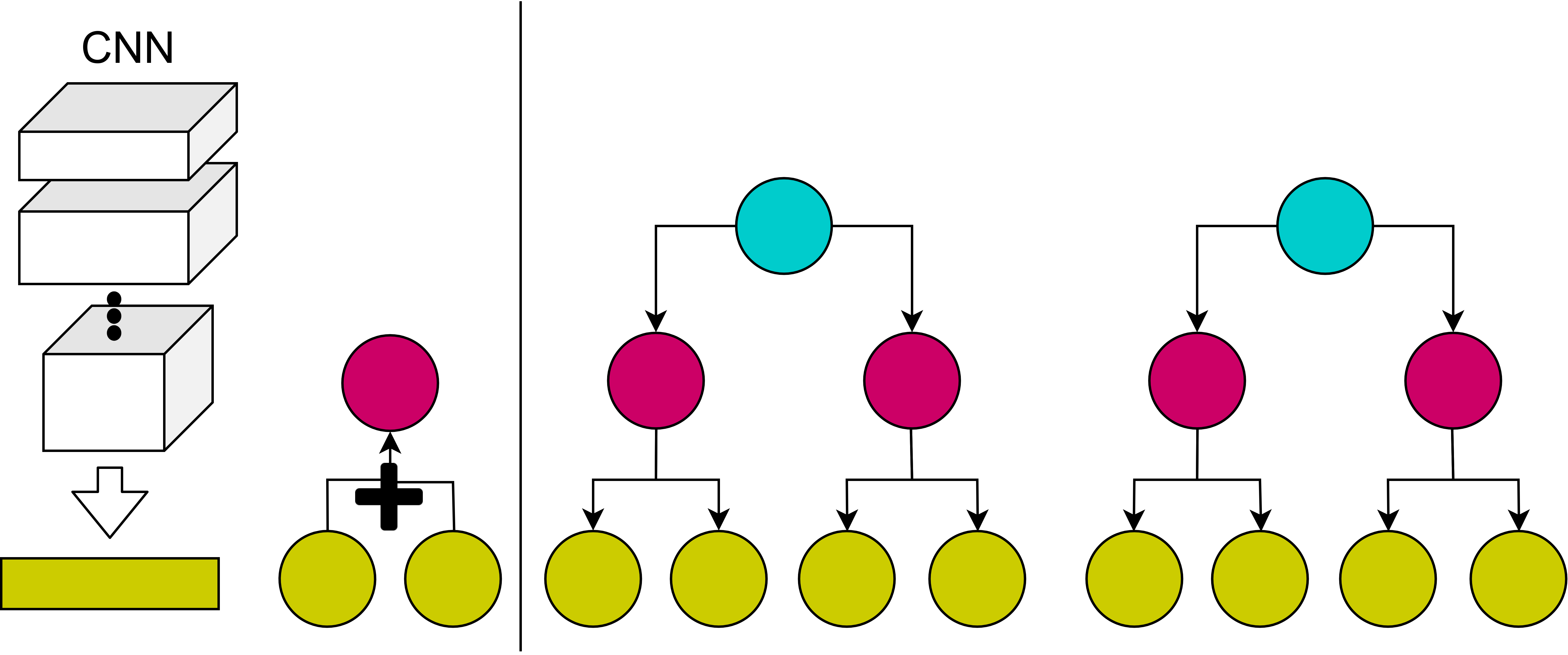}
    \caption{Model schematic for the Marginalization method. Instead of predicting a label per level, the model outputs a probability distribution over the leaves of the hierarchy. Probability for non-leaf nodes is determined by marginalizing over the direct descendants. The Marginalization method models how different nodes are connected among each other in addition to the fact that there are $L$ levels in the label-hierarchy.}
    \label{fig:bs3-marginalization-model-schematic}
\end{figure}

\begin{equation}
\label{eq:Marginalization (bottom up) term}
\mathscr{L}(x, \tau) = \sum_{i=1}^{L} \mathscr{L}_{i}(x_{i}, \tau_{i}) = -\sum_{i=1}^{L} \log\left(p_{i}[\tau_{i}]\right)
\end{equation}

where, $\tau_{i}$ is the true label for the \textit{i-th} level. $x_{i} \in \mathbb{R}^{N_{i}}$, $\tau \in \mathbb{I}_{+}^L$.

$\mathcal{F}(\mathcal{I}) = x$ where, $x$ are the logits from the last layer of a model $\mathcal{F}$ which takes as input image $\mathcal{I}$.

This is the same loss from \cref{eq:multi_level_term} however, the manner in which each $x_{i}$ is computed is different. The difference is that here the model predicts a probability distribution only over the leaf labels. To obtain a probability of a label that is a non-leaf label, the probabilities of the direct children are summed over and this marginalization results in the probability of the parent label. This way a valid probability distribution is obtained for each level in the hierarchy.

\begin{equation}
\label{eq:Marginalization (bottom up) x_i term}
p_{i}[j] = P(v_{i}^{j} | \mathcal{I}) = \sum_{c \in \text{childrenOf}(v_{i}^{j})} P(c | \mathcal{I}),  \forall i \in {1, 2, ..., (L-1)}
\end{equation}

where, $v_{i}^{j}$ is the \textit{j-th} vertex (node) in the \textit{i-th} level.


All but the last level use \cref{eq:Marginalization (bottom up) x_i term} to compute the probabilities for their labels.

\begin{equation}
\label{eq:Marginalization (bottom up) x_L term}
p_{L}[j] = P(v_{L}^{j} | \mathcal{I}) = \left(\frac{\exp(x_{j})}{\sum_{k=1}^{N_{L}} \exp(x_{k})}\right)
\end{equation}

For the final level, we compute the probability distribution over the leaf nodes by directly using the logits output from the model, $\mathcal{F}$. This computation is indicated in \cref{eq:Marginalization (bottom up) x_L term} using softmax. Once $p_{L}$ is determined, $p_{L-1}$ can be calculated. For this reason we compute the probabilities for the complete hierarchy in a bottom up fashion: starting from the bottom-most layer and moving to the upper levels. 

\section{Masked Per-level classifier}
On the upper levels of the hierarchy one has more data per label and fewer labels to choose from. Naturally, this makes classifying relatively accurate closer to the root of the hierarchy. This model exploits knowledge about the parent-child relationship between nodes in a top down manner.

\begin{figure}[!htbp]
    \centering
    \includegraphics[width=0.9\textwidth]{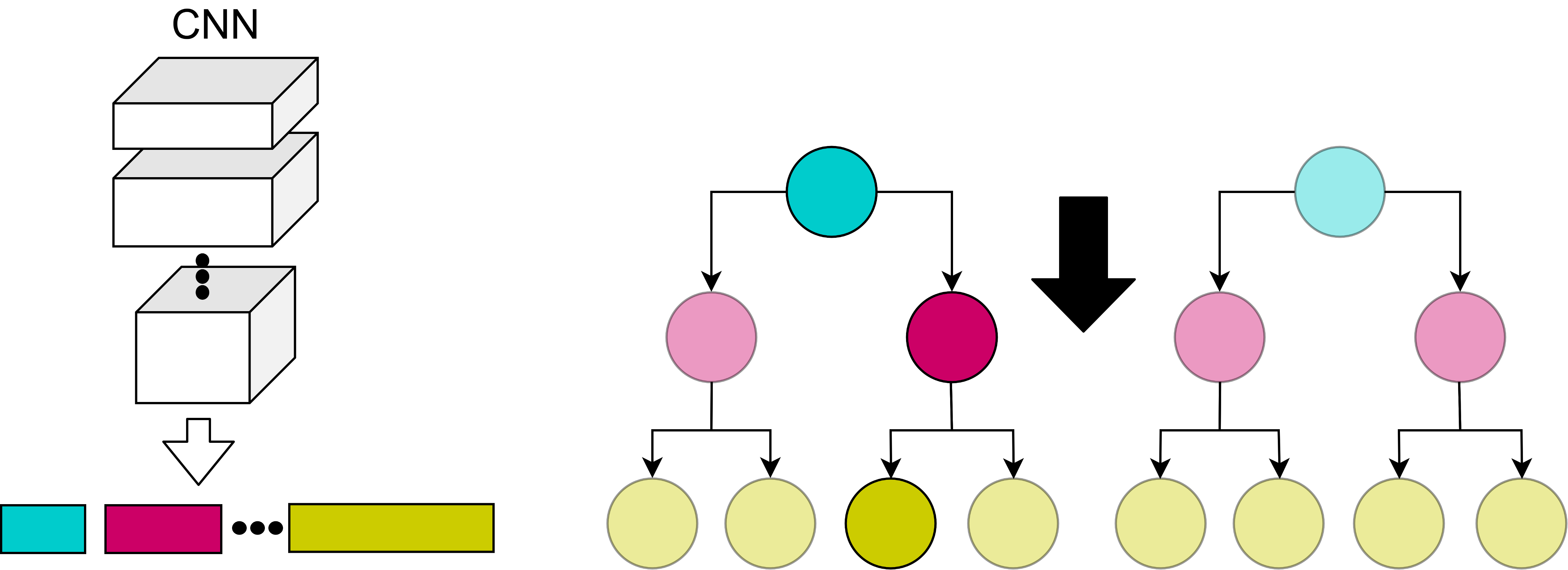}
    \caption{Model schematic for the Masked Per-level classifier. The model is trained exactly like the L $N_{i}$-way classifier. While predicting, one assumes the model performs better for upper levels than lower levels. Keeping this in mind, when predicting a label for a lower level, the model's prediction for the level above is used to mask all infeasible descendant nodes, assuming the model predicts correctly for the level above. This results in competition only among the descendants of the predicted label in the level above.}
    \label{fig:bs3-multi-level-masked-model-schematic}
\end{figure}

Unlike Marginalization (bottom up), here, we have L-classifiers, one for each hierarchical level. For the first level, the model predicts the class with the highest score among the logits. For consequent level $l_{i}$, the information about the models belief i.e. it's prediction for the $l_{i-1}$ level is leveraged. Instead of naively predicting the label with the highest score for level $l_{i}$ (comparing among all possible logits), all nodes except the children of the predicted label for level $l_{i-1}$ are masked. This translates to computing the loss over a subset of the original nodes in level $l_{i}$. With the availability of the parent-child relationship and assuming that the model predicts correctly the parent label (on level $l_{i-1}$, the only possible labels are the children of this predicted parent. As mentioned earlier, classification in the upper levels is more accurate and since we perform this in a top down fashion, this is a reasonable assumption. Another work has shown this to be the case \cite{tasho2018thesis}. For the last L-1 levels, only a subset of the logits (formed by the children of the predicted parent) are compared against each other, ignoring the rest.

While training, the loss is computed over the children of the parent conforming to the ground truth. Even if the model predicts the parent incorrectly, we still use the ground truth to penalize its prediction for the children.

For data with unknown ground truth i.e. during evaluation, the model uses the predictions from level $l_{i-1}$ to make infer about level $l_{i}$ by masking nodes that correspond to labels that are not possible.

\begin{equation}
\label{eq:Masked L-classifiers}
\mathscr{L}(x, \tau) = \sum_{i=1}^{L} \mathscr{L}_{i}(x_{i}, \tau_{i})
\end{equation}

\begin{multline}
\label{eq:Masked L-classifiers term}
\mathscr{L_i}(x_{i}, \tau_{i}) = -\log\left(\frac{\exp(x_{i}[\tau_{i}])}{\sum_{j \in C} \exp(x_{i}[j])}\right) 
                       = -x_{i}[\tau_{i}] + \log\left(\sum_{j \in C} \exp(x_{i}[j])\right)  
\end{multline}




where, $\tau_{i}$ is the true label for the \textit{i-th} level. $x_{i} \in \mathbb{R}^{N_{i}}$, $\tau \in \mathbb{I}_{+}^L$.
$C = \text{childrenOf}(v^{\tau_{i-1}}_{i-1})$.

$v_{i}^{j}$ is the \textit{j-th} vertex (node) in the \textit{i-th} level and consequently, $v^{\tau_{i-1}}_{i-1}$ is the node corresponding to the ground-truth on level $(i-1)$.

$\mathcal{F}(\mathcal{I}) = x$ where, $x$ are the logits from the last layer of a model $\mathcal{F}$ which takes as input image $\mathcal{I}$. $x_{i}$ is a continuous sub-sequence of the predicted logits $x$, i.e. $x_{i} = (x_{i}[N_{i-1}+1], x_{i}[N_{i-1}+2], ..., x_{i}[N_{i-1}+N_{i}])$. 

\section{Hierarchical Softmax}
For this the model predicts logits for every node in the hierarchy. The logits are predicted by dedicated linear layers for each group of siblings and consequently there is a separate probability distribution over each one of these groups. This is probability conditioned on the parent node i.e. $p(v_{i}^{j_{i}}|v_{i-1}^{j_{i-1}}), \forall v_{i}^{j_{i}} \in C$, such that $C = \text{childrenOf}(v_{i-1}^{j_{i-1}})$.

In the context of natural language processing something similar has been discussed in previous work \cite{morin2005hierarchical, mikolov2013hierarchical2}. But their main goal is to reduce the computational complexity over very large vocabularies. In the context of computer vision this is relatively unexplored and we propose to decompose the probability distribution and predict conditional distributions for each set of direct descendants in the hierarchy, in order to exploit the label-hierarchy and boost performance.

\begin{equation}
\label{eq:hierarchical_softmax_cond_prob_dist}
p(v_{i}^{j_{i}}|v_{i-1}^{j_{i-1}}) = \left(\frac{\exp(x_{v_{i-1}^{j_{i-1}}}[j_{i}])}{\sum_{k \in C} \exp(x_{v_{i-1}^{j_{i-1}}}[k])}\right) \
\forall v_{i}^{j_{i}} \in C,\ x_{v_{i-1}^{j_{i-1}}} \in \mathbb{R}^{|C|}
\end{equation}

The vector $x_{v_{i-1}^{j_{i-1}}}$ represents the logits that exclusively correspond to all the children of node $v_{i-1}^{j_{i-1}}$. With this in place, for the set of children of a give node, a conditional probability distribution is output by the model $\mathcal{F}$.

$\mathcal{F}(\mathcal{I}) = p(\cdot)$ where, $p(\cdot)$ is the conditional probability for every child node given the parent, $p(v_{i}^{j_{i}}|v_{i-1}^{j_{i-1}})$. $\mathcal{F}$ takes as input image $\mathcal{I}$.

In order to calculate the joint distribution over the leaves, probabilities along the path from the root to each leaf are multiplied.

\begin{equation}
\label{eq:hierarchical_softmax_factorize}
p(v_{1}^{j_1}, v_{2}^{j_2}, ..., v_{(L-1)}^{j_(L-1)}, v_{L}^{j_L}) = p(v_{1}^{j_1})p(v_{2}^{j_2}|v_{1}^{j_1})... p(v_{L}^{j_L}|v_{(L-1)}^{j_(L-1)})
\end{equation}

where, $v_{i}^{j_i}$ is the parent node of $v_{i+1}^{j_{(i+1)}}$. The nodes belonging to the \textit{i-th} level and the \textit{(i+1)-st} level respectively.

The cross-entropy loss is directly computed only over the leaves but since the distribution over the leaves implicitly uses the internal nodes for calculation, all levels are optimized over indirectly and the performance gradually improves (across all levels).

\begin{equation}
\label{eq:hierarchical_softmax_term}
\mathscr{L}(x, \tau) = -\log\left(p(v_{1}^{j_1}, v_{2}^{j_2}, ..., v_{(L-1)}^{j_(L-1)}, v_{L}^{\tau_{L}})\right) = -\log\left(p(v_{1}^{\tau_{1}}, v_{2}^{\tau_{2}}, ..., v_{(L-1)}^{\tau_{L-1}}, v_{L}^{\tau_{L}})\right)
\end{equation}

where, $\tau_{i}$ is the true label for the \textit{i-th} level. $x_{i} \in \mathbb{R}^{N_{i}}$, $\tau \in \mathbb{I}_{+}^L$.

\cref{eq:hierarchical_softmax_term} can be re-written as \cref{eq:hierarchical_softmax_term_simplified} because when $\tau_{L}$ is known the path to the root is unique and the remaining $\tau_{i}, \forall i \in {1, 2, ..., (L-1)}$ are determined.

\begin{equation}
\label{eq:hierarchical_softmax_term_simplified}
\mathscr{L}(x, \tau) = -\log\left(p(v_{1}^{\tau_{1}}, v_{2}^{\tau_{2}}, ..., v_{(L-1)}^{\tau_{L-1}}, v_{L}^{\tau_{L}})\right)
\end{equation}

\chapter{Empirical Analysis: Injecting label-hierarchy into CNN classifiers}
\label{ch:cnn_analysis}
In this Chapter, we describe the numerical experiments used to evaluate our methods that help classifiers exploit label-hierarchies. Before going into the experimental details, we discuss the choice of performance metrics to compare across different models.

\section{Performance metrics}
In order to quantify the performance we use micro and macro averaged scores. Although micro scores take contributions in proportion to the size of the class they end up overshadowing classes that occur less frequently. Such patterns are very much part of dataset with hierarchical labels as class higher up the hierarchy abstract their descendants and have more samples as compared to classes below with the leaves of the hierarchy having the least number of samples. The macro scores in contrast take an un-weighted average over the scores computed individually for all classes.

Consider a dataset as shown in  \cref{table:comparing_macro_and_micro_scores}. When using a classifier for each level in the hierarchy, the classifier prefers to blindly predict the majority label to boost its micro score. By always predicting Hesperiidae, Pyrginae and Pyrgus\_alveus it obtains a micro-averaged precision, recall and F1-score of (0.5, 0.5, 0.5). This type of behavior is undesirable. However, the macro-averaged scores are (0.1364, 0.2727, 0.1724) which reflect the poor performance of the classifier.

\begin{table}[htbp]
\centering
\begin{tabular}{| c || c | c |} 
 \hline
 Label name & Level & Frequency  \\ [0.5ex] 
 \hline\hline
Hesperiidae & Family & 76 \\
Riodinidae & Family & 16 \\ \hline
Hesperiinae & Subfamily & 36 \\   
Pyrginae & Subfamily & 40 \\  
Nemeobiinae & Subfamily & 16 \\ \hline
Ochlodes venata & Genus + Species & 18 \\
Hesperia comma & Genus + Species & 18 \\
Pyrgus alveus & Genus + Species & 22 \\
Spialia sertorius & Genus + Species & 18 \\
Hamearis lucina & Genus + Species & 14 \\
Polycaena tamerlana & Genus + Species & 2 \\
 [1ex] \hline
\end{tabular}
\caption{A subset of the ETHEC dataset to demonstrate the pros and cons of using macro and micro scoring.}
\label{table:comparing_macro_and_micro_scores}
\end{table}

To get better insight about where the model under-performs micro and macro averaged scores are also computer per level in the hierarchy.

\paragraph{True positive rate}
True positive rate (TPR) is the fraction of actual positives predicted correctly by the method.
\begin{equation}
    \text{TPR}=\frac{tp}{totalPositives}
\label{eq:TPR-formula}
\end{equation}
\paragraph{True negative rate}
True negative rate (TNR) is the fraction of actual negatives predicted correctly by the method.
\begin{equation}
    \text{TNR}=\frac{tn}{totalNegatives}
\label{eq:TNR-formula}
\end{equation}
\paragraph{Precision}
Precision computes what fraction of the labels predicted true by the model are actually true.
\begin{equation}
    \text{P}=\frac{tp}{tp+fp}
\label{eq:precision-formula}
\end{equation}
\paragraph{Recall}
Recall computes what fraction of the true labels were predicted as true.
\begin{equation}
    \text{R}=\frac{tp}{tp+fn}
\label{eq:recall-formula}
\end{equation}
\paragraph{F1-score}
\begin{equation}
    \text{F1}=\frac{2*P*R}{P+R}
\label{eq:f1-formula}
\end{equation}
\paragraph{Hit@k}
\begin{equation}
    \text{Hit@K}=\frac{1}{N} \sum_{i=1}^{N} 1[\text{label}_{i}^{\text{gt}} \in \text{SortedPredictions}(i)]
\label{eq:hit-at-k-formula}
\end{equation}
where, $\text{SortedPredictions}(i) = \{ \text{label}_{0}^{\text{pred}}, \text{label}_{1}^{\text{pred}},..., \text{label}_{k-1}^{\text{pred}}, \text{label}_{k}^{\text{pred}}\}$ is the set of the top-K predictions for the \textit{i}-th data sample.
\paragraph{Macro-averaged score}
A macro-averaged score for a metric is calculated by averaging the metric across all labels.
\begin{equation}
    \text{M-metric}=\frac{1}{N} \sum_{i=1}^{N} \text{metric}(\text{label}_{i})
\label{eq:macro-metric-formula}
\end{equation}



\paragraph{Micro-averaged score}
A micro-averaged score for a metric is calculated by accumulating contributions (to the performance metric) across all labels and these accumulated contributions are used to calculate the micro score.

\section{Hierarchical CIFAR-10}
To see the details of the hierarchy creation please refer to \cref{sec:datasets}.
\subsection{Per-level classifier}
\subsubsection{Influence of training set size on performance}

To test the influence of dataset size to performance, we run experiments by changing the size of the train set. We randomly pick 3 differently sized subsets of the dataset to see the effect on the classification performance and as a sanity check of the implementation.

We choose 3 configurations of the training set (\textit{all samples}, \textit{1000 samples}, \textit{100 samples}) and train for 100 epochs. We keep the validation and test set same through out as discussed in \cref{sec:datasets}. Refer to \cref{table:cifar10_training_set_size} for performance comparison. The numbers are reported on the unseen test set.

\begin{table}[!htbp]
\centering
\begin{tabular}{| c || c | c | c || c | c | c |} 
 \hline
  & m-P & m-R & m-F1 & M-P & M-R & M-F1  \\ [0.5ex] 
 \hline\hline
 \multicolumn{7}{|c|}{ResNet-50 (update all weights)} \\
 \hline\hline
 100 samples & 0.6129 & 0.6129 & 0.6129 & 0.5336 & 0.4924 & 0.4724\\ 
 1000 samples & 0.7947 & 0.7947 & 0.7947 & 0.7217 & 0.7234 & 0.7190\\
 all samples & \textbf{0.9358} & \textbf{0.9358} & \textbf{0.9358} & \textbf{0.9124} & \textbf{0.9101} & \textbf{0.9108} \\
 [1ex] \hline
\end{tabular}
\caption{Performance metrics for Per-level classifier on the Hierarchical CIFAR-10 data when varying the amount of training data. The models used in this experiment are pre-trained on the 1000-class ImageNet data set. All weights are updated with a learning rate of 0.01 and input spatial dimensions are 224x224. \textit{P}, \textit{R} and \textit{F1} represent Precision, Recall and F1-score. Metrics prefixed with \textit{m} are micro-averaged while the ones with \textit{M} are macro-averaged. The top performing models are in bold-face.}
\label{table:cifar10_training_set_size}
\end{table}

\subsection{Hierarchy-agnostic classifier}

In \cref{table:cifar10_n_way_multi_label} we summarize the results of hierarchy-agnostic classifier for the Hierarchical CIFAR-10 dataset. We show performance for different CNN models for feature extraction. In addition to that, models have either only their last layer fine-tuned (keeping the rest of the weights fixed) or all the weights in the model are updated.

\begin{table}[!htbp]
\centering
\begin{tabular}{| c || c | c | c | c |} 
 \hline
  & m-Precision & m-Recall & m-F1 & M-F1  \\ [0.5ex] 
 \hline\hline
 \multicolumn{5}{|c|}{Alexnet} \\
 \hline\hline
 Per-class decision boundary & 0.7311 & 0.7863 & 0.7577 & 0.6814 \\
 One-fits-all decision boundary & 0.7687 & 0.7564 & 0.7625 & 0.6708 \\
 \hline\hline
 \multicolumn{5}{|c|}{Alexnet (update all weights)} \\
 \hline\hline
 Per-class decision boundary & 0.9136 & 0.9033 & 0.9085 & 0.8736\\
 One-fits-all decision boundary & 0.9185 & 0.8968 & 0.9075 & 0.8683 \\
 \hline\hline
 \multicolumn{5}{|c|}{VGG} \\
 \hline\hline
 Per-class decision boundary & 0.7461 & 0.8018 & 0.7729 & 0.7014 \\ 
 One-fits-all decision boundary & 0.7847 & 0.7805 & 0.7826 & 0.6961 \\
 \hline\hline
 \multicolumn{5}{|c|}{VGG (update all weights)} \\
 \hline\hline
 Per-class decision boundary & 0.9296 & 0.9114 & 0.9204 & 0.8888 \\ 
 One-fits-all decision boundary & 0.9300 & 0.9156 & 0.9228 & 0.8910 \\
 \hline\hline
 \multicolumn{5}{|c|}{ResNet-18} \\
 \hline\hline
 Per-class decision boundary & 0.7369 & 0.7649 & 0.7507 & 0.6710 \\
 One-fits-all decision boundary & 0.7591 & 0.7613 & 0.7602 & 0.6705 \\
 \hline\hline
 \multicolumn{5}{|c|}{ResNet-18 (update all weights)} \\
 \hline\hline
 Per-class decision boundary & 0.9437 & 0.9282 & 0.9359 & 0.9098 \\ 
 One-fits-all decision boundary & 0.9450 & 0.9276 & 0.9362 & 0.9107 \\
 \hline\hline
 \multicolumn{5}{|c|}{ResNet-50} \\
 \hline\hline
 Per-class decision boundary & 0.7544 & 0.7946 & 0.7740 & 0.6977 \\
 One-fits-all decision boundary & 0.7922 & 0.7729 & 0.7824 & 0.6980 \\
 \hline\hline
 \multicolumn{5}{|c|}{ResNet-50 (update all weights)} \\
 \hline\hline
 Per-class decision boundary & 0.9448 & 0.9283 & 0.9365 & 0.9097 \\ 
 One-fits-all decision boundary & \textbf{0.9538} & \textbf{0.9361} & \textbf{0.9448} & \textbf{0.9227} \\
 [1ex] \hline
\end{tabular}
\caption{Performance metrics for the hierarchy-agnostic classifier on the Hierarchical CIFAR-10 data. The models used in this experiment are pre-trained on the 1000-class ImageNet data set. For these experiments, only the last layer is fine-tuned (unless mentioned otherwise), fixing the rest of the weights with a learning rate of 0.01 and input spatial dimensions of 224x224 for 100 epochs. Metrics prefixed with \textit{m} are micro-averaged while the ones with \textit{M} are macro-averaged. The top performing models are in bold-face.}
\label{table:cifar10_n_way_multi_label}
\end{table}

\clearpage

\section{Hierarchical Fashion MNIST}
To see the details of the hierarchy creation please refer to \cref{sec:datasets}.

\subsection{Hierarchy-agnostic classifier}

As for the Hierarchical CIFAR-10, we also use the same hierarchy-agnostic model on the Hierarchical Fashion MNIST dataset with results tabulated in \cref{table:FMNIST_n_way_multi_label}. In both cases the ResNet \cite{he2016resnet} backbone seems to be the best performing for micro and macro averaged F1-score.

\begin{table}[!htbp]
\centering
\begin{tabular}{| c || c | c | c | c |} 
 \hline
  & m-Precision & m-Recall & m-F1 & M-F1  \\ [0.5ex] 
 \hline\hline
 \multicolumn{5}{|c|}{Alexnet} \\
 \hline\hline
 Per-class decision boundary & 0.8086 & 0.8451 & 0.8264 & 0.7818 \\
 One-fits-all decision boundary & 0.8822 & 0.8066 & 0.8427 & 0.7767 \\
 \hline\hline
 \multicolumn{5}{|c|}{Alexnet (update all weights)} \\
 \hline\hline
 Per-class decision boundary & 0.9145 & 0.9114 & 0.9129 & 0.8847 \\
 One-fits-all decision boundary & 0.9321 & 0.9004 & 0.9160 & 0.8831 \\
 \hline\hline
 \multicolumn{5}{|c|}{VGG} \\
 \hline\hline
 Per-class decision boundary & 0.7705 & 0.7818 & 0.7761 & 0.7169 \\ 
 One-fits-all decision boundary &  0.8311 & 0.7668 & 0.7976 &  0.7179 \\
 \hline\hline
 \multicolumn{5}{|c|}{VGG (update all weights)} \\
 \hline\hline
 Per-class decision boundary & 0.9349 & 0.9219 & 0.9284 & 0.9030 \\ 
 One-fits-all decision boundary & 0.9390 & 0.9207 & 0.9297 & 0.9040 \\
 \hline\hline
 \multicolumn{5}{|c|}{ResNet-18} \\
 \hline\hline
  Per-class decision boundary & 0.7911 & 0.8339 & 0.8119 & 0.7692 \\
  One-fits-all decision boundary & 0.8424 & 0.7920 & 0.8164 & 0.7531 \\
 \hline\hline
 \multicolumn{5}{|c|}{ResNet-18 (update all weights)} \\
 \hline\hline
 Per-class decision boundary & 0.9372 & 0.9325 & 0.9348 & 0.9124 \\ 
 One-fits-all decision boundary & \textbf{0.9503} & 0.9248 & 0.9374 & 0.9132 \\
 \hline\hline
 \multicolumn{5}{|c|}{ResNet-50} \\
 \hline\hline
 Per-class decision boundary & 0.7924 & 0.8276 & 0.8096 & 0.7706 \\
 One-fits-all decision boundary & 0.8546 & 0.8029 & 0.8280 & 0.7641 \\
 \hline\hline
 \multicolumn{5}{|c|}{ResNet-50 (update all weights)} \\
 \hline\hline
 Per-class decision boundary & 0.9338 & 0.9330 & 0.9334 & 0.9114 \\ 
 One-fits-all decision boundary & 0.9389 & \textbf{0.9383} & \textbf{0.9386} & \textbf{0.9164} \\
 [1ex] \hline
\end{tabular}
\caption{Performance metrics for the hierarchy-agnostic classifier on the Hierarchical FMNIST data. The models used in this experiment are pre-trained on the 1000-class ImageNet data set. For these experiments, only the last layer and the first layer is fine-tuned (unless mentioned otherwise), fixing the rest of the weights with a learning rate of 0.01 and input spatial dimensions of 224x224. Metrics prefixed with \textit{m} are micro-averaged while the ones with \textit{M} are macro-averaged. The top performing models are in bold-face.}
\label{table:FMNIST_n_way_multi_label}
\end{table}

\clearpage

\section{ETHEC Dataset}
This sections describes the experiments and their results on the ETHEC dataset \cite{dhall_20.500.11850/365379}. In order to keep the section compact we perform experiments with best performing CNN backbones on the Hierarchical CIFAR-10 and FMNIST which is the ResNet-50.

The presence of a hierarchy introduces imbalance in the number of samples across labels in different levels of the hierarchy. In addition to this kind of imbalance that the ETHEC dataset also suffers from different number of samples for labels in the same level, unlike CIFAR-10 and FMNIST.

To tackle such imbalance in the ETHEC dataset, experiments are performed where the dataset is re-sampled and the term corresponding to every class in the loss function is re-weighed.

Re-sampling is performed proportional to the inverse frequency of the occurrence of a label in the dataset which are then normalize to sum to 1 to sample using a multinomial distribution. For more details please refer to PyTorch documentation \cite{pytorch}. Similarly, the loss is re-weighted using the per-class inverse frequencies as the to scale the corresponding terms contributed by data belonging to a particular class. 

\subsection{Hierarchy-agnostic classifier}
\label{subsec:multi_label_classifier}

Before, running the experiments on the complete dataset, a small subset of the original dataset was created ETHEC Merged Small. This dataset was used to debug the learning framework and ensure that we can learn to distinguish when there are a handful of classes in the 4-level structure. The labels used at different levels can be found in \cref{table:ethecsmall_dataset}.

\begin{table}[!htbp]
\centering
\begin{tabular}{| c | c || c | c | c || c | c | c || c |} 
 \hline
  cw & rs & m-P & m-R & m-F1 & M-P & M-R & M-F1 & (min, max), $\mu \pm \sigma$ \\ [0.5ex] 
 \hline\hline
 \multicolumn{9}{|c|}{ResNet-50 - Per-class decision boundary} \\
 \hline\hline
  
  \ding{55} & \ding{55} & 0.0355 & 0.7232 & 0.0677 & 0.3066 & 0.4053 & 0.2195 & (3, 351), 81.42 $\pm$ 69.51 \\
  \ding{55} & \ding{51} & 0.7159 & 0.7543 & 0.7346 & \textbf{0.4402} & 0.4362 & \textbf{0.3718} & (0, 13), 4.21 $\pm$ 2.07 \\
  \ding{51} & \ding{55} & 0.0077 & \textbf{0.8702} & 0.0153 & 0.0120 & \textbf{0.8397} & 0.0183 & (84, 718), 451.14 $\pm$ 136.69 \\
  \ding{51} & \ding{51} & 0.0081 & 0.7519 & 0.0161 & 0.0105 & 0.5909 & 0.0165 & (33, 714), 369.96 $\pm$ 120.55 \\
 \hline\hline
 \multicolumn{9}{|c|}{ResNet-50 - One-fits-all decision boundary} \\
 \hline\hline
  \ding{55} & \ding{55} & 0.9324 & 0.7235 & \textbf{0.8147} & 0.1913 & 0.1462 & 0.1568 & (0, 7), 3.10 $\pm$ 1.16 \\
  \ding{55} & \ding{51} & \textbf{0.9500} & 0.6564 & 0.7763 & 0.1078 & 0.0947 & 0.0959 & (0, 5), 2.76 $\pm$ 0.60 \\
   \ding{51} & \ding{55} & 0.2488 & 0.2960 & 0.2704 & 0.0021 & 0.0067 & 0.0030 & (4, 9), 4.76 $\pm$ 0.76 \\
  \ding{51} & \ding{51} & 0.1966 & 0.3800 & 0.2591 & 0.0027 & 0.0110 & 0.0037 & (4, 10), 7.73    $\pm$ 0.61 \\
 [1ex] \hline
\end{tabular}
\caption{Performance metrics for the hierarchy-agnostic classifier on the ETHEC Merged dataset. The models used in this experiment are pre-trained on the 1000-class ImageNet data set. All weights are updated with a learning rate of 0.01, a batch-size of 64 and input spatial dimensions are 224x224 for 100 epochs. \textit{P}, \textit{R} and \textit{F1} represent Precision, Recall and F1-score; \textit{cw} and \textit{rs} represent class weight and re-sampling. Metrics prefixed with \textit{m} are micro-averaged while the ones with \textit{M} are macro-averaged. The top performing models are in bold-face. Since, the model can predict any number of labels (between 0 and $N_{total}$), the table includes the minimum and the maximum number of labels predicted \textit{(min, max)} as well as the number of labels predicted on average $\mu \pm \sigma$. These statistics, like the rest, are calculated for samples in the \texttt{test} set.}
\label{table:ethec_merged_ft_multi_label}
\end{table}

The hierarchy-agnostic classifier disregards the hierarchy in the dataset and treats all levels in the same manner. This classifier model essentially flattens the hierarchical structure and performs multi-label classification to predict the correct labels given an image of a specimen. For every label, the model predicts whether the image is a member or not for that class. There is no restriction on the number of predictions that the model should make; as information about the number of levels (in the hierarchy) is unavailable to the model.

\subsubsection{Per-class decision boundary (PCDB) models}
The ill-effects of such free rein are reflected in \cref{table:ethec_merged_ft_multi_label}. Models with a high average number of predictions, especially the per-class decision boundary (PCDB) models, have high recall as they predict a lot more than just 4 labels for a given image. Predicting the image's membership in a lot of classes improves the chances of predicting the correct label but at the cost of a large number of false positives.

The models have high recall and that is due to them predicting a lot of false positives as the precision is on the other extreme of the spectrum and generally very poor in comparison to the recall scores. The (min, max), $\mu \pm \sigma$ column clearly shows the reckless behavior of the model predicting a maximum of 718 labels for one such sample and 451.14 $\pm$ 136.69 on average for the worst performing multi-label model in our experiments.

\subsubsection{One-fits-all decision boundary (OFADB) models}
On the other hand, the one-fits-all decision boundary (OFADB) models have a high micro scores but the macro scores are worse than those of the PCDB. Since, models with a better micro-F1 score are preferred, and they all share a common decision boundary for every label, the model sets this decision boundary in such a way that labels with a large number of samples are predicted correctly because that boosts the micro scores. Due to a commonly shared global threshold the models are more conservative in exercising the free rein as is evident from the $(min, max), \mu \pm \sigma$ column.

The one-fits-all decision boundary (OFADB) performs better than the same model with per-class decision boundaries (PCDB). We believe that the OFADB prevents over-fitting, especially in the case when many labels have very few data samples to learn from, which is the case for the ETHEC database. Here too, the nature of the multi-label setting allows the model to predict as many labels as it wants however, there is a marked difference between the (min, max), $\mu \pm \sigma$ statistics when comparing between the OFADB and PCDB. The best performing OFADB model predicts 3.10 $\pm$ 1.16 labels on average. This is close to the correct number of labels per specimen which is equal to the 4 levels in the label hierarchy.

\subsubsection{Loss reweighing and Data re-sampling}
Both data re-sampling and loss re-weighing remedy imbalance across different labels but via different paradigms. Instead of modifying what the model sees during training, reweighing the loss instead penalizes different data points differently. We choose to use the inverse-frequency of the label as weights that scale loss corresponding to the data point belonging to a particular label.

re-sampling involves by presenting the model a modified training set. This is done by choosing to show some samples multiple times while omitting others by over-sampling and under-sampling. Since, we wish to prevent the model from being biased by the population of data belonging to a particular label, ideally, we would like to present equal number of samples for each label in the dataset. We perform re-sampling based on the inverse-frequency of a label in the \texttt{train} set. If a label has an exceedingly large number of samples (more than the average) then it would be under-sampled while data belonging to less frequent labels would be over-sampled.

In the context of boosting, \cite{seiffert2008resampling} show that empirically, re-sampling is a better strategy than reweighing when the dataset consists of over-represented and under-represented classes. In our experiments as well, re-sampling significantly outperforms loss reweighing.


\begin{figure*}
    \centering
    \begin{subfigure}[b]{0.480\textwidth}
        \centering
        \includegraphics[width=\textwidth]{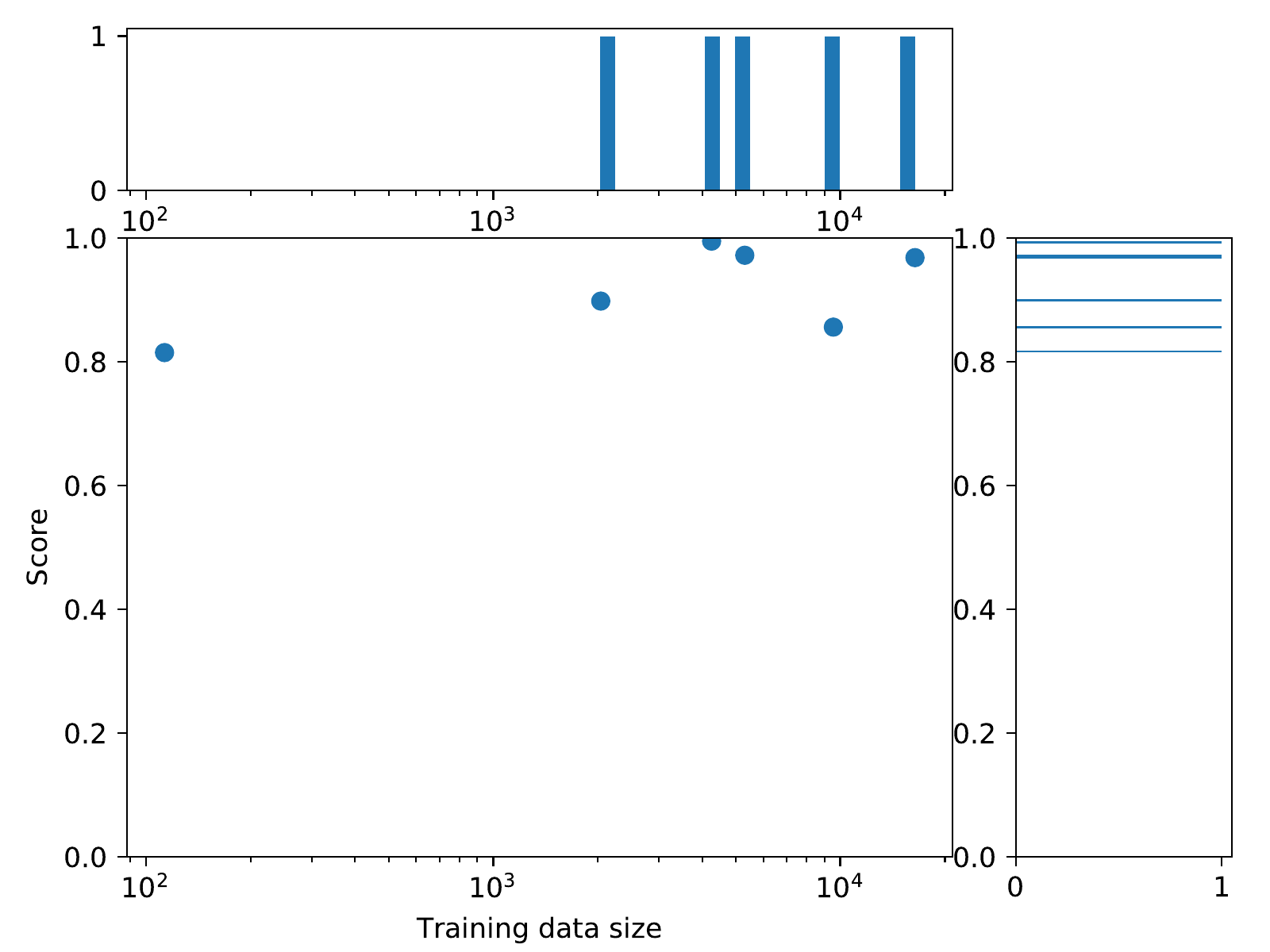}
        \caption[Network2]%
        {{\small \textit{family}}}    
        \label{fig:family multi_label}
    \end{subfigure}
    \hfill
    \begin{subfigure}[b]{0.480\textwidth}  
        \centering 
        \includegraphics[width=\textwidth]{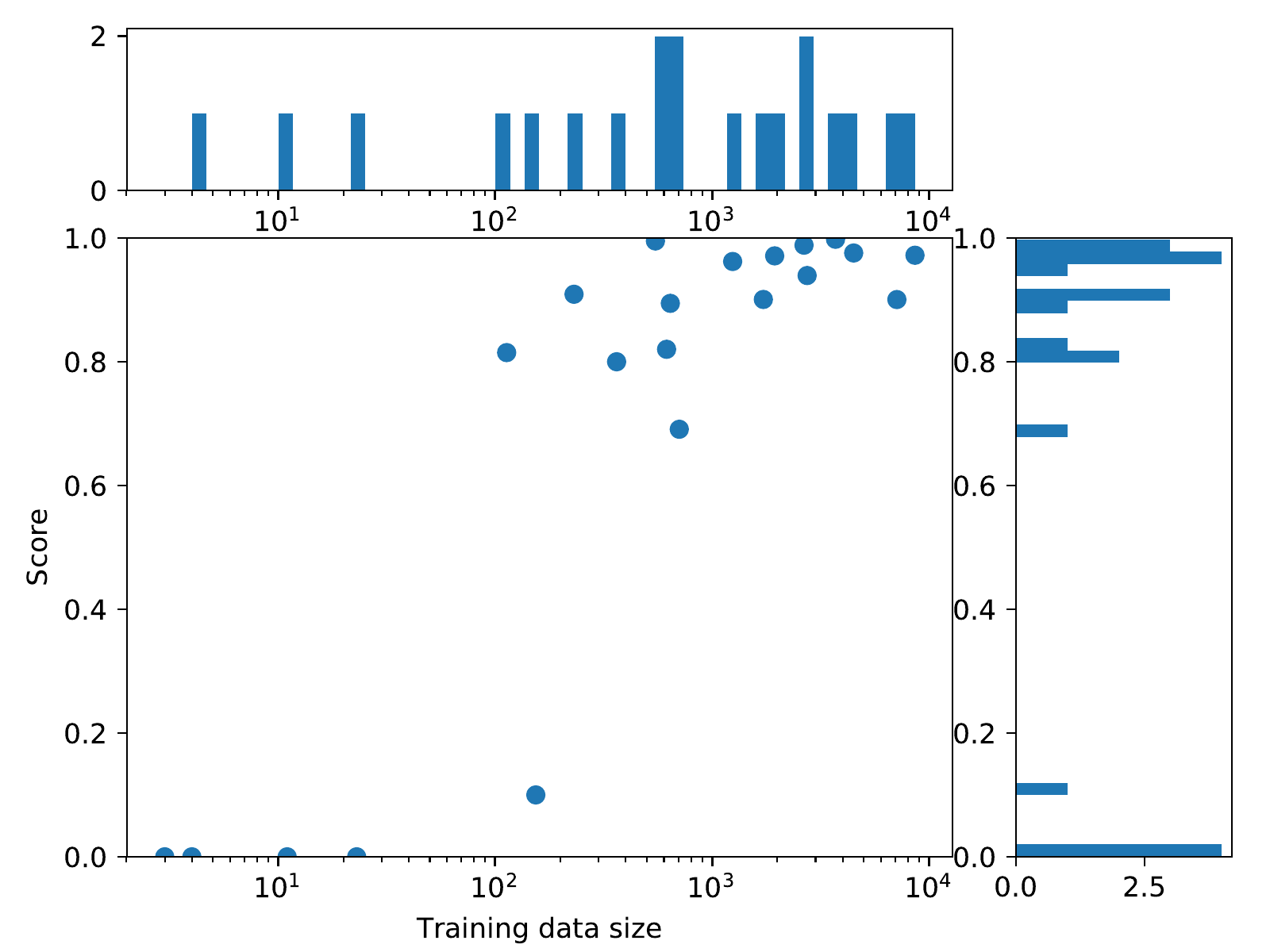}
        \caption[]%
        {{\small \textit{subfamily}}}    
        \label{fig:subfamily multi_label}
    \end{subfigure}
    \vskip\baselineskip
    \begin{subfigure}[b]{0.480\textwidth}   
        \centering 
        \includegraphics[width=\textwidth]{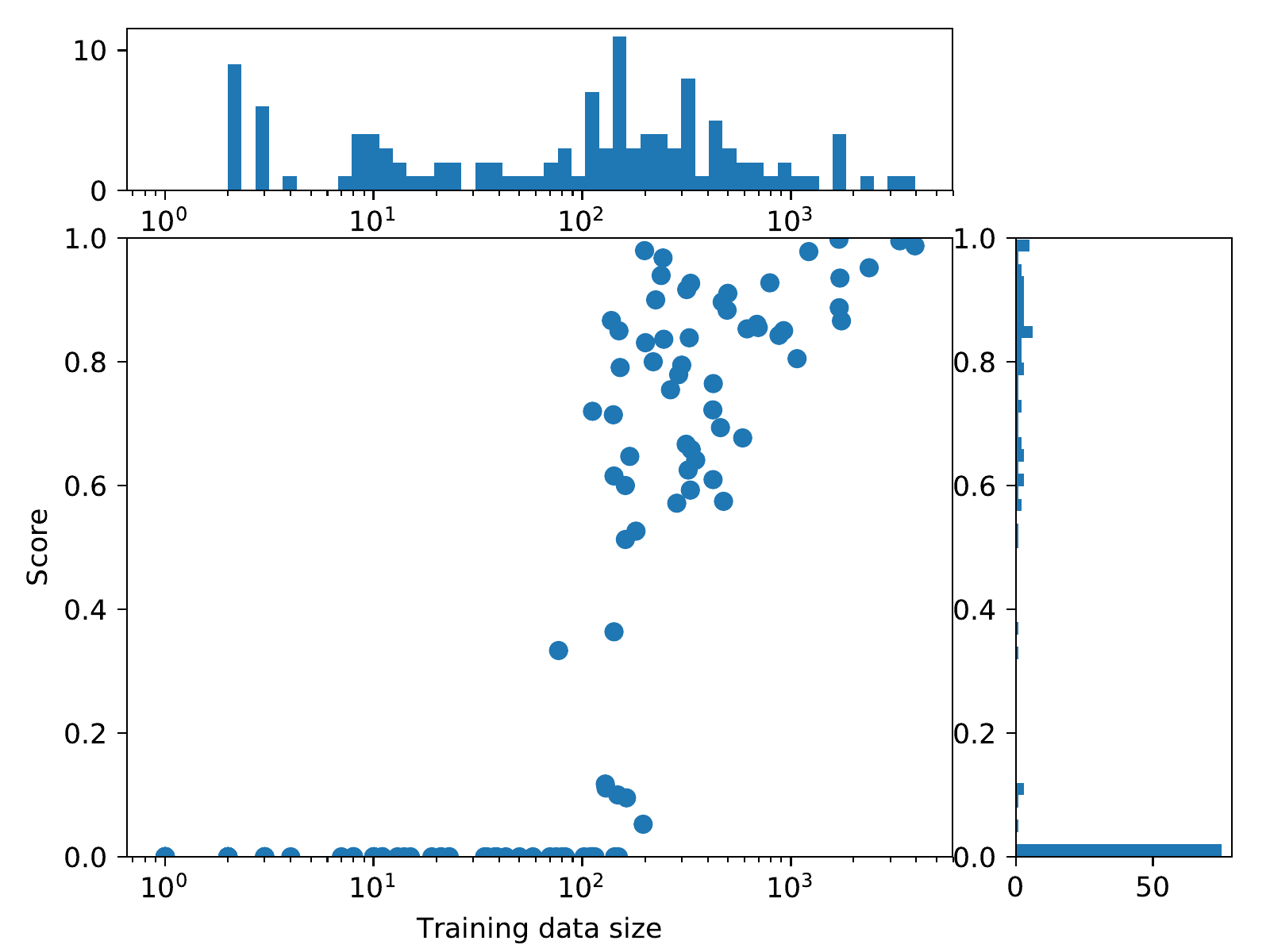}
        \caption[]%
        {{\small \textit{genus}}}    
        \label{fig:genus multi_label}
    \end{subfigure}
    \quad
    \begin{subfigure}[b]{0.480\textwidth}   
        \centering 
        \includegraphics[width=\textwidth]{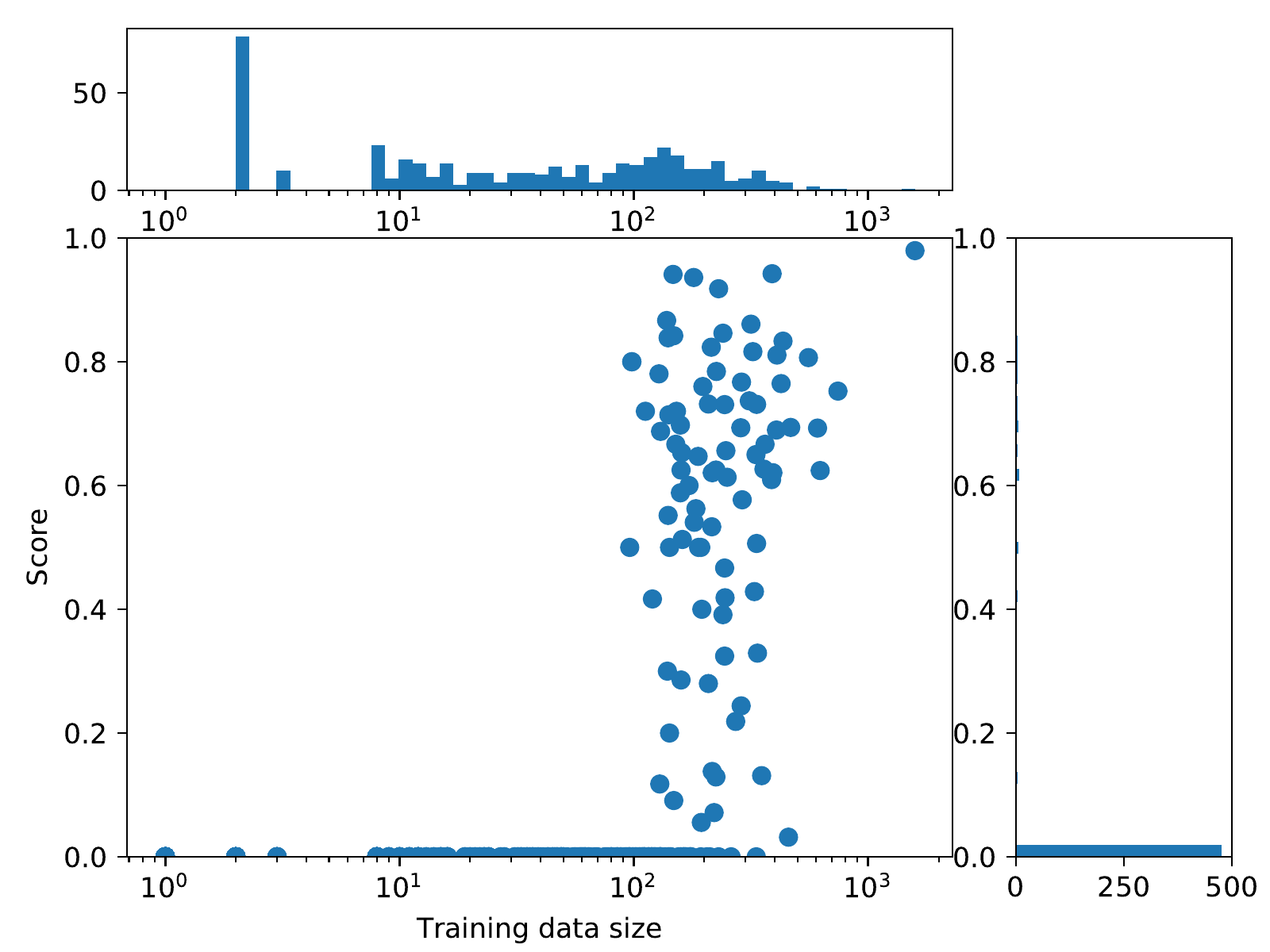}
        \caption[]%
        {{\small \textit{genus + specific epithet}}}    
        \label{fig:genus + specific epithet multi_label}
    \end{subfigure}
    \caption[]
    {\small \centering Per-label F1 performance across levels plotted against number of training samples for hierarchy-agnostic classifier using ResNet-50 (cw: \ding{55}, rs: \ding{55}). For each panel, a point represents a label with its position indicating the number of samples corresponding to that label in the \texttt{train} set and the F1-score that particular label achieve on the \texttt{test} set. To better visualize, the distribution of the samples in the \texttt{train} set and also the distribution of the performance, we display the marginal histograms for both training data size (on top of each scatter plot) and the F1-score (on the right of each scatter plot).} 
    \label{fig:multi_label_wo_cw_wo_rs_4x}
\end{figure*}

\subsubsection{Performance breakdown}

From \cref{fig:multi_label_wo_cw_wo_rs_4x} one can see the relation between number of training samples for a particular label and it's F1 performance. The 5 out of the 6 \textit{families} have more than 1,000 training data and the model performs well for these classes. However, the importance of a larger dataset is visible in the performance versus training data size plots in \cref{fig:subfamily multi_label}, \cref{fig:genus multi_label} and \cref{fig:genus + specific epithet multi_label}. There seems to be a ``S-shaped'' or sigmoid like trend in performance as the number of training data increases. Labels, across the 4 levels, that have more than 1,000 training samples have very high F1 scores while labels that have less than 100 training points are on the lower end of the spectrum.

The micro-F1 performance for the levels: \textit{family} and \textit{subfamily} is comparable to the best performing multi-level classifier (see \cref{table:ethec_merged_ft_multi_level_level_split}). But moving down to the lower levels in the hierarchy the performance is much worse than the best performing multi-level classifier as detailed in \cref{subsec:multi_level_classifier}.

\begin{table}[!htbp]
\centering
\begin{tabular}{| c | c || c | c | c || c | c | c |} 
 \hline
  Level & $N_{i}$ & m-P & m-R & m-F1 & M-P & M-R & M-F1  \\ [0.5ex] 
 \hline\hline
 \multicolumn{8}{|c|}{ResNet-50 (OFADB) with resampler (cw: \ding{55}, rs: \ding{55})} \\
 \hline\hline
  \textit{family} & 6 & 0.9861 & 0.9012 & 0.9417 & 0.9718 & 0.8801 & 0.9173 \\
  \textit{subfamily} & 21 & 0.9860 & 0.9065 & 0.9446 & 0.7941 & 0.6548 & 0.6968 \\
  \textit{genus} & 135 & 0.9290 & 0.7518 & 0.8311 & 0.3918 & 0.2961 & 0.3212 \\
  \textit{genus} + \textit{specific epithet} & 561 & 0.7249 & 0.3345 & 0.4578 & 0.1121 & 0.0832 & 0.0888  \\
 [1ex] \hline
\end{tabular}
\caption{Performance metrics for hierarchy-agnostic classifier on the ETHEC Merged dataset with the best performing configuration in its category. The model used in this experiment are pre-trained on the 1000-class ImageNet data set. All weights are updated with a learning rate of 0.01, a batch-size of 64 and input spatial dimensions are 224x224 for 100 epochs. \textit{P}, \textit{R} and \textit{F1} represent Precision, Recall and F1-score. Metrics prefixed with \textit{m} are micro-averaged while the ones with \textit{M} are macro-averaged.}
\label{table:ethec_merged_multi_label_level_split}
\end{table}

\clearpage

\subsection{Per-level classifiers}
\label{subsec:multi_level_classifier}
For the experiments reported here, we use the complete dataset instead of a subset.

As mentioned in \cref{subsec:multi_label_classifier}, due to the imbalanced nature of the ETHEC dataset we perform experiments with a combination of data re-sampling and loss re-weighing. The results are summarized in \cref{table:ethec_merged_ft_multi_level}.

Right off the bat, it is empirically seen that this model outperforms the model from \cref{subsec:multi_label_classifier}. Not only does it perform better on the global metrics but also the 4-level split metrics shown in \cref{table:ethec_merged_ft_multi_level_level_split} in comparison to \cref{table:ethec_merged_multi_label_level_split}.

The fact that the model is informed about the presence of different levels and that each level has one and only one correct label makes a significant difference in the performance. Instead of being able to predict as many labels as it wants (which was the case in the multi-label classifier) here, the model is designed such that is constrained to predict a single label per level in the hierarchy. 

Additionally, only logits belonging to the same hierarchical level compete against each other. By incorporating more information about the hierarchy the model outperforms the hierarchy-agnostic classifier. The best hierarchy-agnostic classifier has a m-F1 and M-F1 of 0.8147 and 0.3718 (from two separate models) while the best performing per-level classifier has a m-F1 and M-F1 of 0.9084 and 0.6888 (from the same model).

\begin{table}[!htbp]
\centering
\begin{tabular}{| c | c || c | c | c || c | c | c |} 
 \hline
  cw & rs & m-P & m-R & m-F1 & M-P & M-R & M-F1  \\ [0.5ex] 
 \hline\hline
 \multicolumn{8}{|c|}{ResNet-50} \\
 \hline\hline
  \ding{51} & \ding{55} & 0.8483 & 0.8483 & 0.8483 & 0.6648 & 0.6789 & 0.6411 \\
  \ding{55} & \ding{55} & 0.8930 & 0.8930 & 0.8930 & 0.6854 & 0.7094 & 0.6677 \\
  \ding{55} & \ding{51} & \textbf{0.9084} & \textbf{0.9084} & \textbf{0.9084} & \textbf{0.7134} & \textbf{0.7223} & \textbf{0.6888} \\
  \ding{51} & \ding{51} & 0.8760 & 0.8760 & 0.8760 & 0.6782 & 0.6874 & 0.6537 \\ \hline
  \ding{55} & \texttt{sqrt} & 0.9067 & 0.9067 & 0.9067 & 0.6941 & 0.7073 & 0.6700 \\ [1ex]
  \hline
\end{tabular}
\caption{Performance metrics for Per-level classifier on the ETHEC Merged dataset. The models used in this experiment are pre-trained on the 1000-class ImageNet data set. All weights are updated with a learning rate of 0.01, a batch-size of 64 and input spatial dimensions are 224x224 for 100 epochs. \textit{P}, \textit{R} and \textit{F1} represent Precision, Recall and F1-score; \textit{cw} and \textit{rs} represent class weight and re-sampling. Metrics prefixed with \textit{m} are micro-averaged while the ones with \textit{M} are macro-averaged. The top performing models are in bold-face. \texttt{sqrt} in the ``rs'' denotes re-sampling using the inverse of the square-root of the class frequency as weights, in other cases the inverse of the class frequency is used.}
\label{table:ethec_merged_ft_multi_level}
\end{table}

It is important to notice the trend between the macro and micro-averaged scores. In \cref{table:ethec_merged_ft_multi_level} all micro-averaged scores are always higher than their macro-averaged counterparts which is a direct result of optimizing the model for best performance measured by the micro-F1 score, which is a function of the micro precision and micro recall.

Micro-averaged scores are calculated by computing a ``combined" confusion matrix across all labels and then computing the (averaged) score, in this case the precision, recall and F1-score. Here, the notion of ``averaging'' comes from the fact that individual, label-wise confusion matrix elements (true positive, false positives, true negatives and false negatives) are combined in to a single global confusion matrix. While for the macro-averaged scores, the per-label scores i.e. precision, recall and F1-score are calculated and then their mean across all labels results in the macro-averaged scores.

If the micro scores are higher than the macro scores it is an indicator of the classes with fewer samples being classified incorrectly while the more popular classes are being classified correctly and eventually inflating the micro score. On the other hand, higher macro scores would indicate that labels with a large number of samples are being misclassified.

In our case it is the former, where the micro scores dominate the macro scores implying that the model performs well for classes with more training data which is intuitive as the model has more data to train on and sees a variety of samples for the same label in contrast to classes with only a handful of samples.

\begin{figure*}[!htbp]
    \centering
    \begin{subfigure}[b]{0.482\textwidth}
        \centering
        \includegraphics[width=\textwidth]{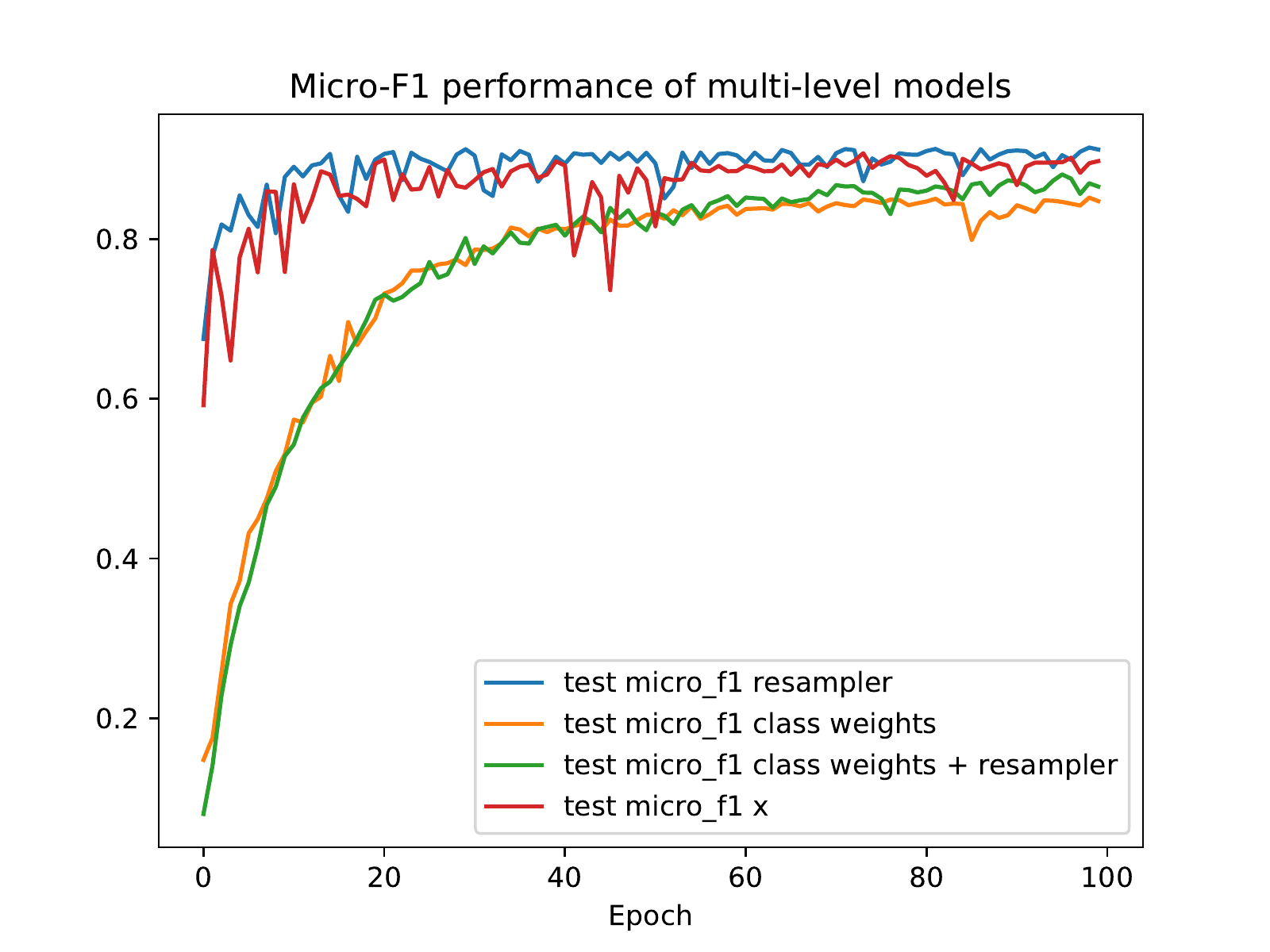}
        \caption[]%
        {{\small Micro-F1 performance over 100 epochs for ResNet-50 multi-level classifier with different combinations of resampler and class weights. Legend: \texttt{x} = cw: \ding{55}, rs: \ding{55}}}    
        \label{fig:micro_f1 rs cw combination combined plot}
    \end{subfigure}
    \hfill
    \begin{subfigure}[b]{0.482\textwidth}  
        \centering 
        \includegraphics[width=\textwidth]{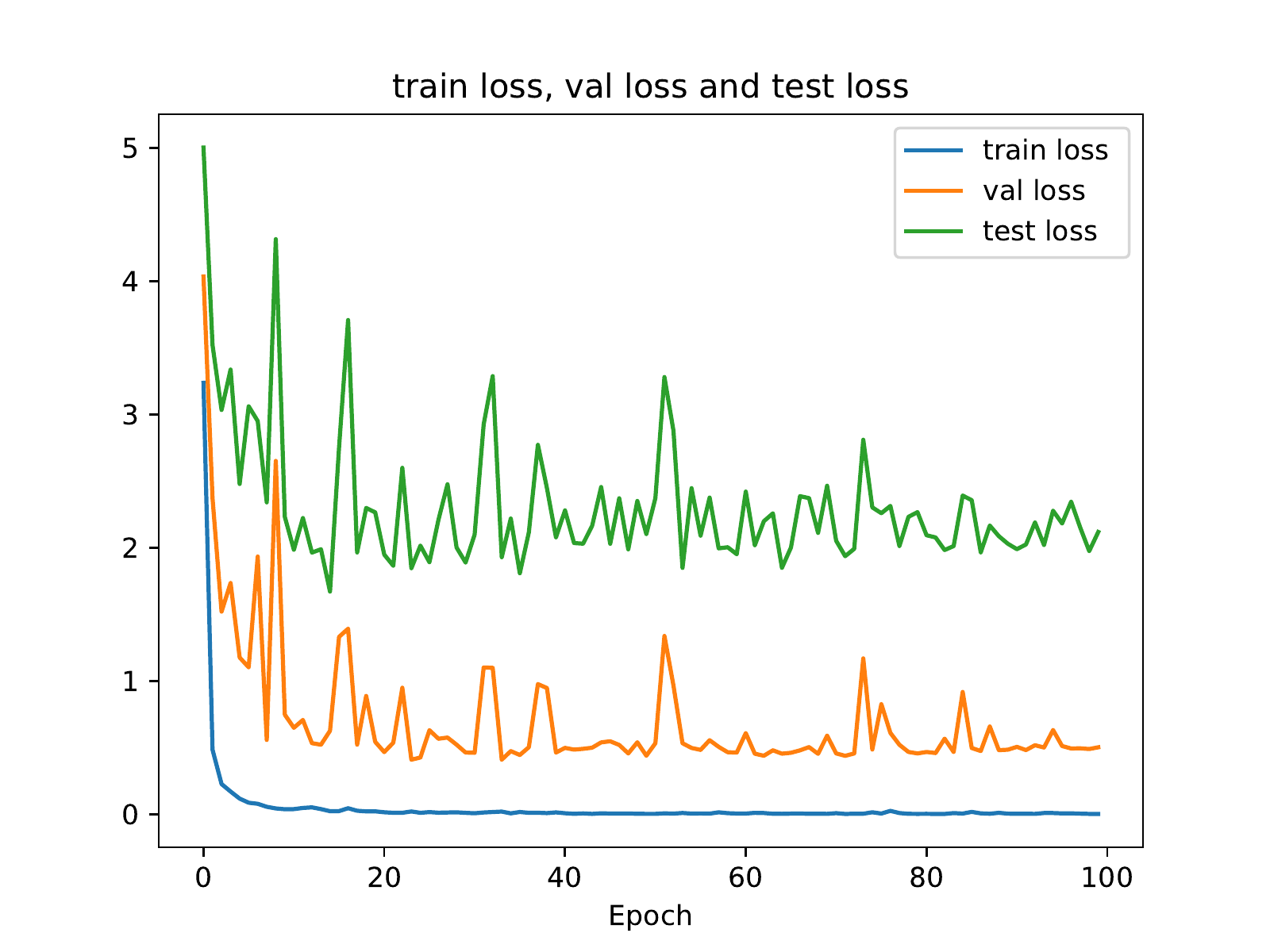}
        \caption[]%
        {{\small Loss evolution over 100 epochs on the \texttt{train}, \texttt{val} and \texttt{test} datasets for ResNet-50 multi-level classifier with resampler (cw: \ding{55}, rs: \ding{51}).}}    
        \label{fig:loss multi_level rs}
    \end{subfigure}
    \vskip\baselineskip
    \begin{subfigure}[b]{0.482\textwidth}   
        \centering 
        \includegraphics[width=\textwidth]{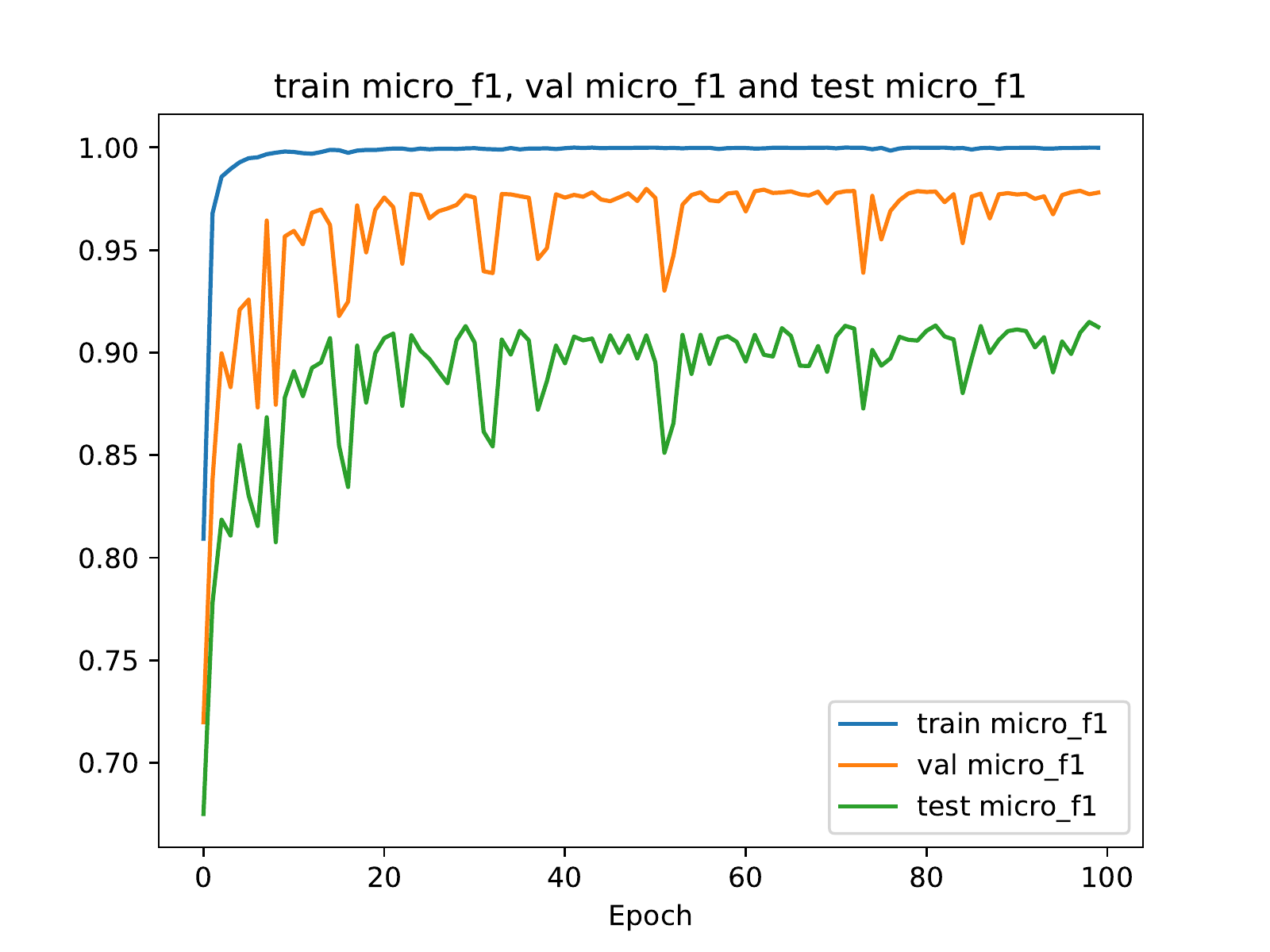}
        \caption[]%
        {{\small \texttt{train}, \texttt{val} and \texttt{test} micro-F1 performance over 100 epochs for ResNet-50 multi-level classifier with resampler (cw: \ding{55}, rs: \ding{51}).}}    
        \label{fig:micro_f1 multi_level rs}
    \end{subfigure}
    \quad
    \begin{subfigure}[b]{0.482\textwidth}   
        \centering 
        \includegraphics[width=\textwidth]{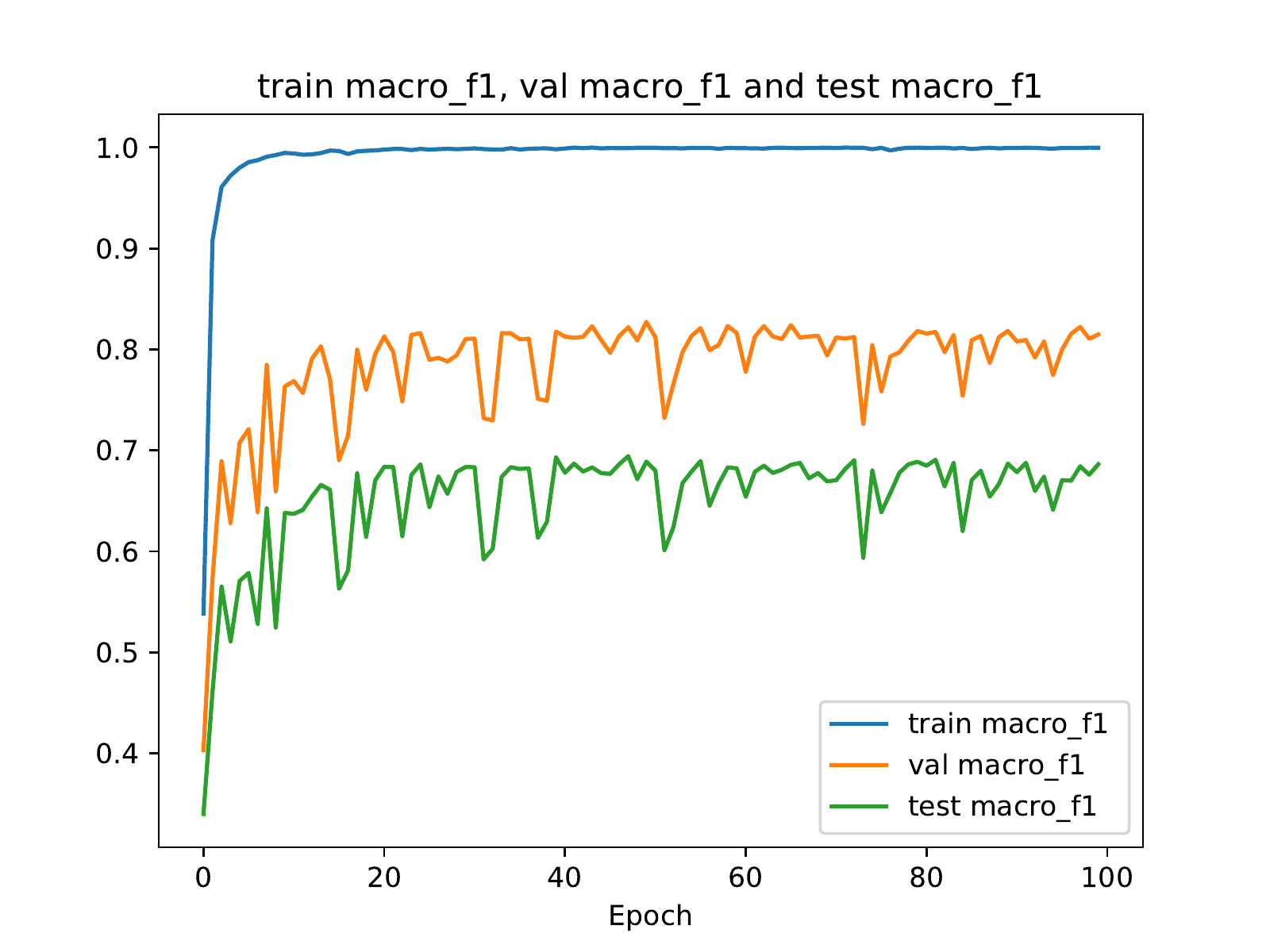}
        \caption[]%
        {{\small \texttt{train}, \texttt{val} and \texttt{test} macro-F1 performance over 100 epochs for ResNet-50 multi-level classifier with resampler (cw: \ding{55}, rs: \ding{51}).}}    
        \label{fig:macro_f1 multi_level rs}
    \end{subfigure}
    \caption[]
    {\small \centering Comparing micro-F1 performance across different combinations of class weights and resampler for the multi-level classifier. Refer to \cref{table:ethec_merged_ft_multi_level} for best scores. Loss, micro-F1 and macro-F1 for ResNet-50 with resampler (cw: \ding{55}, rs: \ding{51}) across 100 epochs.} 
    \label{fig:multi_level_4x cw rs combinations}
\end{figure*}

Refer to \cref{fig:macro_f1 multi_level rs} for the macro-F1, \cref{fig:micro_f1 multi_level rs} for the micro-F1 score and \cref{fig:loss multi_level rs} for the training, validation and testing loss over 100 epochs. The micro-F1 performance for experiments with different combinations of class weights and re-sampling are compared in \cref{fig:micro_f1 rs cw combination combined plot}.

\subsubsection{\texttt{sqrt}: square root resampler}
In order to magnify the extremely under-represented classes, instead of using the inverse of the frequency for a particular label we use the square-root of the inverse of the frequency to perform re-sampling.

The use of an inverse square-root is to perform an even more aggressive re-sampling for classes with only a handful of data points. However, we notice empirically that this does not make a significant difference in the model's performance.

\subsubsection{Performance breakdown}

In order to see where the model under-performs we break down the best performing model's score such that we have metrics across different hierarchical levels. From \cref{table:ethec_merged_ft_multi_level_level_split} it is evident that the performance is much better when there are less classes since classes higher up the hierarchy agglomerate the descendant nodes together, ending up with more data and less labels to differentiate between. 

\begin{figure*}[!htbp]
    \centering
    \begin{subfigure}[b]{0.480\textwidth}
        \centering
        \includegraphics[width=\textwidth]{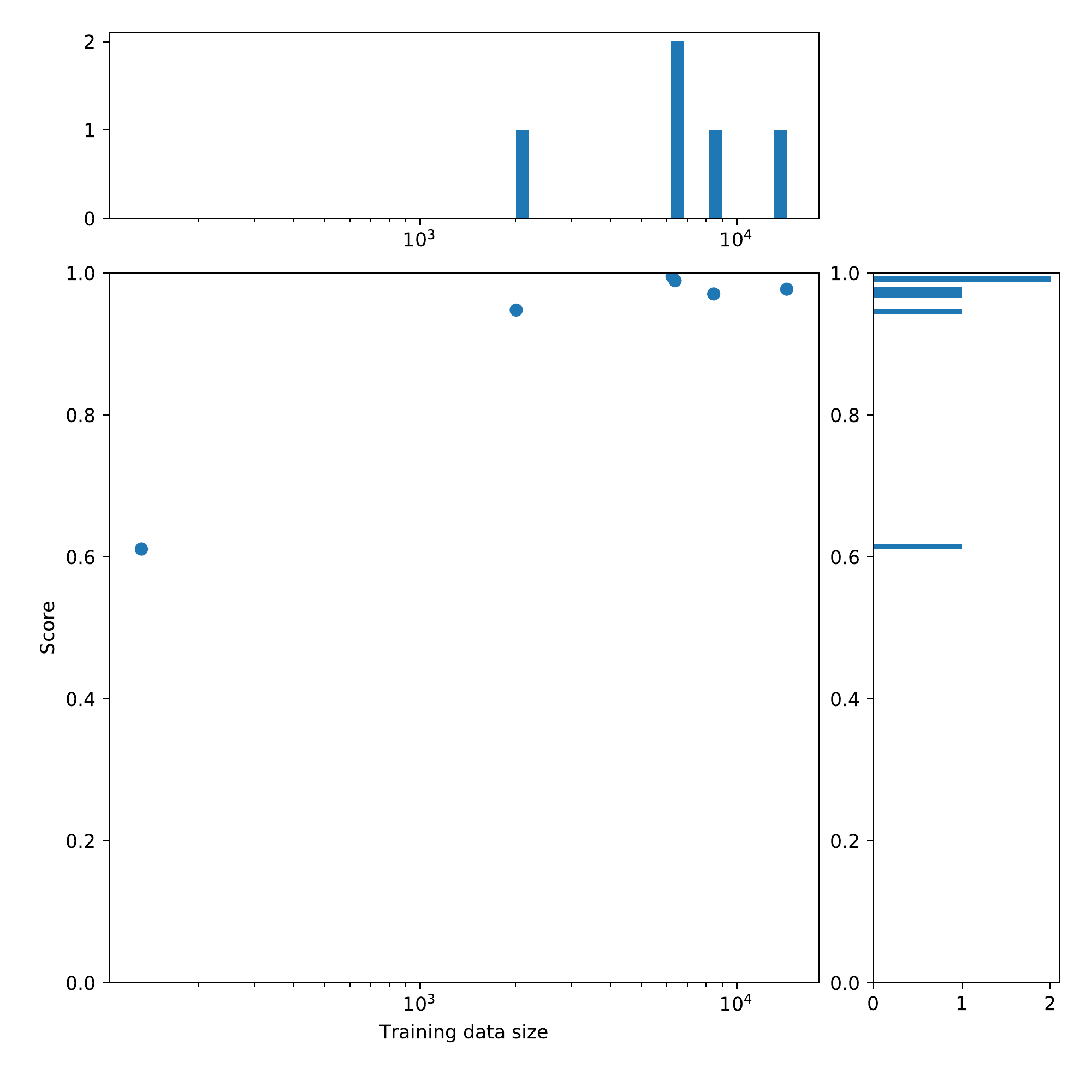}
        \caption[Network2]%
        {{\small \textit{family}}}    
        \label{fig:family multi_level}
    \end{subfigure}
    \hfill
    \begin{subfigure}[b]{0.480\textwidth}  
        \centering 
        \includegraphics[width=\textwidth]{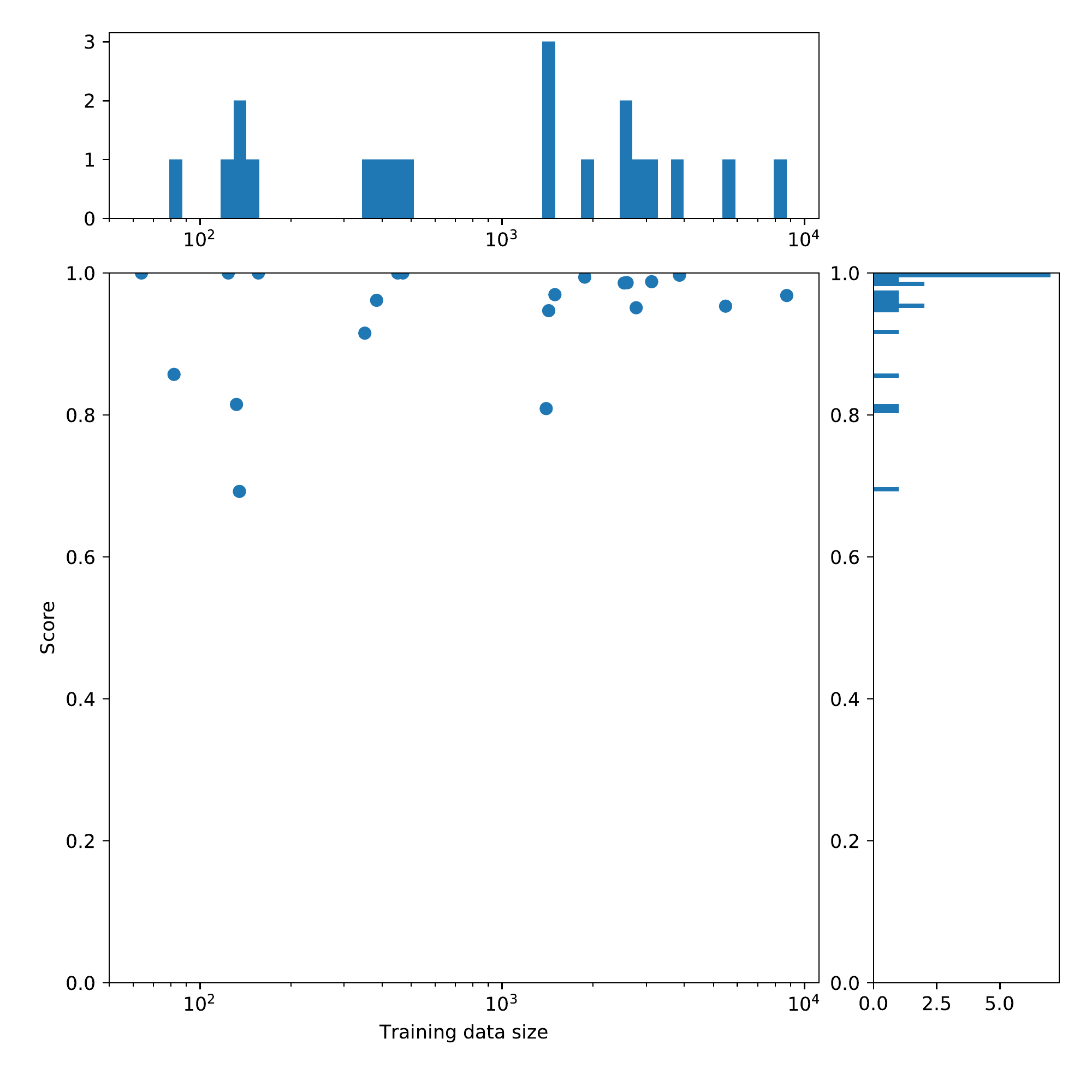}
        \caption[]%
        {{\small \textit{subfamily}}}    
        \label{fig:subfamily multi_level}
    \end{subfigure}
    \vskip\baselineskip
    \begin{subfigure}[b]{0.480\textwidth}   
        \centering 
        \includegraphics[width=\textwidth]{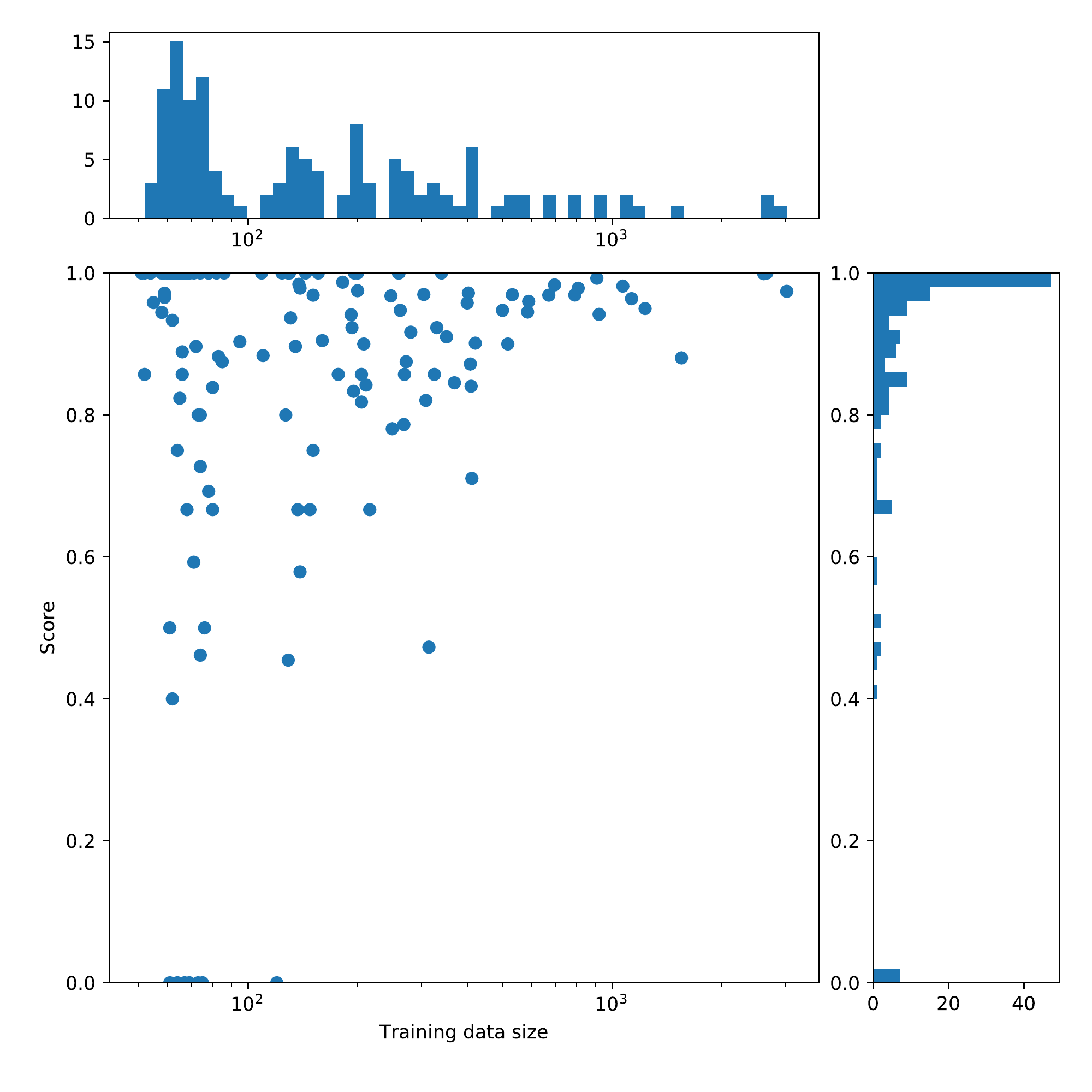}
        \caption[]%
        {{\small \textit{genus}}}    
        \label{fig:genus multi_level}
    \end{subfigure}
    \quad
    \begin{subfigure}[b]{0.480\textwidth}   
        \centering 
        \includegraphics[width=\textwidth]{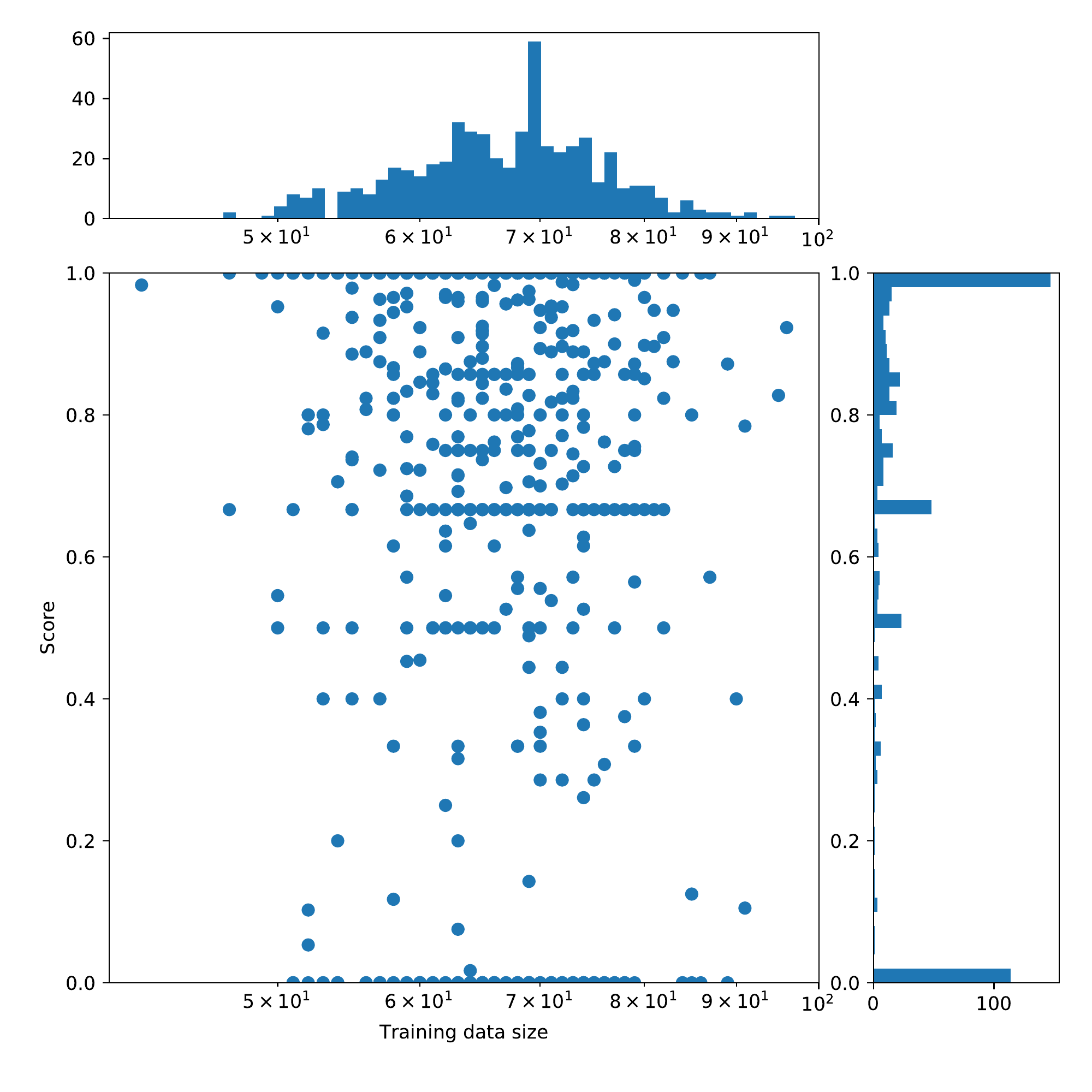}
        \caption[]%
        {{\small \textit{genus + specific epithet}}}    
        \label{fig:genus + specific epithet multi_level}
    \end{subfigure}
    \caption[]
    {\small \centering Per-label F1 performance across 4 hierarchical levels plotted against number of training samples for Per-level classifier using ResNet-50 with resampler (cw: \ding{55}, rs: \ding{51}). It is important to note that the population statistics, especially of lower levels in the hierarchy (\cref{fig:genus multi_level} and \cref{fig:genus + specific epithet multi_level}) are skewed to the higher end as an effect of re-sampling the less frequent classes.} 
    \label{fig:multi_level_rs_4x}
\end{figure*}

\cref{fig:multi_level_rs_4x} gives more insight into the data distribution for each level and the corresponding performance measured by the F1 score. The performance of the model deteriorates as one moves to the lower levels in the hierarchy. At the leaves the performance is the worst among the four levels due to extensive branching in the hierarchy and only several data samples per leaf label. With the help of inverse frequency weighted re-sampling the ill-effect of data deficiency is mitigated to an extent with an improved performance as compared to when there is no re-sampling at all (refer to \cref{table:ethec_merged_ft_multi_level}). 

\begin{table}[!htbp]
\centering
\begin{tabular}{| c | c || c | c | c || c | c | c |} 
 \hline
  Level & $N_{i}$ & m-P & m-R & m-F1 & M-P & M-R & M-F1  \\ [0.5ex] 
 \hline\hline
 \multicolumn{8}{|c|}{ResNet-50 with resampler (cw: \ding{55}, rs: \ding{51})} \\
 \hline\hline
  \textit{family} & 6 & 0.9766 & 0.9766 & 0.9766 & 0.9005 & 0.9328 & 0.9152 \\
  \textit{subfamily} & 21 & 0.9661 & 0.9661 & 0.9661 & 0.9433 & 0.9542 & 0.9424 \\
  \textit{genus} & 135 & 0.9204 & 0.9204 & 0.9204 & 0.8845 & 0.8482 & 0.8497 \\
  \textit{genus} + \textit{specific epithet} & 561 & 0.7704 & 0.7704 & 0.7704 & 0.6616 & 0.6811 & 0.6382  \\
 [1ex] \hline
\end{tabular}
\caption{Performance metrics for Per-level classifier on the ETHEC Merged dataset when using a resampler which is the best performing model in its category. The models used in this experiment are pre-trained on the 1000-class ImageNet data set. All weights are updated with a learning rate of 0.01, a batch-size of 64 and input spatial dimensions are 224x224 for 100 epochs. \textit{P}, \textit{R} and \textit{F1} represent Precision, Recall and F1-score. Metrics prefixed with \textit{m} are micro-averaged while the ones with \textit{M} are macro-averaged.}
\label{table:ethec_merged_ft_multi_level_level_split}
\end{table}

\clearpage

\subsection{Marginalization}

\begin{table}[!htbp]
\centering
\begin{tabular}{| c || c | c | c || c | c | c |} 
 \hline
  model & m-P & m-R & m-F1 & M-P & M-R & M-F1  \\ [0.5ex] \hline\hline
  \multicolumn{7}{|c|}{Models trained using grayscale images} \\
  \hline\hline
  ResNet-50 & 0.8586 & 0.8586 & 0.8586 & 0.6071 & 0.6070 & 0.5765 \\
  \hline\hline
  \multicolumn{7}{|c|}{Models trained using normal color images} \\
 \hline\hline
  ResNet-50 & \textbf{0.9223} & \textbf{0.9223} & \textbf{0.9223} & \textbf{0.7095} & \textbf{0.7231} & \textbf{0.6927} \\  
  ResNet-101 & 0.9110 & 0.9110 & 0.9110 & 0.7327 & 0.7262 & 0.7023 \\
  ResNet-152 & 0.9162 & 0.9162 & 0.9162 & 0.7181 & 0.7271 & 0.6954 \\ [1ex]
  \hline
\end{tabular}
\caption{Performance metrics for Marginalization classifiers on the ETHEC Merged dataset. The models used in this experiment are pre-trained on the 1000-class ImageNet data set. All weights are updated with a learning rate of $10^{-5}$, a batch-size of 64 and input spatial dimensions are 224x224 for 100 epochs. \textit{P}, \textit{R} and \textit{F1} represent Precision, Recall and F1-score. For all models, data re-sampling proportional to the inverse of the class frequency is performed while training. Metrics prefixed with \textit{m} are micro-averaged while the ones with \textit{M} are macro-averaged. The top performing models are in bold-face. All these models use level weights [1, 1, 1, 1].}
\label{table:ethec_baseline3_bottom_up}
\end{table}

\begin{table}[!htbp]
\centering
\begin{tabular}{| c | c | c | c || c || c | c | c | c |} 
 \hline
  $L_{1}$ & $L_{2}$ & $L_{3}$ & $L_{4}$ & m-F1 & m-F1 $L_{1}$ & m-F1 $L_{2}$ & m-F1 $L_{3}$ & m-F1 $L_{4}$  \\ [0.5ex] 
 \hline\hline
 \multicolumn{4}{|c||}{term $L_{i}$ in loss} &
 \multicolumn{1}{c||}{} &
 \multicolumn{4}{c|}{Per-level micro-F1} \\
 \hline\hline
  & & & \ding{51} & 0.9137 & 0.9814 & 0.9638 & 0.9134 & 0.7962 \\
  & & \ding{51} & \ding{51} & 0.9070 & 0.9774 & 0.9626 & 0.9077 & 0.7804 \\
  & \ding{51} & \ding{51} & \ding{51} & 0.9207 & 0.9891 & 0.9733 & 0.9255 & 0.7948 \\
  \ding{51} & \ding{51} & \ding{51} & \ding{51} & 0.9223 & 0.9887 & 0.9758 & 0.9273 & 0.7972 \\
 [1ex] \hline
\end{tabular}
\caption{We compare performance of Marginalization models when trained with and without losses from each hierarchical level. Computing the losses over more levels improves model performance. The models used in this experiment are pre-trained on the 1000-class ImageNet data set. All weights are updated with a learning rate of $10^{-5}$, a batch-size of 64 and input spatial dimensions are 224x224 for 100 epochs. \textit{P}, \textit{R} and \textit{F1} represent Precision, Recall and F1-score. Metrics prefixed with \textit{m} are micro-averaged while the ones with \textit{M} are macro-averaged.}
\label{table:ethec_baseline3_bottom_up_level_combination_loss}
\end{table}

This section compiles the general results for image classification using the marginalization model. The ResNet-50 is the best performing model. This model predicts the non-leaf labels in the hierarchy by marginalizing over children labels whose are probabilities explicitly predicted by the model. We also notice that a huge performance boost is obtained when normal color images are used as compared to grayscale images. It is not just about the patterns but also the colors on the specimen that help distinguish them.

We also show the best performing model's performance when different loss terms are used to compute the loss. We train models where we sum up classification losses across different levels in the hierarchy and observe that when losses are computed over more levels it yields better performance.


\subsection{Masked Per-level classifier}
This model predicts in a top-down manner where it assumes prediction made on the upper levels are more accurate than the ones made for lower levels in the label-hierarchy. A similar phenomenon is observed in a previous work \cite{tasho2018thesis}.

Here, the model masks out all immediate non-descendants of the previous label (which is a part of the level above the current in the hierarchy). This scheme is made to trickle down all the way down to the leaf nodes. When predicting the label for any given level, the model chooses the label with the best score among all direct descendants of the label predicted for the level above the current.

\begin{table}[!htbp]
\centering
\begin{tabular}{| c || c | c | c || c | c | c |} 
 \hline
  model & m-P & m-R & m-F1 & M-P & M-R & M-F1  \\ [0.5ex] \hline\hline
  \multicolumn{7}{|c|}{Models trained using grayscale images} \\
  \hline\hline
  ResNet-50 & 0.8443 & 0.8443 & 0.8443 & 0.6002 & 0.5931 & 0.5619 \\
  \hline\hline
  \multicolumn{7}{|c|}{Models trained using normal color images} \\
  \hline\hline
  ResNet-50 & \textbf{0.9173} & \textbf{0.9173} & \textbf{0.9173} &  0.7107 & 0.7227 & 0.6915 \\
  ResNet-101 & 0.9169 & 0.9169 & 0.9169 &  0.7119 & 0.7260 & 0.6921 \\
  ResNet-152 & 0.9152 & 0.9152 & 0.9152 & \textbf{0.7167} & \textbf{0.7281} & \textbf{0.6958} \\ [1ex]
  \hline
\end{tabular}
\caption{Performance metrics for Masked Per-level classifier on the ETHEC Merged dataset. The models used in this experiment are pre-trained on the 1000-class ImageNet data set. All weights are updated with a learning rate of $10^{-5}$, a batch-size of 64 and input spatial dimensions are 224x224 for 200 epochs. \textit{P}, \textit{R} and \textit{F1} represent Precision, Recall and F1-score. For all models, data re-sampling proportional to the inverse of the class frequency is performed while training. Metrics prefixed with \textit{m} are micro-averaged while the ones with \textit{M} are macro-averaged. The top performing models are in bold-face. In addition to the normal experiments, we also include results from models trained on grayscale images.}
\label{table:ethec_baseline3_top_down}
\end{table}

This is the second best performing CNN-based model and we also look at the level-wise performance split in \cref{table:ethec_baseline3_top_down_level_split}.

\begin{table}[!htbp]
\centering
\begin{tabular}{| c | c || c | c | c || c | c | c |} 
 \hline
  Level & $N_{i}$ & m-P & m-R & m-F1 & M-P & M-R & M-F1  \\ [0.5ex] 
 \hline\hline
 \multicolumn{8}{|c|}{ResNet-50 Performance Breakdown} \\
 \hline\hline
  \textit{family} & 6 & 0.9828 & 0.9828 & 0.9828 & 0.9735 & 0.9361 & 0.9495 \\
  \textit{subfamily} & 21 & 0.9701 & 0.9701 & 0.9701 & 0.9684 & 0.9252 & 0.9356 \\
  \textit{genus} & 135 & 0.9233 & 0.9233 & 0.9233 & 0.8916 & 0.8432 & 0.8525 \\
  \textit{genus} + \textit{specific epithet} & 561 & 0.7930 & 0.7930 & 0.7930 & 0.6548 & 0.6838 & 0.6409  \\
 [1ex] \hline
\end{tabular}
\caption{Performance metrics for Masked Per-level classifier on the ETHEC Merged dataset for the best performing model. The models used in this experiment are pre-trained on the 1000-class ImageNet data set. All weights are updated with a learning rate of $10^{-5}$, a batch-size of 64 and input spatial dimensions are 224x224 for 200 epochs. \textit{P}, \textit{R} and \textit{F1} represent Precision, Recall and F1-score. Metrics prefixed with \textit{m} are micro-averaged while the ones with \textit{M} are macro-averaged.}
\label{table:ethec_baseline3_top_down_level_split}
\end{table}

\begin{table}[!htbp]
\centering
\begin{tabular}{| c | c | c | c || c || c | c | c | c |} 
 \hline
  $L_{1}$ & $L_{2}$ & $L_{3}$ & $L_{4}$ & m-F1 & m-F1 $L_{1}$ & m-F1 $L_{2}$ & m-F1 $L_{3}$ & m-F1 $L_{4}$  \\ [0.5ex] 
 \hline\hline
 \multicolumn{4}{|c||}{term $L_{i}$ in loss} &
 \multicolumn{1}{c||}{} &
 \multicolumn{4}{c|}{Per-level micro-F1} \\
 \hline\hline
  & & & \ding{51} & 0.0633 & 0.2325 & 0.0162 & 0.0022 & 0.0022 \\
  & & \ding{51} & \ding{51} & 0.1043 & 0.3058 & 0.0410 & 0.0386 & 0.0319 \\
  & \ding{51} & \ding{51} & \ding{51} & 0.0848 & 0.0970 & 0.0919 & 0.0879 & 0.0622 \\
  \ding{51} & \ding{51} & \ding{51} & \ding{51} & 0.9098 & 0.9808 & 0.9616 & 0.9116 & 0.7853 \\
 [1ex] \hline
\end{tabular}
\caption{We compare performance of Masked Per-level classifier when trained with and without losses from each hierarchical level. Computing the losses over more levels improves model performance. The models used in this experiment are pre-trained on the 1000-class ImageNet data set. All weights are updated with a learning rate of $10^{-5}$, a batch-size of 64 and input spatial dimensions are 224x224 for 100 epochs. \textit{P}, \textit{R} and \textit{F1} represent Precision, Recall and F1-score. Metrics prefixed with \textit{m} are micro-averaged while the ones with \textit{M} are macro-averaged.}
\label{table:ethec_baseline3_top_down_level_combination_loss}
\end{table}

\clearpage

\subsection{Hierarchical Softmax}

\begin{table}[!htbp]
\centering
\begin{tabular}{| c || c | c | c || c | c | c |} 
 \hline
  model & m-P & m-R & m-F1 & M-P & M-R & M-F1  \\
  \hline\hline
  ResNet-50 & 0.9055 & 0.9055 & 0.9055 & 0.6899 & 0.7049 & 0.6723 \\
  ResNet-101 & 0.9122 & 0.9122 & 0.9122 & 0.7049 & 0.7072 & 0.6780 \\
  ResNet-152 & \textbf{0.9180} & \textbf{0.9180} & \textbf{0.9180} & \textbf{0.7119} & \textbf{0.7174} & \textbf{0.6869} \\ [1ex]
  \hline
\end{tabular}
\caption{Performance metrics for Hierarchical Softmax on the ETHEC Merged dataset. The models used in this experiment are pre-trained on the 1000-class ImageNet data set. All weights are updated with a learning rate of $10^{-5}$, a batch-size of 64 and input spatial dimensions are 224x224 for 100 epochs. \textit{P}, \textit{R} and \textit{F1} represent Precision, Recall and F1-score. For all models, data re-sampling proportional to the inverse of the class frequency is performed while training. Metrics prefixed with \textit{m} are micro-averaged while the ones with \textit{M} are macro-averaged. The top performing models are in bold-face.}
\label{table:ethec_hierarchical_softmax}
\end{table}

The Hierarchical Softmax seems to be less prone to over-fitting and has the best performing model with the ResNet-152 backbone. To recall, the model predicts conditional distribution $p(\text{child}|\text{parent})$ for each label in the hierarchy. The joint distribution is calculated by multiplying probabilities of all labels along a specific path to obtain the probability for the leaf.

\begin{table}[!htbp]
\centering
\begin{tabular}{| c | c || c | c | c || c | c | c |} 
 \hline
  Level & $N_{i}$ & m-P & m-R & m-F1 & M-P & M-R & M-F1  \\ [0.5ex] 
 \hline\hline
 \multicolumn{8}{|c|}{ResNet-152 with Hierarchical Softmax | Performance Breakdown} \\
 \hline\hline
  \textit{family} & 6 & 0.9879 & 0.9879 & 0.9879 & 0.9605 & 0.9452 & 0.9522 \\
  \textit{subfamily} & 21 & 0.9731 & 0.9731 & 0.9731 & 0.9605 & 0.9452 & 0.9522 \\
  \textit{genus} & 135 & 0.9253 & 0.9253 & 0.9253 & 0.8972 & 0.8504 & 0.8574 \\
  \textit{genus} + \textit{specific epithet} & 561 & 0.7855 & 0.7855 & 0.7855 & 0.6572 & 0.6756 & 0.6347  \\
 [1ex] \hline
\end{tabular}
\caption{Performance metrics for Hierarchical Softmax on the ETHEC Merged dataset for the best performing model. The models used in this experiment are pre-trained on the 1000-class ImageNet data set. All weights are updated with a learning rate of $10^{-5}$, a batch-size of 64 and input spatial dimensions are 224x224 for 200 epochs. \textit{P}, \textit{R} and \textit{F1} represent Precision, Recall and F1-score. Metrics prefixed with \textit{m} are micro-averaged while the ones with \textit{M} are macro-averaged.}
\label{table:ethec_hsoftmax_level_split}
\end{table}

\begin{figure*}[!htbp]
    \centering
    \begin{subfigure}[b]{0.480\textwidth}
        \centering
        \includegraphics[width=\textwidth]{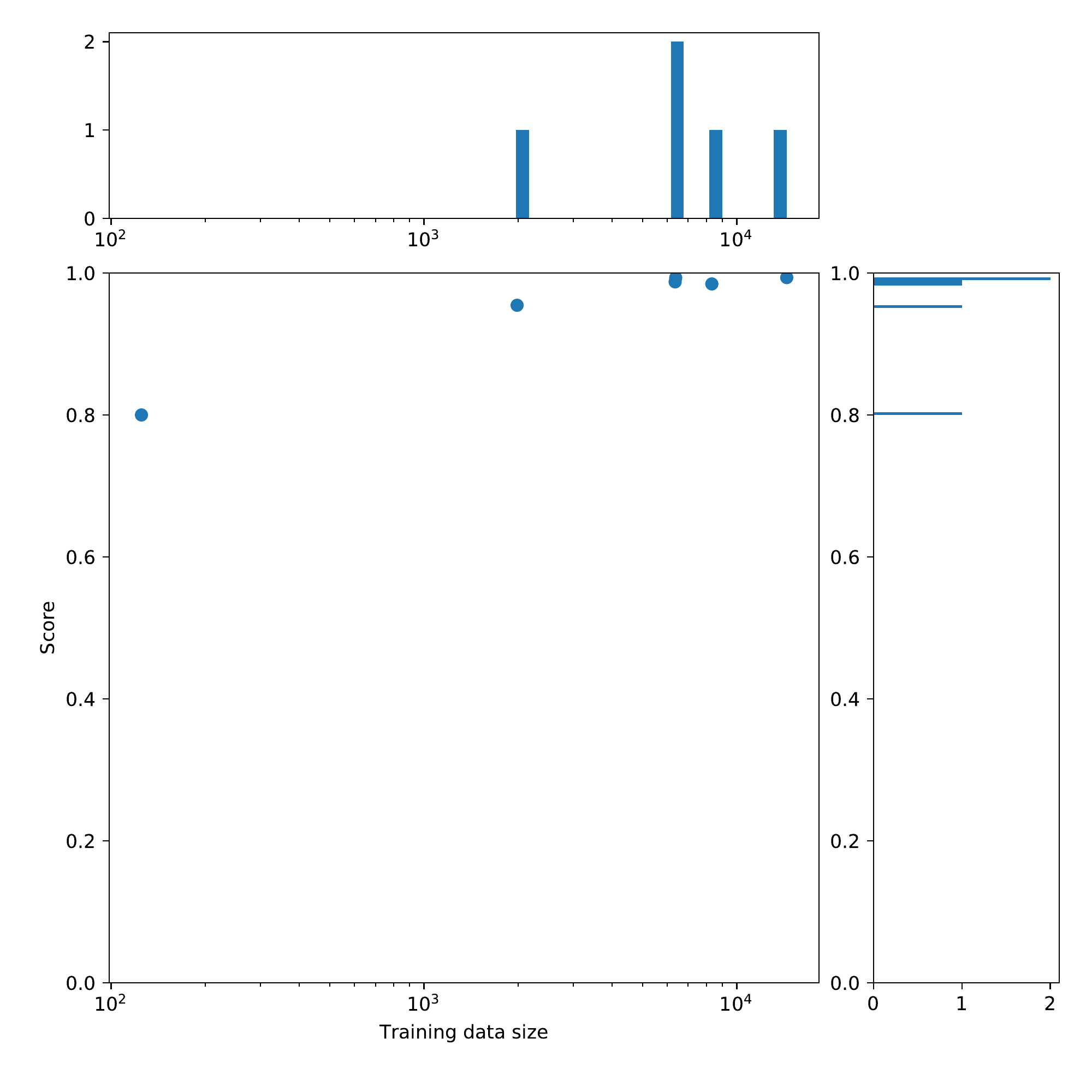}
        \caption[Network2]%
        {{\small \textit{family}}}    
        \label{fig:family hsoftmax}
    \end{subfigure}
    \hfill
    \begin{subfigure}[b]{0.480\textwidth}  
        \centering 
        \includegraphics[width=\textwidth]{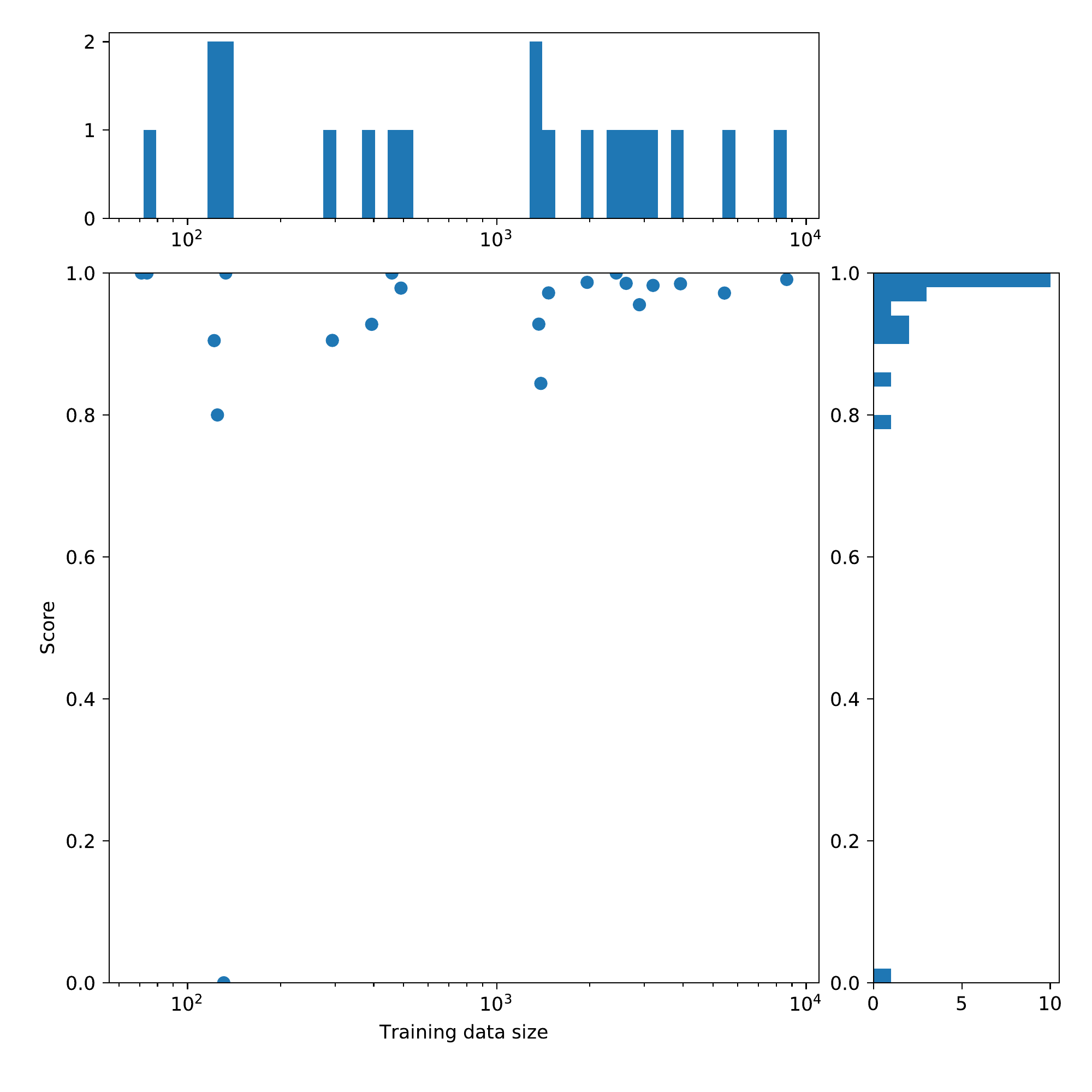}
        \caption[]%
        {{\small \textit{subfamily}}}    
        \label{fig:subfamily hsoftmax}
    \end{subfigure}
    \vskip\baselineskip
    \begin{subfigure}[b]{0.480\textwidth}   
        \centering 
        \includegraphics[width=\textwidth]{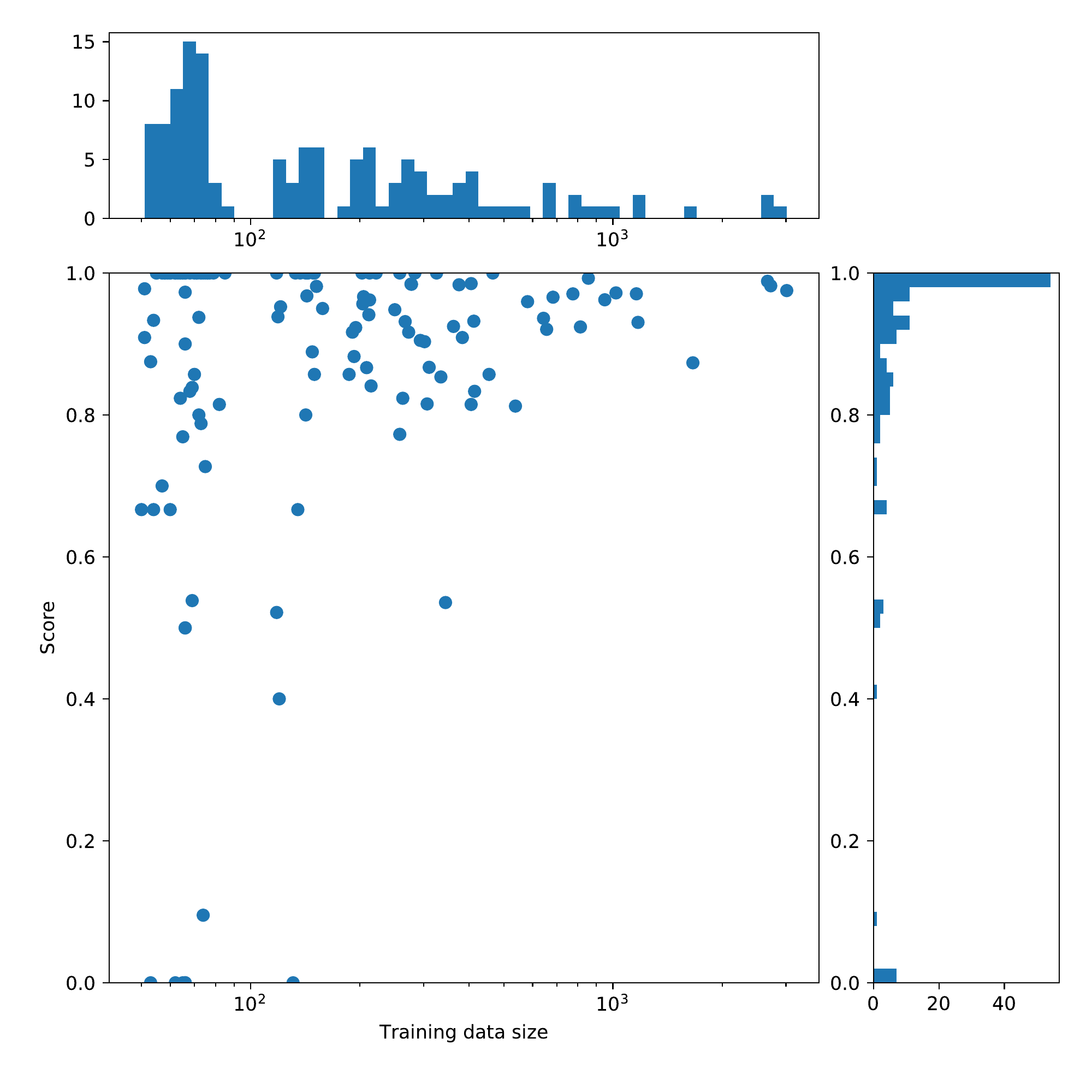}
        \caption[]%
        {{\small \textit{genus}}}    
        \label{fig:genus hsoftmax}
    \end{subfigure}
    \quad
    \begin{subfigure}[b]{0.480\textwidth}   
        \centering 
        \includegraphics[width=\textwidth]{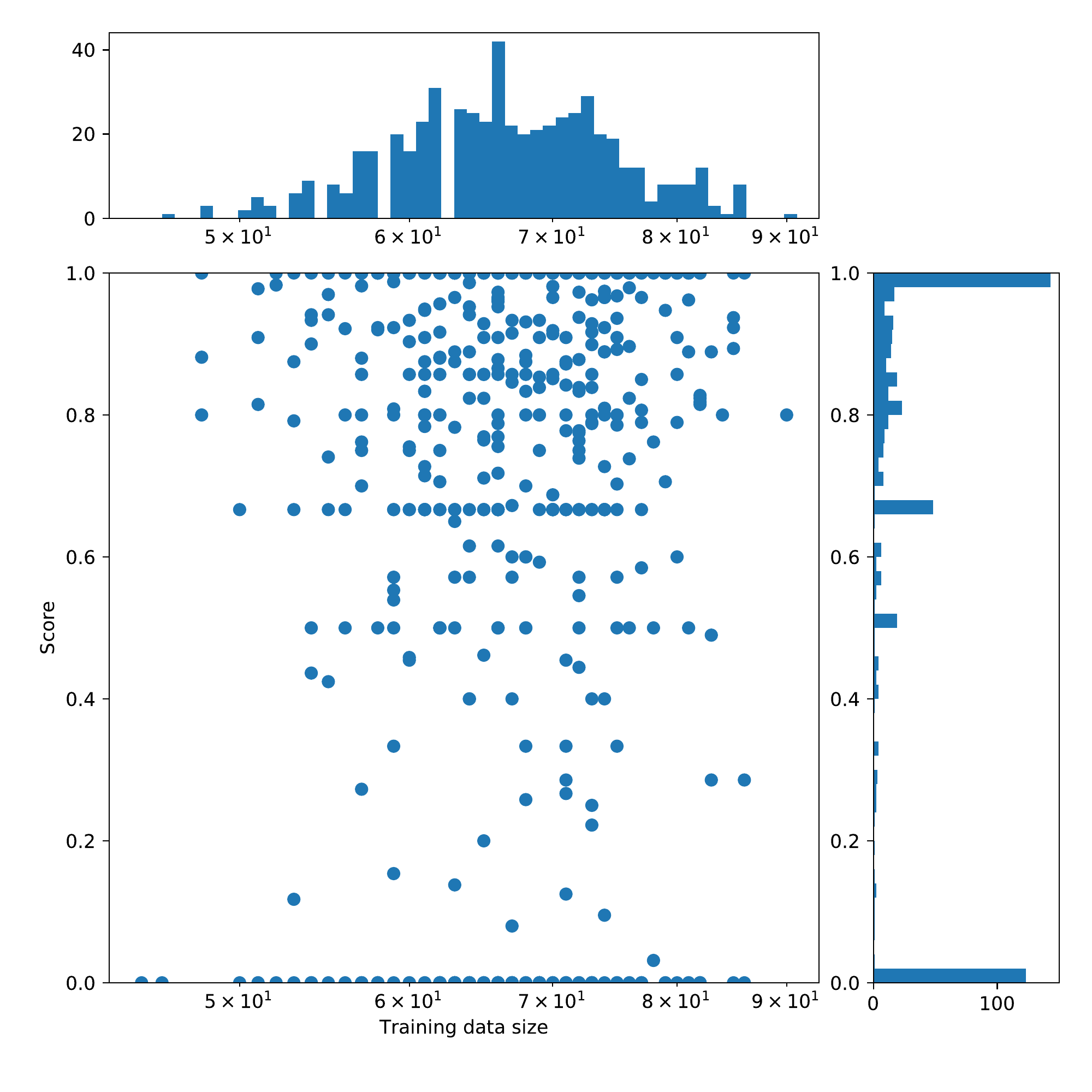}
        \caption[]%
        {{\small \textit{genus + specific epithet}}}    
        \label{fig:genus + specific epithet hsoftmax}
    \end{subfigure}
    \caption[]
    {\small \centering Per-label F1 performance across 4 hierarchical levels plotted against number of training samples for ResNet-152 with Hierarchical Softmax and resampler (cw: \ding{55}, rs: \ding{51}). It is important to note that the population statistics, especially of lower levels in the hierarchy (\cref{fig:genus multi_level} and \cref{fig:genus + specific epithet multi_level}) are skewed to the higher end as an effect of re-sampling the less frequent classes.} 
    \label{fig:hierarchical_softmax_4x}
\end{figure*}

\clearpage

\subsection{Results Summary}

\begin{table}[!htbp]
\centering
\begin{tabular}{| c | c || c | c | c || c | c | c |} 
 \hline
  baseline & model & m-P & m-R & m-F1 & M-P & M-R & M-F1  \\ [0.5ex] 
 \hline\hline
  hierarchy-agnostic classifier & ResNet-50 & 0.9324 & 0.7235 & 0.8147 & 0.1913 & 0.1462 & 0.1568 \\
  Per-level classifier & ResNet-50 & 0.9084 & 0.9084 & 0.9084 & \textbf{0.7134} & 0.7223 & 0.6888 \\
  Marginalization & ResNet-50 & \textbf{0.9223} & \textbf{0.9223} & \textbf{0.9223} & 0.7095 & \textbf{0.7231} & \textbf{0.6927} \\
  Masked Per-level classifier & ResNet-50 & 0.9173 & 0.9173 & 0.9173 &  0.7107 & 0.7227 & 0.6915 \\
  Hierarchical softmax & ResNet-152 & 0.9180 & 0.9180 & 0.9180 & 0.7119 & 0.7174 & 0.6869
  \\ [1ex]
  \hline
\end{tabular}
\caption{Comparing best performing baseline classifiers on the ETHEC Merged dataset. The models used in this experiment are pre-trained on the 1000-class ImageNet data set. \textit{P}, \textit{R} and \textit{F1} represent Precision, Recall and F1-score. For details regarding the specific baselines please refer to the respective sections. Metrics prefixed with \textit{m} are micro-averaged while the ones with \textit{M} are macro-averaged. The top performing models are in bold-face.}
\label{table:ethec_baseline_summary}
\end{table}


\chapter{Methods: Order-preserving embedding-based models}
\label{ch:emb_models}

\section{Cosine Embeddings}

\begin{figure}[htbp]
    \centering
    \includegraphics[width=0.4\textwidth]{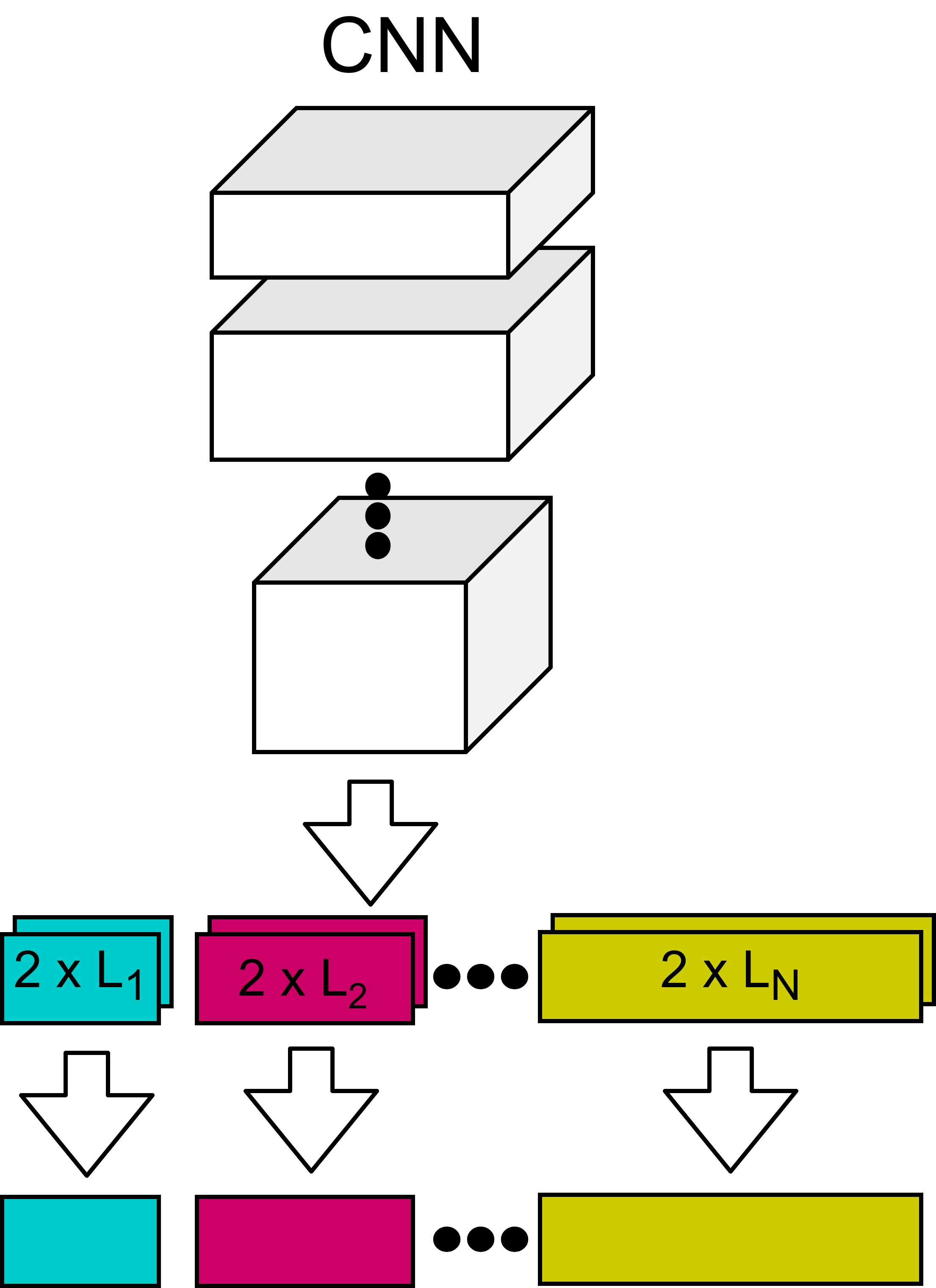}
    \caption{The latent space of the modified the per-level model is used to extract label embeddings. An additional layer is added after the final layer in the original model. The model is trained exactly like the L $N_{i}$-way classifier. The weights of the layer labeled $N \times L_{i}$ hold the N-dimensional representations of the labels Here, N=2, however it can be extended to any N-dimensional embedding space.}
    \label{fig:cosine-emb-model-schematic}
\end{figure}

For cosine embeddings we extract label representations from the latent space of a CNN trained for image classification. The learned representations are a by-product of the model being explicitly trained only for image classification. It is important to note that cosine embeddings are not necessarily order-preserving but are presented in this chapter with all the other embedding based models.

We modify the Per-level classifier model by adding a linear layer that projects the final fully-connected layer of the original model to a latent space which is interpreted as the label embedding.

The additional layer projects onto the N-dimensional embedding space for every label. In the \cref{fig:cosine-emb-model-schematic}, the weights of the layer labeled $2 \times L_{i}$ holds the 2-dimensional representation, one for each label for the \textit{i}-th level.

When performing image classification, matrix multiplication of layer weights and image representation from the upper layer yield the logits for each label.
The weights in the last layer represent the label embeddings. The label logits that represent the similarity between an image and the labels are computed by the dot-product between the image's representation and the representation of each label. The larger the magnitude of the dot-product between the representations of an image and a particular label, the larger the corresponding logit. A larger logit implies the likelihood of the image belonging to the particular label is high.

\begin{figure}[htbp]
    \centering
    \includegraphics[width=0.8\textwidth]{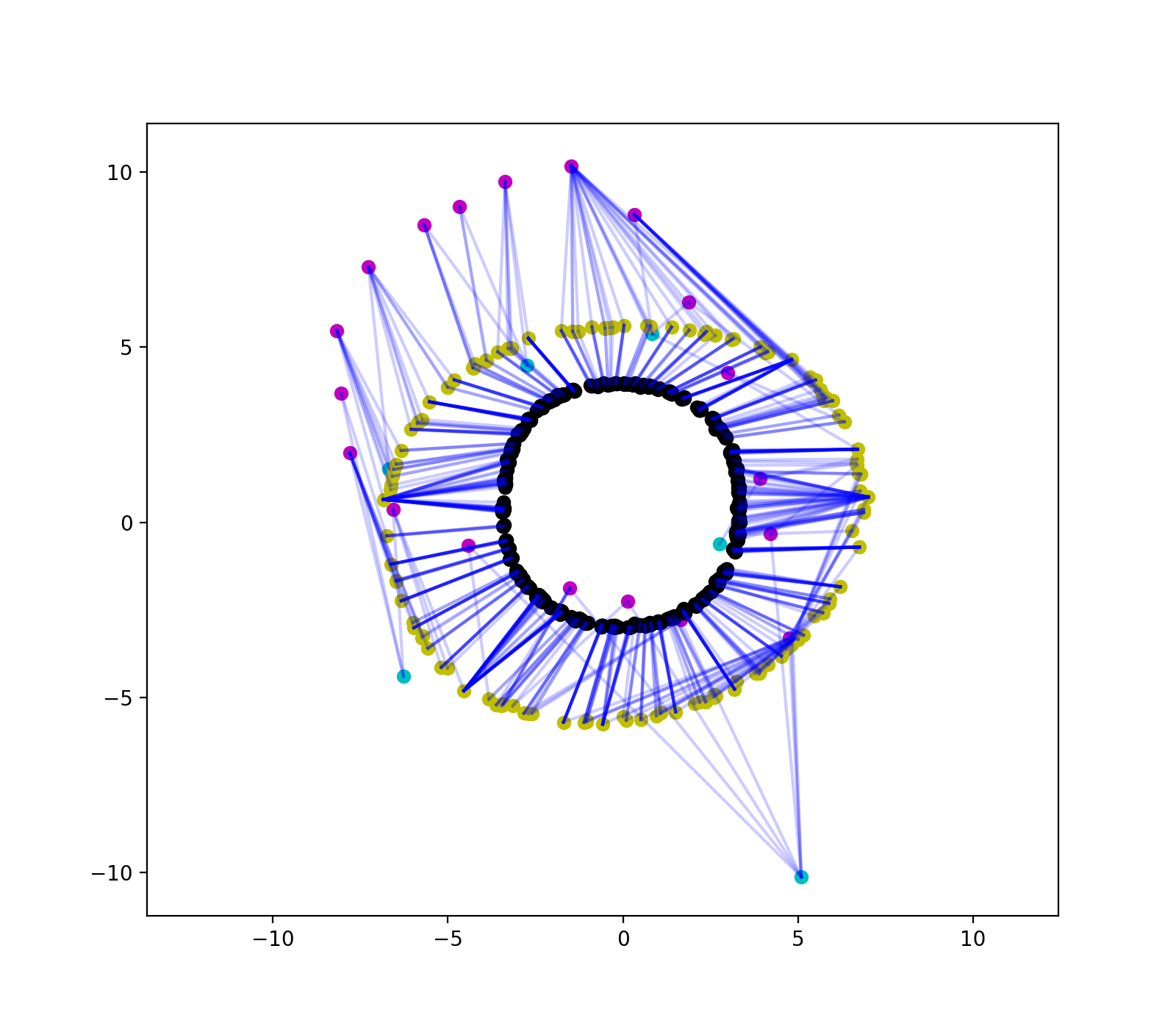}
    \caption{Visualizing the latent 2D space of the modified per-level model for the ETHEC dataset \cite{dhall_20.500.11850/365379}. Each label's embedding is plotted. The weights of the layer labeled $2 \times L_{i}$ hold the 2-dimensional representations of the labels. We visualize the relations between nodes by adding edges from the original label hierarchy.  Legend=\texttt{cyan}: family, \texttt{magenta}: subfamily, \texttt{yellow}: genus, \texttt{black}: genus+species. We see that for labels from a given level, the model is more confident about those that have a larger norm in the embedding space. For the lower levels (genus, genus+species) the labels form roughly a circular pattern meaning that the model has the same confident across the labels. We also see that \texttt{cyan} nodes are collapsed towards the center even though they have most samples per label (as they are the top-most level in the hierarchy) however since they capture images with a large intra-label variance they have a smaller norm than the \texttt{magenta} nodes.}
    \label{fig:cosine-emb-plot}
\end{figure}

\section{Order-Embeddings}
In this part we introduce learning representations for both concepts and images via embeddings. Recent advances show how unconventional loss functions (instead of the widely used vanilla inner-product or their p-norm distance) can model the asymmetric relations between concepts. Directed graphs also model asymmetric relations between two nodes connected by a directed edge.

We treat our label hierarchy as a directed-acyclic graph. More specifically, due to the nature of defining the taxonomy and how we create the dataset, it is a directed tree graph. The dataset $\mathcal{X}$ consists of entailment relations $(u, v)$ connected via a directed edge from $u$ to $v$. (following the definition in \cite{ganea2018entailment_cones}). These directed edges or hypernym links convey that $v$ is a sub-concept of $u$.

We train our model using the max-margin loss proposed in \cite{vendrov2015order}. Unlike in the work \cite{vendrov2015order} we do not restrict the embeddings to have positive coordinates only and get rid of the absolute function used by \cite{vendrov2015order}, granting the embeddings more freedom.

\begin{figure}[htbp]
    \centering
    \includegraphics[width=0.4\textwidth]{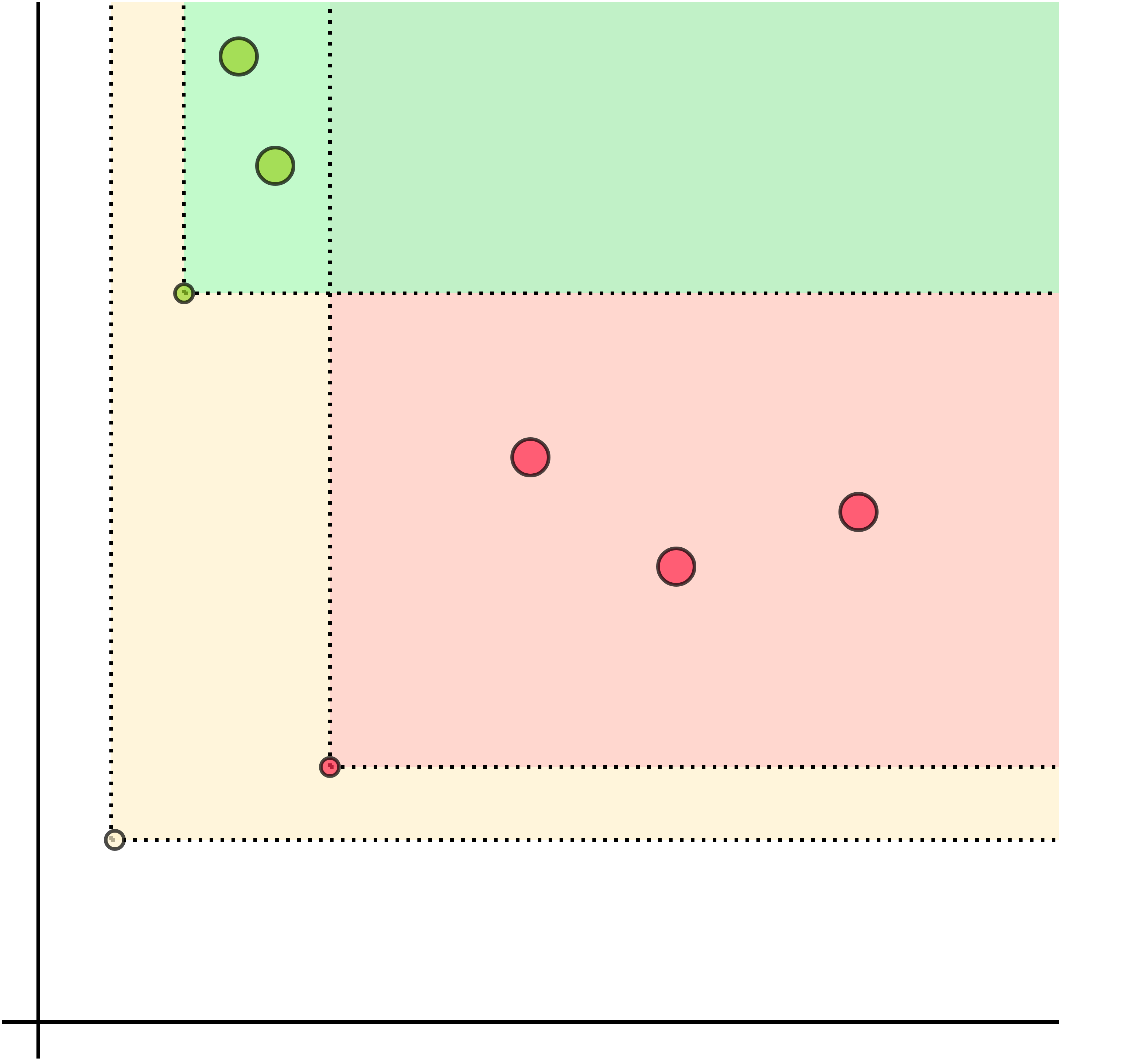}
    \caption{An illustration of the order-embedding space. In an ideal scenario, if $(u, v)$ is a positive edge i.e. $v$ is a sub-concept of $u$, then, all such sub-concepts lie within a translated orthant in the embedding space (a quadrant in $\mathbb{R}^{2}$) originating at $u$'s embedding. Both the immediate and non-immediate descendants (of a parent concept) should lie within the translated orthant that belongs to the parent concept.}
    \label{fig:order-emb-cartoon}
\end{figure}

\section{Euclidean Cones}
Euclidean cones is a generalization of the order-embeddings method. If $v$ \texttt{is-a} $u$ then in an ideal order-embeddings $f(v)$ lies within that orthant that has its origin at $f(u)$. Euclidean cones is a generalization of order-embeddings and allows for more flexible volumes in $\mathbb{R}^N$ to define \texttt{is-a} relations.

\begin{figure}[htbp]
    \centering
    \includegraphics[width=0.4\textwidth]{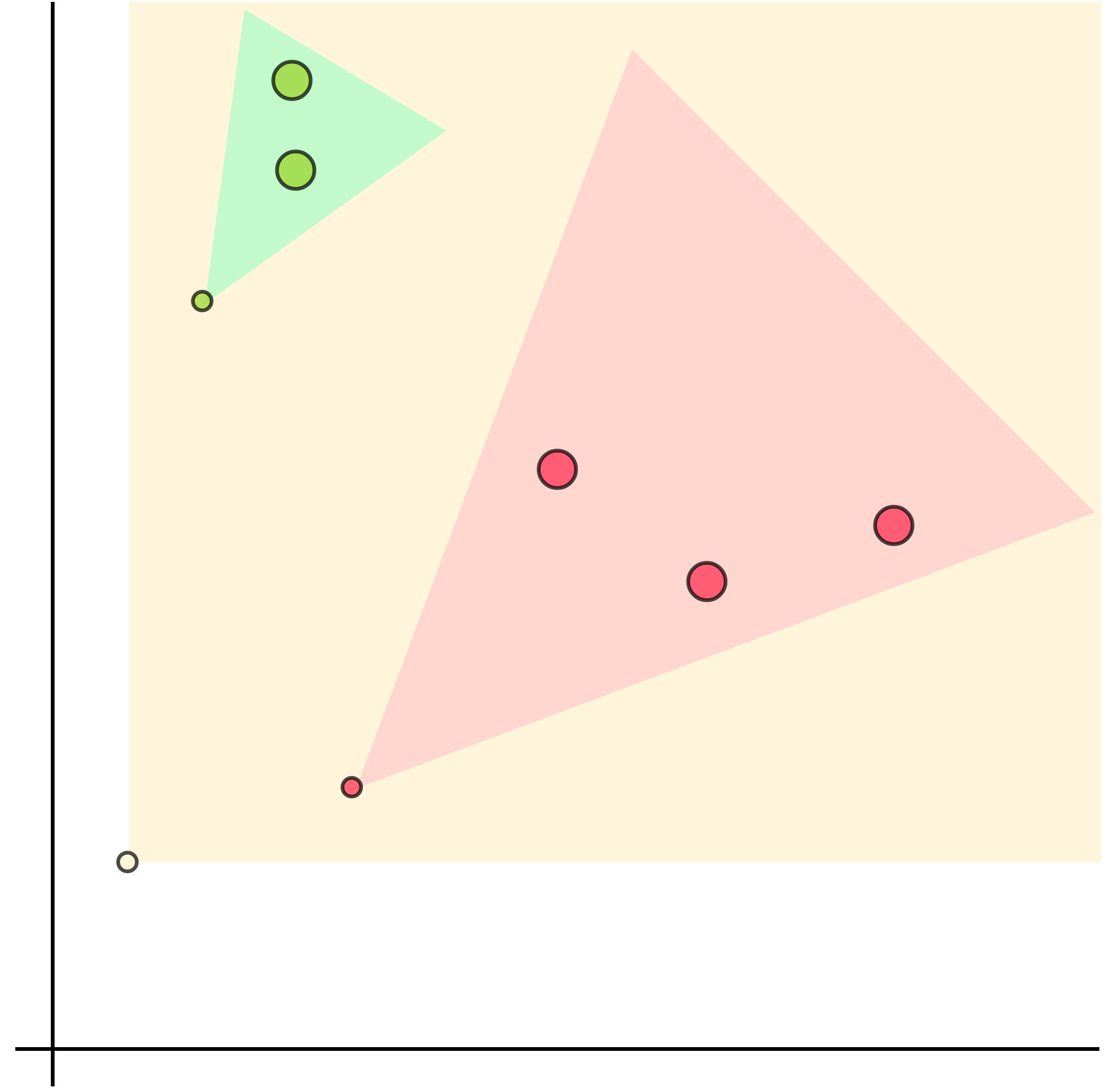}
    \caption{An illustration of euclidean cones. In an ideal scenario, if $(u, v)$ is a positive edge i.e. $v$ is a sub-concept of $u$, then, all such sub-concepts lie within a cone in the embedding space originating at $u$'s location in the embedding space. Both the immediate and non-immediate descendants (of a parent concept) should lie within the cone that belongs to the parent concept. Euclidean Cones is a generalization of order-embeddings. Like the orthants in order-embeddings the cones here extend to infinitely.}
    \label{fig:euc-cones-cartoon}
\end{figure}

A concept that is more abstract than others and consequently entails a lot more concepts can have more volume than the concepts that are entailed by it.

The general form of the loss is the same as in \cref{eq:order_embeddings_loss}. The violation penalty, $E$, differs from that of order-embeddings and takes a more general form as defined by \cref{eq:euc_cones_E}.






\section{Hyperbolic Cones}
Instead of the cones and the embedding space conforming to Euclidean geometry one can formulate it to live in non-Euclidean space. Hyperbolic and spherical geometry both are non-Euclidean geometries (non-zero curvature) with negative and positive sectional curvature respectively.

The hyperbolic space can be modeled in 5 different ways. We use the Poincar\'e ball like previous work \cite{nickel2017poincare, ganea2018entailment_cones}.

The hyperbolic cones like the rest of the models are implemented in PyTorch \cite{pytorch}. We follow the schemes from \cite{ganea2018hyperbolicNN} to avoid numerical instabilities when learning parameters in the hyperbolic space, more specifically the Poincar\'e ball. 

\paragraph{Label embeddings.} For our implementation of the hyperbolic cones, the label-embeddings live in the hyperbolic space $\mathbb{D}^n$ and are optimized using the RSGD as per \cref{eq:riemannian_grad} and \cref{eq:riemannian_grad_update} with the help of the exponential-map from \cref{eq:exp_x_v}. RSGD is implemented by modifying the SGD gradients in PyTorch as it is not a part of the standard library.

\paragraph{Image embeddings.} For images, features from the final layer of the backbone of the best performing CNN-based model are used ($\in \mathbb{R}^{2048}$). In order to map them to $\mathbb{D}^n$ we use a linear transform $W \in \mathbb{R}^{2048 \times n}$ and then apply a projection into $\mathbb{D}^n$ via the exponential-map at zero which is equivalent to $\text{exp}_{0}(x)$. This bring the image embeddings to the hyperbolic space with Euclidean parameters. This allows for optimizing the parameters with well know optimization schemes such as Adam \cite{kingma2014adam}.

\section{Embedding Label-Hierarchy}

First we begin by learning to represent the hierarchy alone with this model. Considering only the label-hierarchy and excluding the images (momentarily), we model this problem as hypernym prediction where a hypernym pair represents two concepts $(x, y)$ such that $y$ \texttt{is-a} $x$. 

Hypernyms occur naturally in our ETHEC dataset \cite{dhall_20.500.11850/365379} where edges through different levels in the label-hierarchy represent the \texttt{is-a} relation with a directed edge from node $x$ to node $y$ representing $y$ \texttt{is-a} $x$.

\subsubsection{Data splitting}
We split the data into \texttt{train}, \texttt{val} and \texttt{test} in a similar manner to that of \cite{ganea2018entailment_cones}. They first compute the transitive reduction of the directed-acyclic graph. However, since it is a tree it is already in the most minimal form and we use the tree to form the ``basic'' edges for which the transitive closure can be fully recovered. If these edges are not present in the \texttt{train} set, the information about them is unrecoverable and therefore they are always included in the \texttt{train} set. Now, we randomly pick edges from the transitive closure (=1974 edges) minus the ``basic'' edges (=723 edges) to form a set of ``non-basic'' edges (=1257 edges). We use the ``non-basic'' edges to create \texttt{val} (5\%) (=62 edges) and \texttt{test} (5\%) (=62 edges) splits and a proportion of the rest are reserved for training (see Training details).

\subsubsection{Training details}
We follow the training details from \cite{ganea2018entailment_cones}. We augment both the validation and test set by generating 5 negative pairs each for $(x, y)$ (a positive pair): of the type $(x', y)$ and $(x, y')$ with a randomly chosen edge that is not present in the full transitive closure of the graph. Generating 10 negatives for each positive. For the training set, negative pairs are generated on-the-fly in the same manner. We report performance on different training set sizes. We vary the training set to include 0\%, 10\%, 25\%, 50\% of the ``non-basic'' edges selected randomly. We train for 500 epochs with a batch size of 10 and a learning rate of 0.01. We run two sets of experiments: one, we fix $\alpha=1.0$ as mentioned in \cite{vendrov2015order} and two, tune $\alpha$ based on the F1-score on the \texttt{val} set \cite{ganea2018entailment_cones}.

\subsubsection{\texttt{pick-per-level} strategy} During the experiments, we found a better strategy to sample negative edges. Instead of sampling a negative edge $(x', y)$ from candidates where $x'$ is any node that makes $(x', y)$ a negative edge, we pick each $x'$ from a different level in the hierarchy. This serves a dual purpose. Because the hierarchy is a tree, 78.24\% of the nodes belong to the final level in the hierarchy and if a \texttt{pick-per-level} strategy is not applied one would always sample edges where the corrupted end would be from the last level majority of the times. This makes training and convergence excruciatingly slow. Secondly, with this \texttt{pick-per-level} strategy we are able to sample nodes that give hard negatives edges from the same level as the non-corrupted node $y$, helping embeddings to disentangle and spread out in space.

\subsubsection{Optimization details}
We use Adam optimizer \cite{kingma2014adam} for order-embeddings and Euclidean cones. For hyperbolic cones we use Riemannian SGD \cite{ganea2018entailment_cones}.

\section{Jointly Embedding Images with Label-Hierarchy}
In the order-embeddings paper \cite{vendrov2015order}, the authors propose a two-level hierarchy for image-caption retrieval task. In their formulation, using a 2-level hierarchy, images are put on the lower-level and the captions on the upper level; the reason being: images are more detailed while captions represent concepts more abstract than the image itself. They also use a different loss from the one proposed in the hypernym prediction task.

For jointly embedding the images together with the labels we use the same hypernym loss from \cref{eq:order_embeddings_loss}. The only change being that now in addition to the labels,  $\mathcal{G}$ (the graph representing the hierarchy) also contains images as nodes as leaves at the lowest level.

$\mathcal{G}$ constitutes of two types of edges: an edge $(u, v)$ can be such that $u, v \in \text{labels}$  or $u \in \text{labels}, v \in \text{images}$. This is not of concern as the only difference is the way the embeddings are computed for images and labels. In the end, both $f_i$ and $f_l$ map respective inputs to the same space.

\subsubsection{Classification with Embeddings}
Since our problem does not concern hypernym prediction but rather assigning multiple labels to an image; instead of performing edge prediction (as the case would be in a hypernym prediction task) we use the embeddings for the task of classification.

To classify an image we compute the order-violation in \cref{eq:order_embeddings_E} between the given image and each label and pick the label corresponding to the minimum violation.

\begin{equation}
\label{eq:order_embeddings_classification_fn}
\text{arg}\,\min\limits_{l}\, E(f_{l}(l), f_{i}(i)), \forall l \in \text{labels}
\end{equation}

\subsubsection{Image and Label Embeddings}
To generate image embeddings we use the best performing CNN model trained on the ETHEC dataset \cite{dhall_20.500.11850/365379} and use it extract \textit{fc7}-features from the penultimate layer. We use a learnable linear transformation, a matrix $W$, on top of the \textit{fc7}-features to be able to adjust the \textit{fc7}-features and map them into the joint embedding space in $\mathbb{R}^{N}$ for Euclidean models and $\mathbb{D}^{N}$ for hyperbolic models (Poincar\'e disk).

\begin{equation}
\label{eq:order_embeddings_image_f}
f_{i}(i) = W * \text{CNN}(i) \in \mathbb{R}^{N}
\end{equation}

where, $\text{CNN}(i)$ represent the \textit{fc7}-features from our best performing CNN model and $W$ is a matrix. The weights of the CNN are frozen to calculate the \textit{fc7}-features with only $W$ that can be learned.

For the labels, $f_{l}(l)$ is just a lookup table that stores vectors in $\mathbb{R}^{N}$.

\subsubsection{Data splitting} For these experiments, we split the data into \texttt{train} (80\%), \texttt{val} (10\%) and \texttt{test} (10\%) based on images only as done for the CNN-based models. Since, now we embed images together with the labels, we create a combined graph $\mathcal{G}$ to represent both. The graph contains directed edges from each label that ``describes'' the image to the image itself as well as edges between related labels.

\subsubsection{Training details}
Let $\mathcal{G}$ represent the graph to be embedded. All edges in $\mathcal{G}_{tc}$, the transitive closure of $\mathcal{G}$, are considered as positive edges. To obtain negative edges, $\mathcal{G}_{neg}$ is constructed by removing the edges in $\mathcal{G}_{tc}$ from a fully-connected di-graph with the same nodes as $\mathcal{G}$.

While training, for each positive edge $(x, y)$ 10 negatives are sampled, 5 each by randomly corrupting either side of the positive edge ($5 \times (x', y) + 5 \times (x, y')$). We generate negatives by corrupting the edge with nodes from each level in the hierarchy including the images (the lowest level) because the images outnumber the labels and we would like to embed the label-hierarchy in addition to the images. We use the \texttt{pick-per-level} strategy as described in the previous section.

We make sure that we do not sample a negative edge such that either side of the edge is an image. This is to ensure that two images are not forced apart unless their labels require them to do so because the images for the final level and there are no cones that are nested with the cone formed by the image embedding (as it is the last level in the graph $\mathcal{G}$).

For the validation and test set, we generate 5 negative pairs each for $(x, y)$ (a positive pair): of the type $(x', y)$ and $(x, y')$ with a randomly chosen edge that is not present in the graph $\mathcal{G}_{neg}$. The validation and test sets are generated in the beginning and are fixed during training. We follow the training details from \cite{ganea2018entailment_cones}.

Negatives are sampled only for the training set and are generated on-the-fly. The model's performance is measured on its ability to classify images correctly. Since negative edges are not required for measuring classification performance, negative edges are sampled only during training. For validation and testing, we measure the model's classification the \texttt{val} and \texttt{test} set images respectively.

\paragraph{Graph reconstruction task.} In addition to the classification task, we also check the quality of reconstruction of the label-hierarchy itself. Here, all the edges in $\mathcal{G}$ that correspond to edges between labels are treated as positive edges, while the the edges in $\mathcal{G}_{neg}$ that correspond to edges between labels are treated as negative edges. We compute $E(u, v)\; \forall e \in \mathcal{P} \cup \mathcal{N}$ where $e=(u, v)$ and choose a threshold to classify edges as positive and negative using that yields the best F1-score on this label-hierarchy reconstruction task. This task does not use any edges that have an image on any side to check the quality of reconstruction.

For $W$ we use a linear transformation, a matrix $\mathbb{R}^{2048 \times N}$. Non-linearity is not applied to the output that maps to the embedding space.

\subsubsection{Optimization details}
For jointly embedding labels and images, we empirically found using vanilla Adam \cite{kingma2014adam} optimizer instead of the Riemannian SGD. The drawback being that the label embeddings are parameterized in the Euclidean space and we use the exponential map at 0 from \cref{eq:exp_x_v} to map them to the hyperbolic space. This was observed to be more stable and help converge the joint embeddings. Also, with this implementation of the hyperbolic cones, for both labels and joint embeddings, it was not necessary to initialize the embeddings with the Poincar\'e embeddings \cite{nickel2017poincare} as suggested in \cite{ganea2018entailment_cones}.

We even notice that when training jointly one does not need to initialize the labels with separate labels-only embedding. The model is still able to attain decent image classification performance when the label embeddings are randomly initialized. However, a performance boost is obtained when initialized with values from embedding only the label-hierarchy.

\chapter{Empirical Analysis: Order-preserving embedding-based models}
\label{ch:emb_analysis}

\section{Embedding Label-Hierarchy}
We first embed the label-hierarchy exclusively. We report results where we perform embed only the complete ETHEC Merged Dataset label-hierarchy without any images. From \cref{table:order_embeddings_label_emb} one can see that even with very few dimensions, one is able to achieve a high F1 score on the hidden edges from the \texttt{test} set. Also, in the same low-dimensions the entailment cones perform better than the order-embeddings confirming that they are a generalization of the latter. The metrics in \cref{table:order_embeddings_label_emb} are calculated on a set of hidden edges that form the \texttt{test} set of edges. These metrics are different from the ones in the section below that discusses the graph reconstruction task.

\begin{figure}[!htbp]
    \centering
    \includegraphics[width=0.9\textwidth]{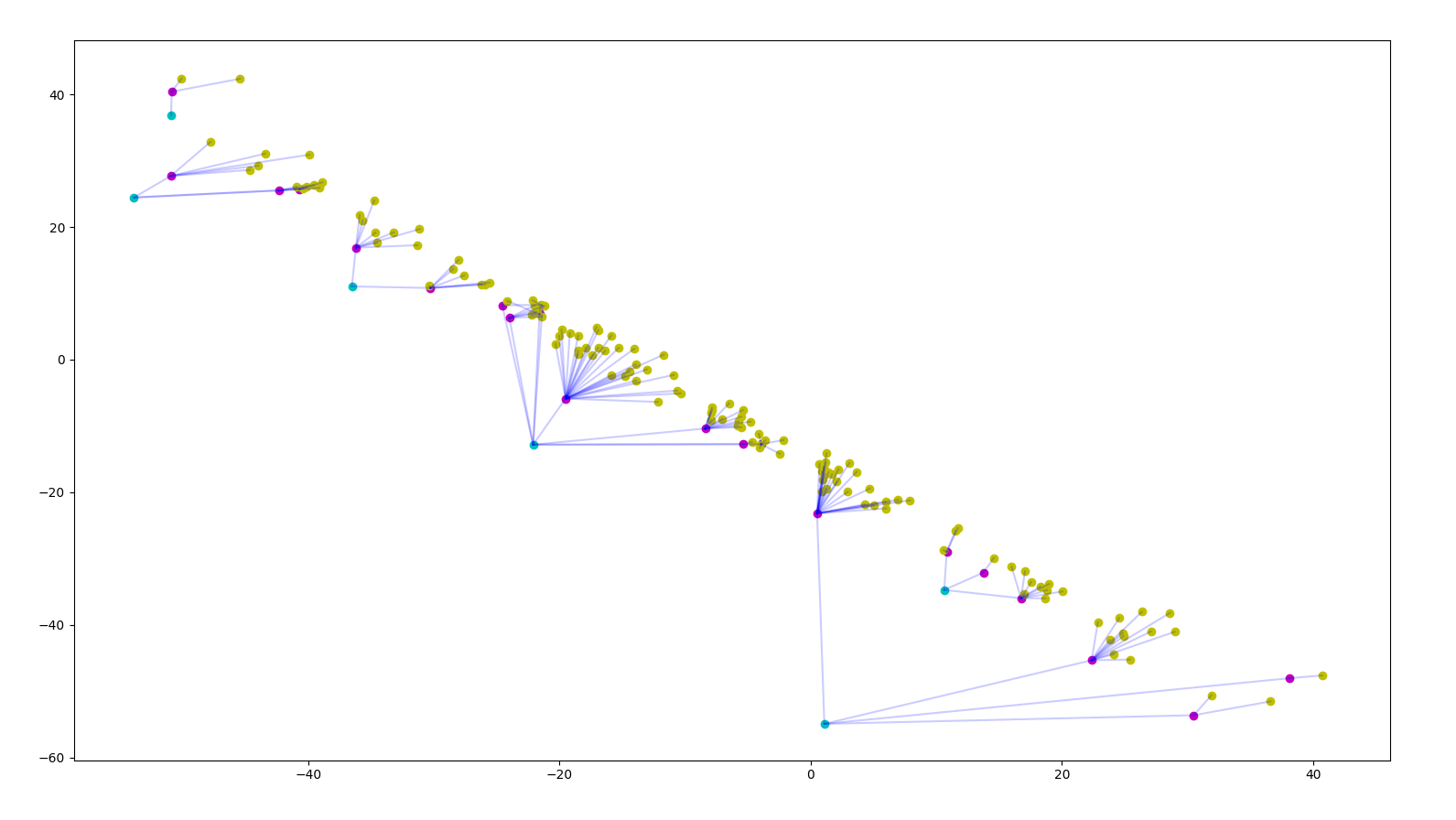}
    \caption{Visualization of the label-hierarchy using order-embeddings in 2 dimensions. The cyan nodes represent \texttt{family}, the magenta nodes represent \texttt{sub-family}, the yellow nodes \texttt{genus}. \texttt{genus+species} nodes are omitted to visualize better.}
    \label{fig:oe_2d_labels}
\end{figure}

\begin{figure}[!htbp]
    \centering
    \includegraphics[width=0.9\textwidth]{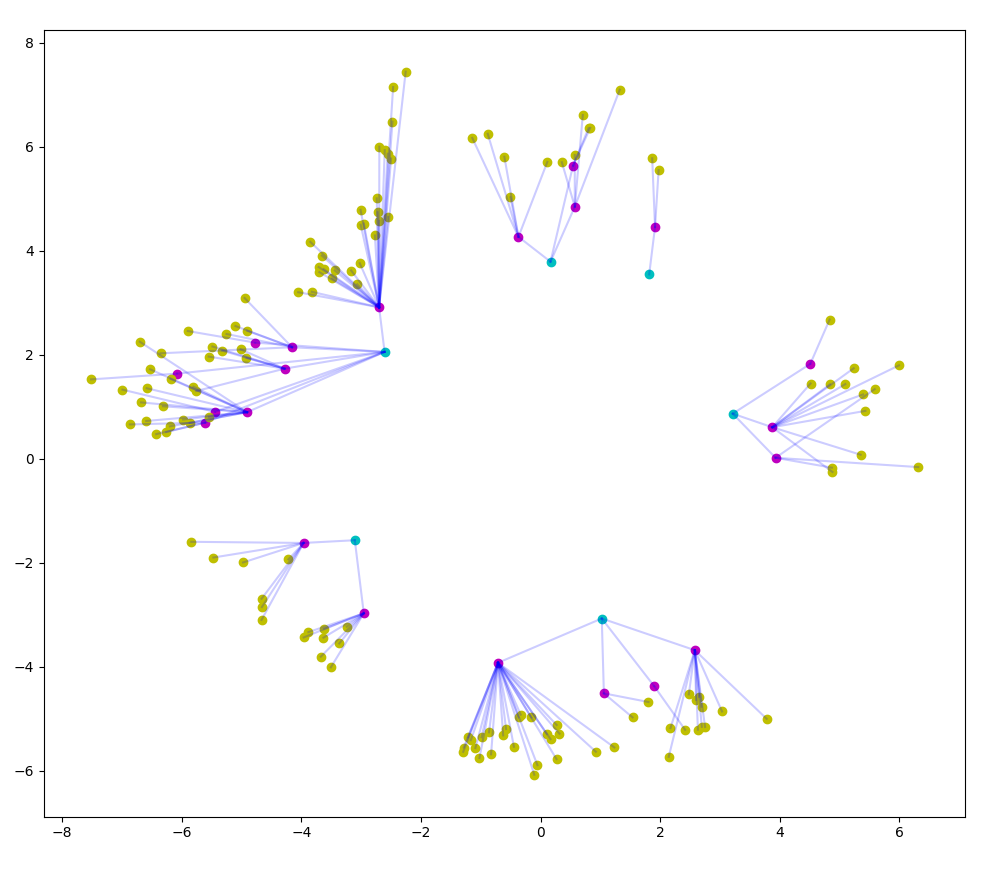}
    \caption{Visualization of the label-hierarchy using Euclidean cones in 2 dimensions. The cyan nodes represent \texttt{family}, the magenta nodes represent \texttt{sub-family}, the yellow nodes \texttt{genus}. \texttt{genus+species} nodes are omitted to visualize better.}
    \label{fig:ec_2d_labels}
\end{figure}

\begin{table}[!htbp]
\centering
\begin{tabular}{| c || c | c | c | c | c |} 
  \hline
  Model & d=2 & d=3 & d=5 & d=10 & d=100 \\ [0.5ex] 
  \hline\hline
  Order-embeddings & 0.8271 & 0.9302 & 0.9457 & 0.9920 & 0.9920 \\ \hline
  Euclidean Cones & 0.8550 & 0.9979 & 0.9593 & 0.9919 & 0.9752 \\ \hline
\end{tabular}
\caption{Micro-F1 score on the \texttt{test} set for embeddings on the label hierarchy of ETHEC Merged dataset. We find the classification threshold that yields the best \texttt{val} set performance. For these experiments we train for 200 epochs with $\alpha=1.0$ for order-embeddings and $\alpha=0.01$ for Euclidean cones, a batch-size of 10 and a learning-rate of 0.1 with Adam optimizer. The F1-score corresponding to the \texttt{test} set for the epoch with the best F1-score on the \texttt{val} set is reported. \texttt{train} set is composed of all the ``basic'' edges and an additional 50\% of the ``non-basic'' edges for all experiments reported here. We vary the dimensionality of the embedding space, d = \{2, 3, 5, 10, 100\}.}
\label{table:order_embeddings_label_emb}
\end{table}

\begin{figure*}[!htbp]
    \centering
    \begin{subfigure}[b]{0.480\textwidth}
        \centering
        \includegraphics[width=\textwidth]{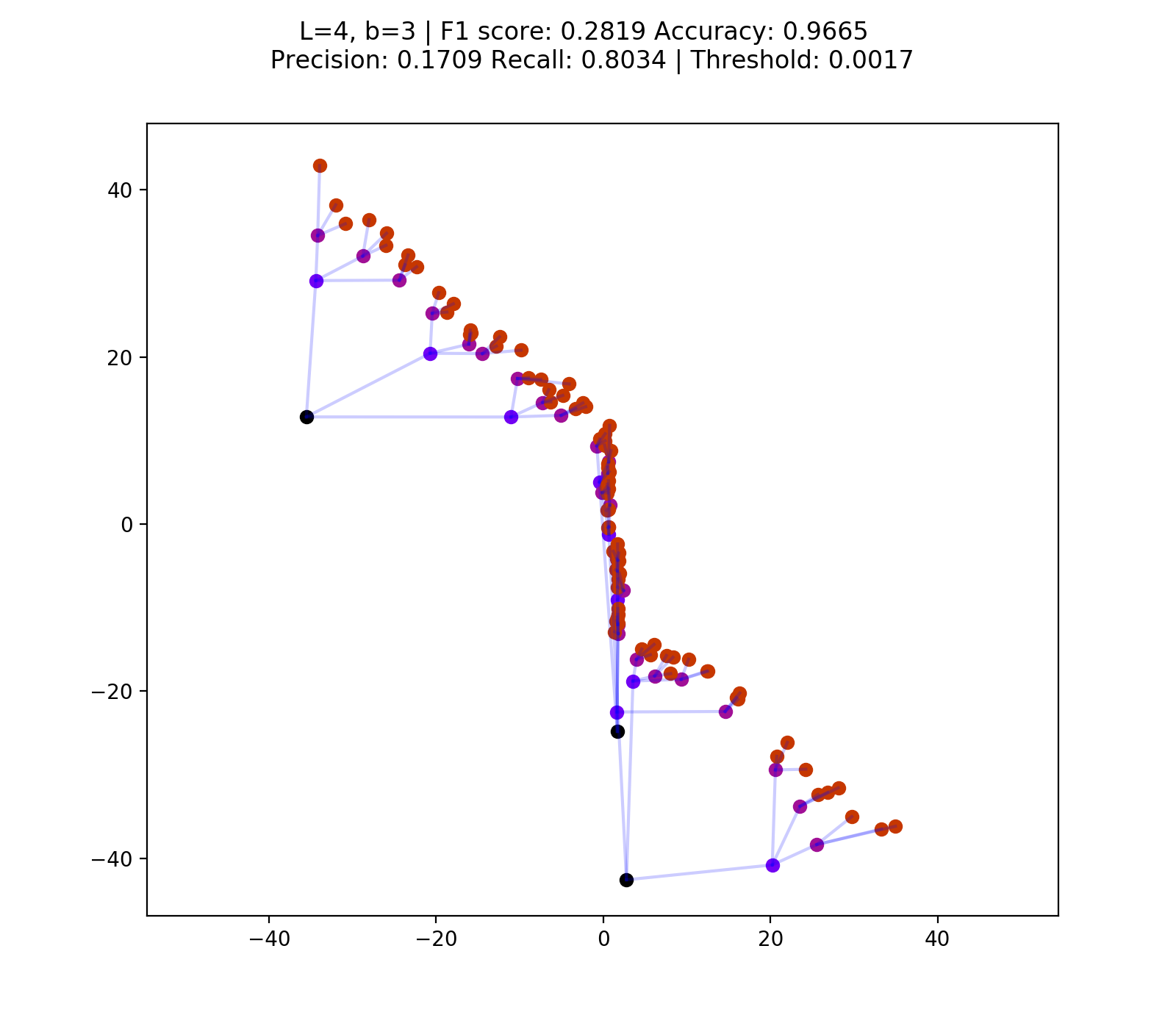}
        \caption[]%
        {{\small \textit{Order-embeddings L=4, b=3}}}    
        \label{fig:oe_l4b3}
    \end{subfigure}
    \hfill
    \begin{subfigure}[b]{0.480\textwidth}  
        \centering 
        \includegraphics[width=\textwidth]{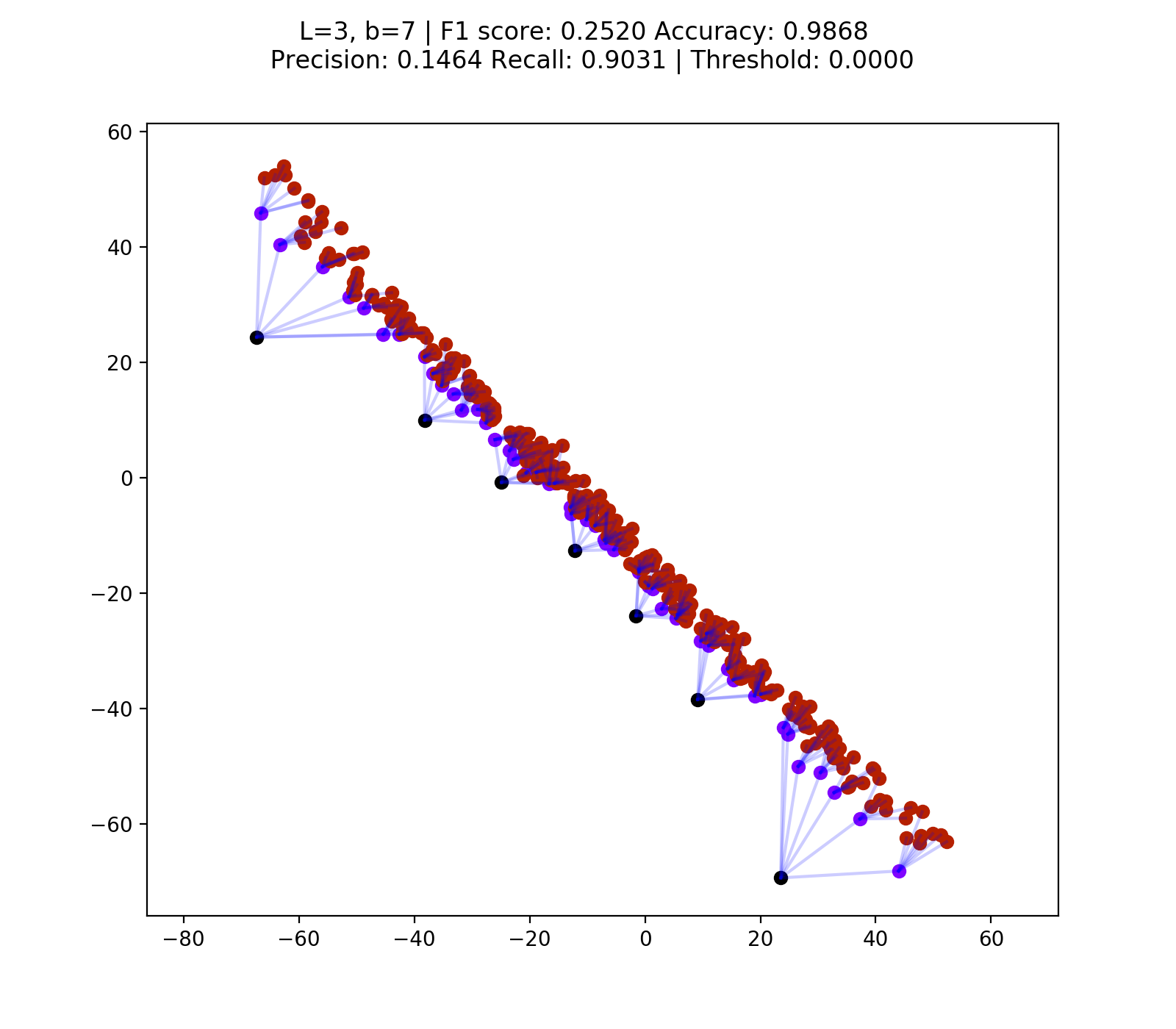}
        \caption[]%
        {{\small Order-embeddings L=3, b=7}}    
        \label{fig:oe_l3b7}
    \end{subfigure}
    \vskip\baselineskip
    \begin{subfigure}[b]{0.480\textwidth}   
        \centering 
        \includegraphics[width=\textwidth]{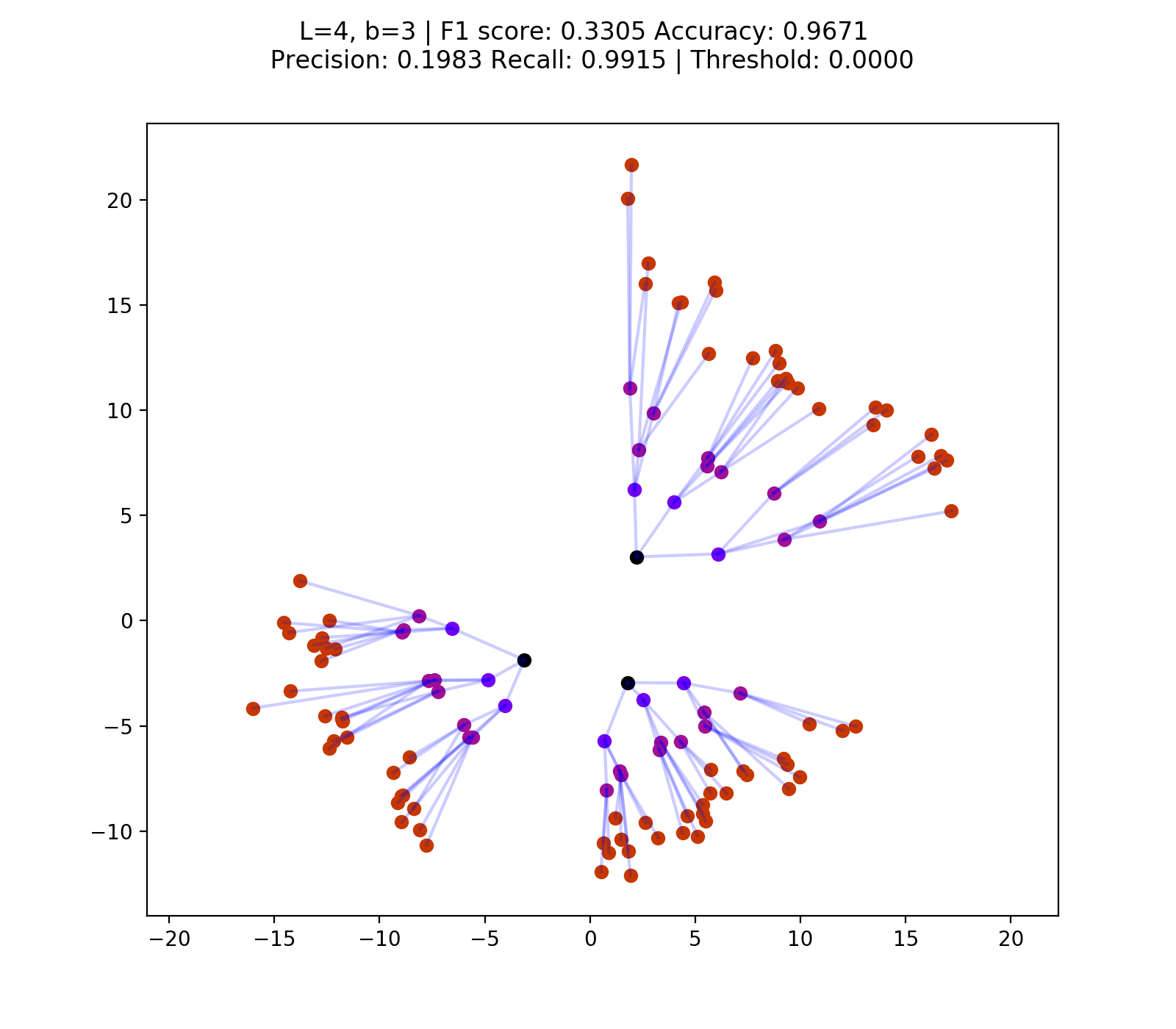}
        \caption[]%
        {{\small Euclidean cones L=4, b=3}}    
        \label{fig:ec_l4b3}
    \end{subfigure}
    \quad
    \begin{subfigure}[b]{0.480\textwidth}   
        \centering 
        \includegraphics[width=\textwidth]{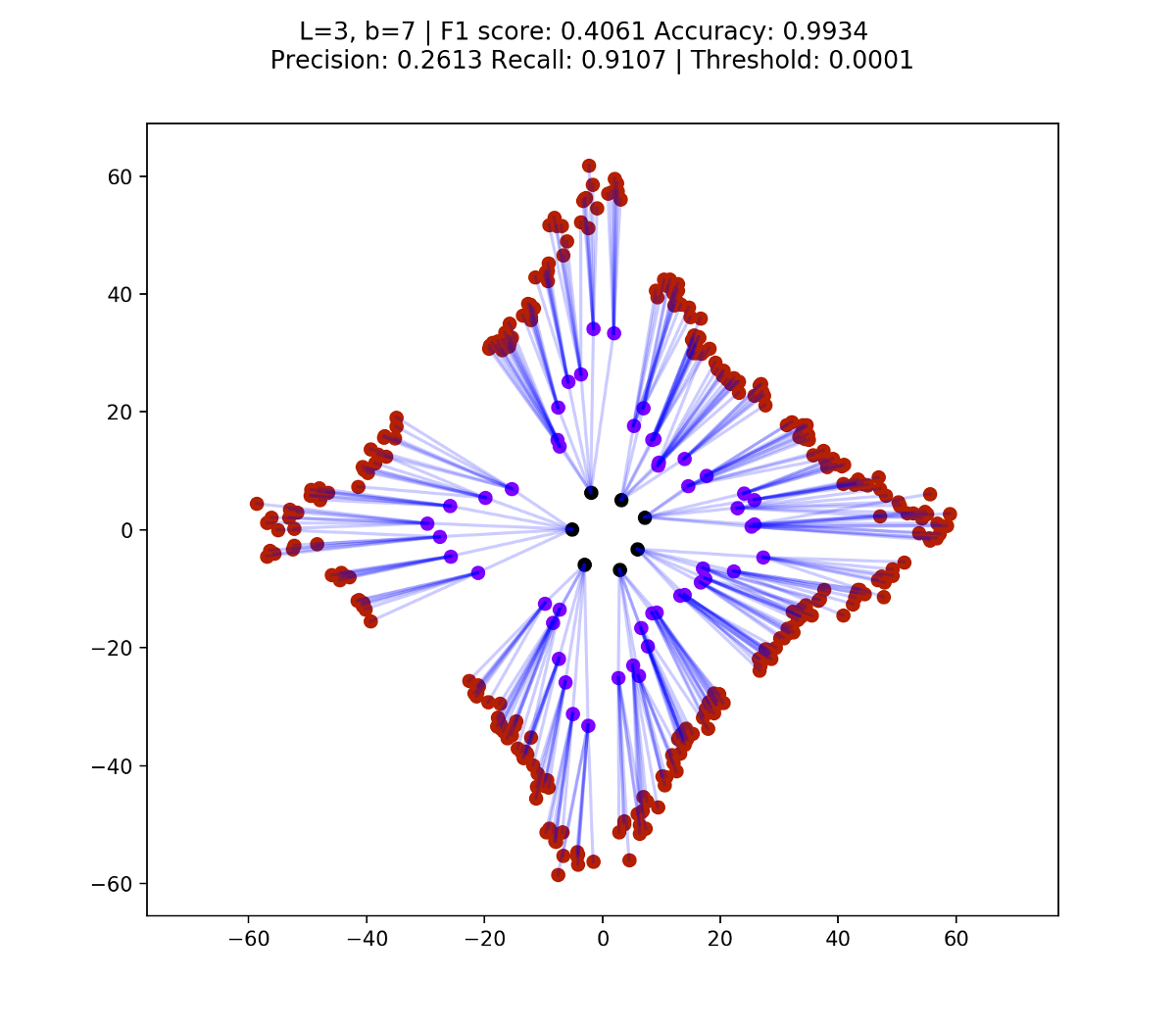}
        \caption[]%
        {{\small Euclidean cones L=3, b=7}}
        \label{fig:ec_l3b7}
    \end{subfigure}
    \caption[]
    {\small \centering We embed 2 different toy graphs. One with 4 levels and a branching factor of 4 and another one with 3 levels and a branching factor of 7. The model is trained for 1000 epochs with Adam (learning rate of 0.01). The toy graphs are embedded using both order-embeddings and euclidean cones in $\mathcal{R}^2$. We draw an edge between each node that is connected in the original in order to better visualize the embedding quality. Nodes from different levels are colored differently. The illustrations show the levels and branching factor, the edges are split into \texttt{train}, \texttt{val} and \texttt{test} and report F1-score, precision, recall and accuracy; and the threshold to decide if a pair of nodes have a directed edge or equivalently if they are hypernyms.} 
    \label{fig:toy_trees}
\end{figure*}

\subsection{Graph reconstruction quality for label-embeddings}

In addition to the entailment prediction task between two given concepts \cite{ganea2018entailment_cones, nickel2017poincare, suzuki2019hyperbolic_disk} we also check the reconstruction of the complete graph which basically checks the ability of the embedding to reproduce the asymmetric relations in the original label hierarchy. This consists of positive and negative directed edges from the original labels where the positive edges are present in the label-hierarchy while negative edges are non-existent edges. It is important to note that only a handful of edges are positive while a vast majority of the edges in the fully-connected digraph for the set of negative, non-existent edges (in the original label-hierarchy).

For the ETHEC dataset, to obtain the full-F1 (measuring the reconstruction of the label-hierarchy) we classify all the 723 positive and 521,289 negative edges. Due to this very large imbalance between the negative and positive edges we refrain from using accuracy or micro/macro F1 score and used the TPR, TNR and full-F1 instead.

\begin{table}[!htbp]
\centering
\begin{tabular}{| c || c | c | c |} 
  \hline
   & d=2 & d=100 & d=1000 \\ [0.5ex] 
  \hline
   & $\text{TPR} /\ \text{TNR} /\ (\text{full-F1})$ & $\text{TPR} /\ \text{TNR} /\ (\text{full-F1})$ & $\text{TPR} /\ \text{TNR} /\ (\text{full-F1})$ \\
  \hline\hline
  OE & 0.2309 /\ 0.9708 /\ (0.1372) & 0.4686 /\ 0.9880 /\ (0.3894) & 0.3788 /\ 0.9878 /\ (0.3489) \\ \hline
  EC & 0.3617 /\ 0.9975 /\ (0.3573) & 0.4802 /\ 0.9985 /\ (0.4151) & 0.5790 /\ 0.9973 /\ (0.4091) \\ \hline
  HC & 0.4443 /\ 0.9907 /\ (0.2296) & 0.9336 /\ 0.9986 /\ (0.8060) & \textbf{0.9721} /\ \textbf{0.9986} /\ \textbf{(0.8257)} \\ \hline
\end{tabular}
\caption{Graph reconstruction performance. Here we measure the ability to reconstruct all positive and negative edges in the original label-hierarchy using the embeddings. We choose the threshold boundary by picking the one that maximizes the full-F1 score. This is the F1-score on all positive and negative edges in the (full) label-hierarchy. Due to the imbalance the F1-score may not completely represent the reconstruction ability so we also report the true positive rate (TPR) and true negative rate (TNR). The cells in table are formatted as $\text{TPR} /\ \text{TNR} /\ (\text{full-F1})$. The model is trained with labels only for 5000 epochs. HC is trained using RSGD with a $\text{lr}=0.001$, OE and EC are optimized with Adam\cite{kingma2014adam} with a $\text{lr}=0.1$. For both the margin used is $\alpha=0.1$ and sampling 10 negatives for each positive and a batch size of 100 for HC and 10 for OE and HC. \texttt{Legend:} OE: Order-embeddings, EC: Euclidean cones, HC: Hyperbolic cones.}
\label{table:label_emb_full_comp}
\end{table}

In \cref{table:label_emb_full_comp} we compare the embedding performance for labels-only from the ETHEC dataset. We employ order-preserving embedding techniques as discussed in previous chapters from order-embeddings \cite{vendrov2015order} and euclidean and hyperbolic variants of \cite{ganea2018entailment_cones}.

Positive edges constitute of only about 0.1\% of the total edges in the DAG representing the label-hierarchy; making it very difficult to predict these. This is also evident empirically as the TNR is quite high for extremely low-dimensions. For 2D order-embeddings the TNR=0.9708 despite being the lowest for 2D embeddings.

The performance boost achieved by entailment cones, a generalization of order-embeddings, can be seen from the table where the euclidean variant is always better than order-embeddings. By parameterizing the cones to live in hyperbolic space and use the corresponding hyperbolic geometry, they are able to achieve almost twice the TPR of OE and EC in 100 dimensions. Further increasing the dimensions to 1000 dimensions improves performance over 100-dimensional HC from 0.8060 to 0.8267 in full-F1 score, exhibiting the representative capacity of HC.

We also note that for 1000-D EC and OE, the model seems to overfit and perform worse than the 100-D counterparts. However the HC improves with increase in the embedding dimensionality.

\begin{figure*}[!htbp]
    \centering
    \begin{subfigure}[b]{0.49\textwidth}
        \centering
        \includegraphics[width=\textwidth]{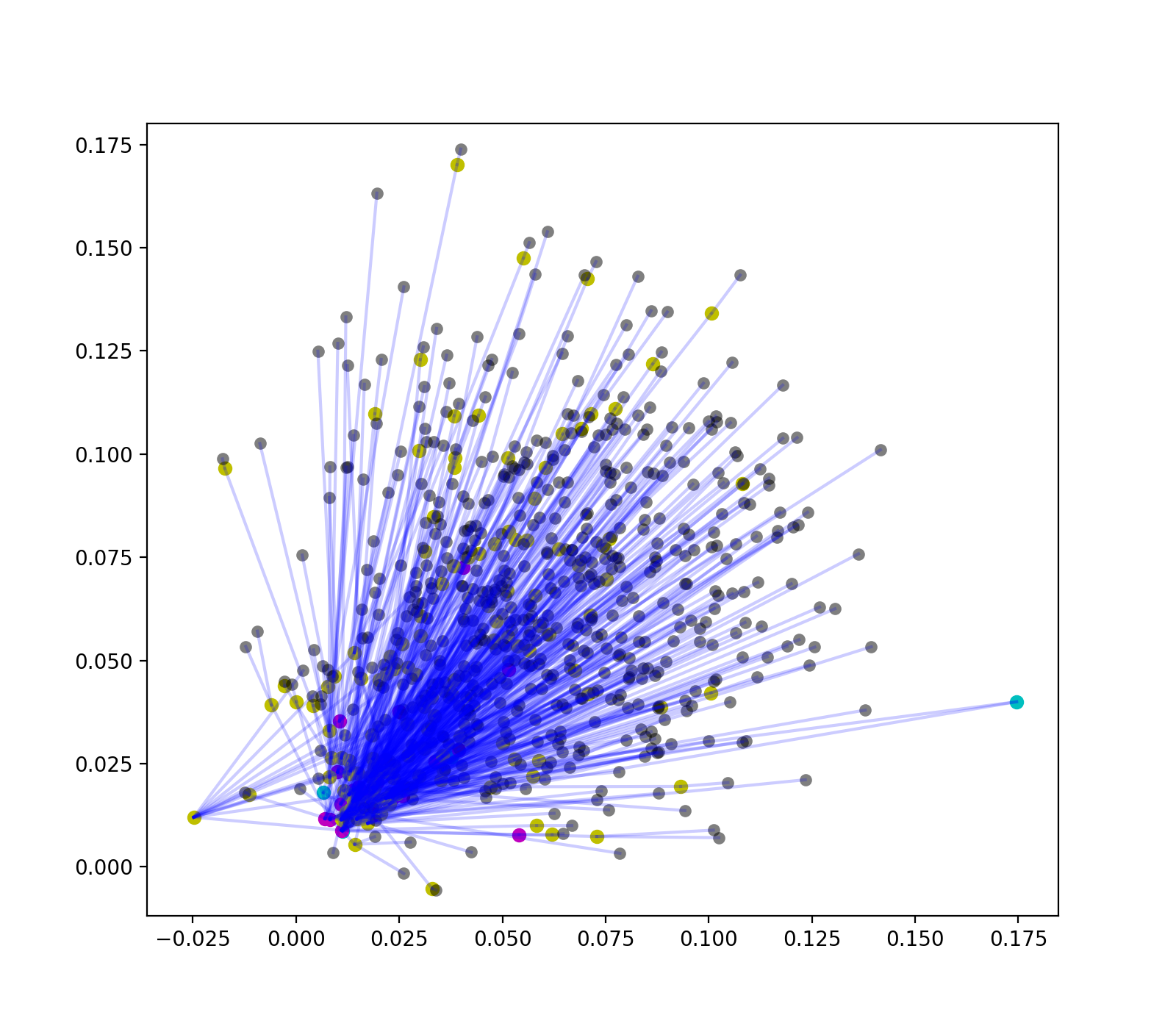}
        \caption[]%
        {{\small Hyperbolic Cones 100-D}}    
        \label{fig:hc_100d_labels}
    \end{subfigure}
    \hfill
    \begin{subfigure}[b]{0.49\textwidth}  
        \centering 
        \includegraphics[width=\textwidth]{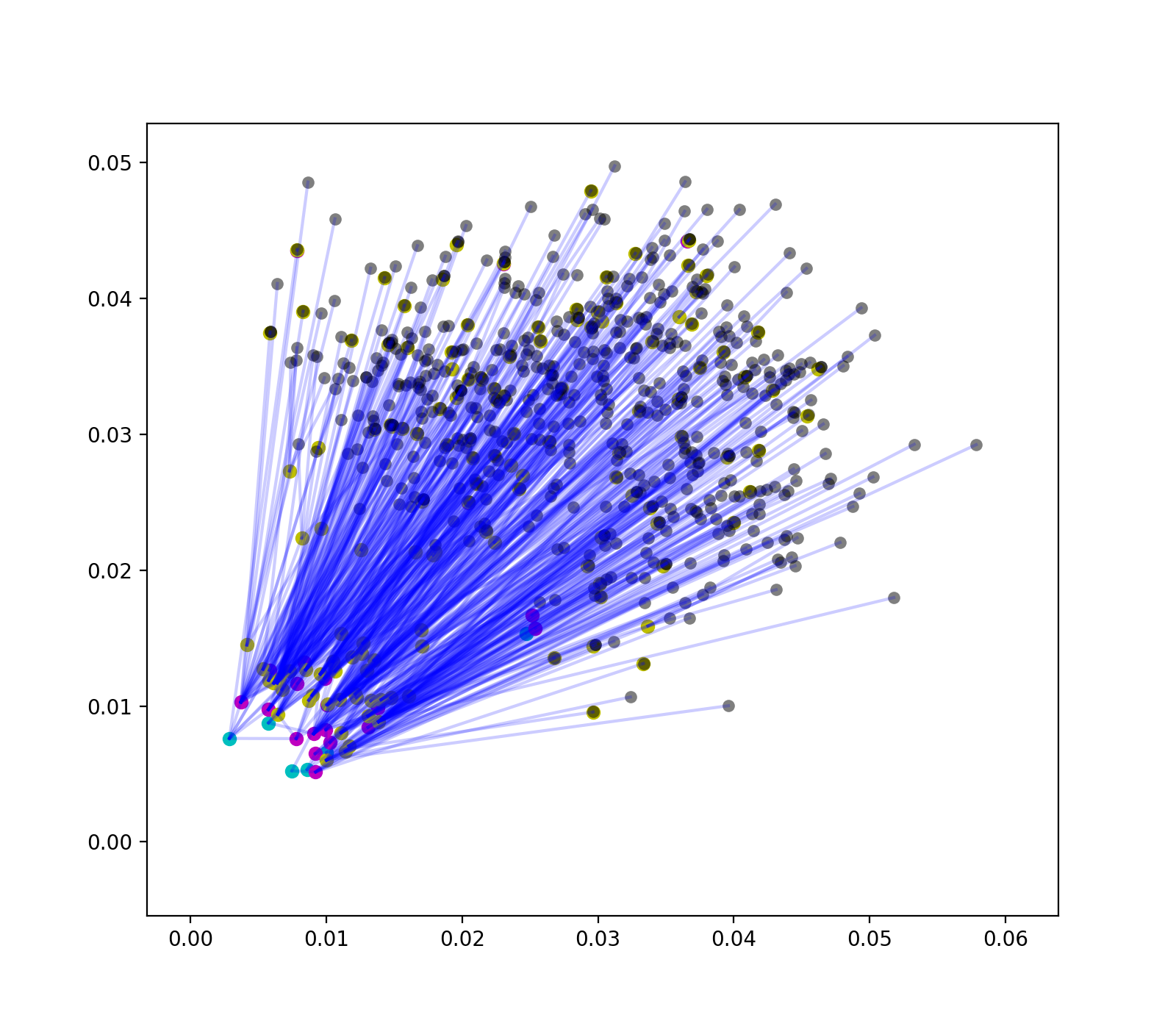}
        \caption[]%
        {{\small Hyperbolic Cones 1000-D}}    
        \label{fig:hc_1000d_labels}
    \end{subfigure}
    \vskip\baselineskip
    \caption[]
    {Projected visualization of labels embedded using hyperbolic cones in 100 and 1000 dimensions. The cyan nodes represent \texttt{family}, the magenta nodes represent \texttt{sub-family}, the yellow nodes \texttt{genus} and black nodes \texttt{genus+species}. This resembles a flower-like shape where the more generic concepts are closer to the origin and at the base of this flower-like shape and most specific concepts at the tip of the petals which forms the periphery are a visible the most (=black nodes).} 
    \label{fig:hc_100d_1000d}
\end{figure*}

\subsection{Optimization}
Initially, the experiments used a batch-size of 10 and the models still had room for improvement at the end of 5000 epochs. However, with a batch-size of 100 the models converged faster and also performed better. In general it was easier to find hyperparameters for euclidean models that the non-euclidean ones. We also noticed better, more stable training when parameterizing the euclidean cones in cosine space rather than angle space. For euclidean cones experiments we implement and use the cosine space formulation to compute the energy $E$ from \cref{eq:euc_cones_E}.

\section{Jointly Embedding Images with Label-Hierarchy}

When performing joint embedding, we simply add images to the existing graph of the label-hierarchy. The notion behind which is to treat the images and labels in the same manner. This allows to re-use the label-embeddings to additionally embed images for the task of image classification.

To perform image classification using embeddings, the least violating order-penalty $E(f_{l}(l), f_{i}(i))$ across all possible labels for a given image is considered as the predicted label. For every level in the label-hierarchy this is done once, with the labels from that particular level.

\begin{figure}[!htbp]
    \centering
    \includegraphics[width=0.9\textwidth]{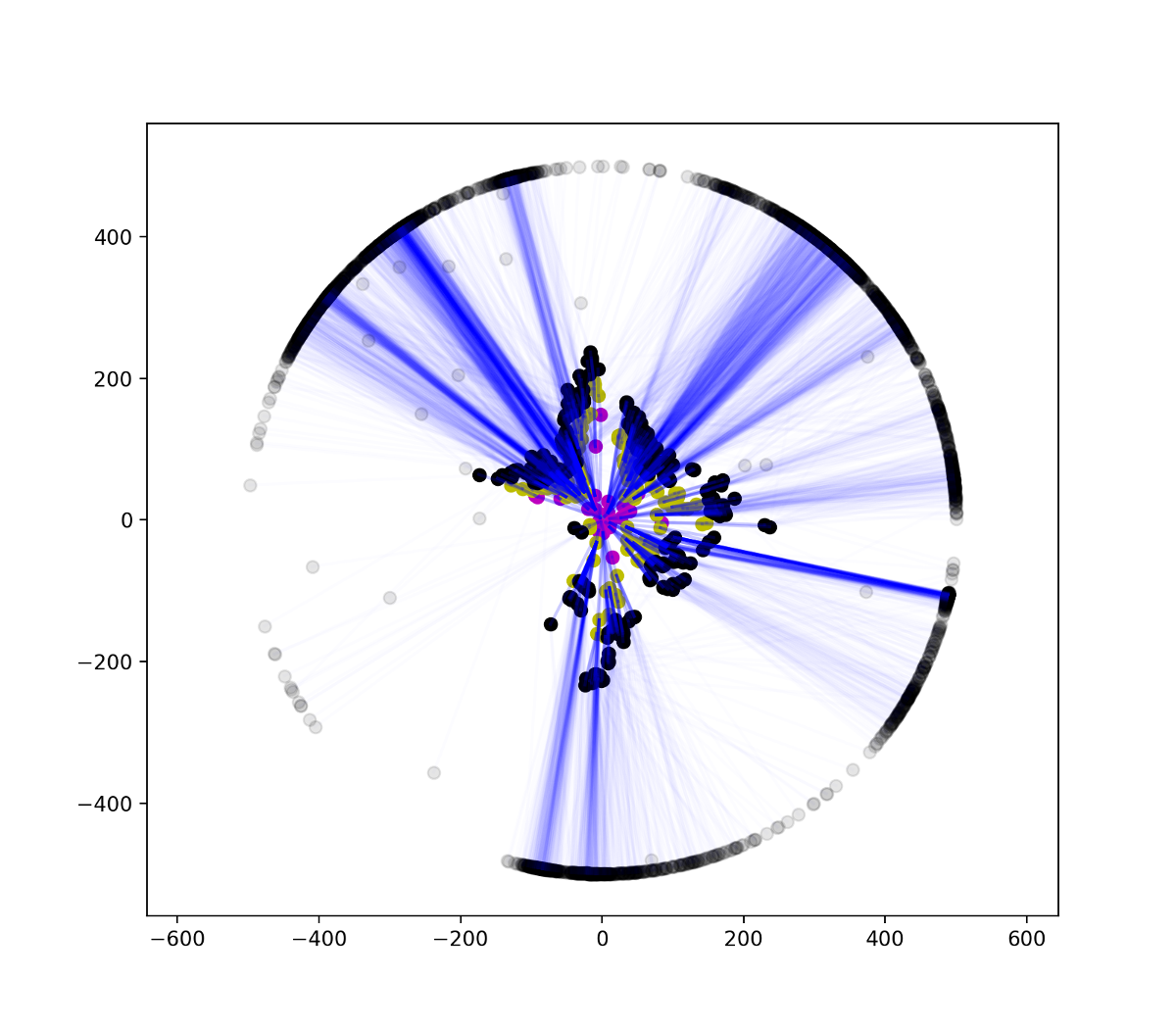}
    \caption{Visualization of jointly embedding labels and images using Euclidean cones in 2 dimensions. The cyan nodes represent \texttt{family}, the magenta nodes represent \texttt{sub-family}, the yellow nodes \texttt{genus} and black nodes \texttt{genus+species}. The images are depicted using semi-transparent nodes which are accumulated away from the origin around the periphery. To better visualize this, we clamp the norms of image embeddings (to 500 units) in order to visualize them with labels which are embedded much closer to the origin (due to them being more abstract).}
    \label{fig:ec_2d_labels_images}
\end{figure}

\begin{table}[!htbp]
\centering
\begin{tabular}{| c || c | c | c || c | c | c |}
  \hline
  & \multicolumn{3}{|c|}{classify \texttt{test} set images} & \multicolumn{3}{|c|}{graph reconstruction} \\ \hline
  
  Model & \textbf{m-F1} & hit@3 & hit@5 & TPR & TNR & full-F1 \\ [0.5ex] 
  \hline\hline
  \multicolumn{7}{|c|}{Euclidean Cones} \\ \hline
  d=10 & 0.7795 & 0.8893 & 0.9204 & 0.8045 & 0.9982 & 0.7040 \\ \hline \hline
  d=100 & 0.8350 & 0.9018 & 0.9425 & \textbf{0.9630} & \textbf{0.9986} & \textbf{0.8210} \\ \hline
  d=1000 & 0.8013 & 0.8971 & 0.9278 & 0.8146 & 0.9981 & 0.7073 \\ \hline
  \hline\hline
  \multicolumn{7}{|c|}{Hyperbolic Cones} \\ \hline
  d=100 & \textbf{0.8404} & \textbf{0.9200} & \textbf{0.9386} & 0.6418 & 0.9978 & 0.5756 \\ \hline
  d=1000 & 0.8045 & 0.9023 & 0.9281 & 0.5233 & 0.9973 & 0.4832 \\
  \hline\hline

\end{tabular}
\caption{The table summarizes the embedding model performance when used to classify images. The metrics are reported on the hidden images of the \texttt{test} set of the ETHEC dataset. The joint image and label embeddings live in $\mathbb{R}^{d}$ or $\mathbb{D}^{d}$ (d=dimensionality of embedding space). The main metric to look at is the m-F1 for image classification performance using the embeddings. This is directly comparable to the CNN-based models and the hierarchy-agnostic model (baseline) as well. Since these models are embedding based, in addition, to the classification task, we report the quality of the reconstruction for the label-hierarchy obtained during the joint embedding. The EC models use $\alpha=1.0$ trained for 200 epochs with a learning-rate of $10^{-2}$ for the label embeddings, $10^{-3}$ for the image embeddings using Adam. For the HC we train for 100 epochs with a learning-rate of $10^{-4}$ for the label embeddings, $10^{-3}$ for the image embeddings using Adam. We use a batch-size of 100 for all with 10 negatives per positive \texttt{pick-per-level} sampling. We initialize the models' label embeddings using label-only embeddings for all. Best models are bold-faced. EC: euclidean cones, HC: hyperbolic cones.}
\label{table:emb_classification_full_results}
\end{table}

\subsection{Optimization}
The EC models use $\alpha=1.0$ trained for 200 epochs with a learning-rate of $10^{-2}$ for the label embeddings, $10^{-3}$ for the image embeddings using Adam. For the HC we train for 100 epochs with a learning-rate of $10^{-4}$ for the label embeddings, $10^{-3}$ for the image embeddings using Adam. We use a batch-size of 100 for all with 10 negatives per positive \texttt{pick-per-level} sampling. We initialize the models' label embeddings using label-only embeddings for all.

\subsection{Hierarchical level-wise classification performance}

In this section we compare level-wise classification performance for the order-preserving embedding models. For easier, side-by-side comparison, performance of CNN-based models proposed earlier is also included.

\begin{table}[!htbp]
\centering
\begin{tabular}{| c || c || c | c | c | c |} 
 \hline
 \multicolumn{1}{|c||}{} &
 \multicolumn{1}{c||}{} &
 \multicolumn{4}{c|}{Per-level micro-F1} \\
 \hline\hline
   Model & m-F1 & m-F1 $L_{1}$ & m-F1 $L_{2}$ & m-F1 $L_{3}$ & m-F1 $L_{4}$  \\ [0.5ex] 
 \hline\hline
 \multicolumn{6}{|c|}{CNN-based methods} \\
 \hline \hline
  Hierarchy-agnostic (baseline) & 0.8147 & 0.9417 & 0.9446 & 0.8311 & 0.4578 \\ \hline \hline
  Per-level classifier & 0.9084 & 0.9766 & 0.9661 & 0.9204 & 0.7704 \\ \hline
  Marginalization classifier & \underline{\textbf{0.9223}} & \underline{\textbf{0.9887}} & \underline{\textbf{0.9758}} & \underline{\textbf{0.9273}} & \underline{\textbf{0.7972}} \\ \hline
  Masked Per-level classifier & 0.9173 & 0.9828 & 0.9701 & 0.9233 & 0.7930 \\ \hline
  Hierarchical-softmax & 0.9180 & 0.9879 & 0.9731 & 0.9253 & 0.7855 \\ \hline
  \multicolumn{6}{|c|}{Order-preserving (joint) embedding models} \\
  \hline \hline
  Euclidean cones d=100 & 0.8350 & 0.9728 & 0.9370 & 0.8336 & 0.5967 \\ \hline
  Hyperbolic cones d=100$*$ & 0.7627 & 0.9695 & 0.9205 & 0.7523 & 0.4246 \\ \hline
  Hyperbolic cones d=100 & \textbf{0.8404} & \textbf{0.9800} & \textbf{0.9439} & \textbf{0.8477} & \textbf{0.5977} \\ \hline
  
\end{tabular}
\caption{Comparing level-wise performance across different models both CNN-based and embeddings based as proposed in the main body of the work. All models that exploit any information from the hierarchy outperform the hierarchy-agnostic classifier baseline. All scores are micro-averaged F1. We also include the overall m-F1 in addition to the separate m-F1 across the 4 levels in the ETHEC dataset. The best overall model is underlined and the best model in the model category is bold-faced. Label embeddings for all joint-embeddings models are initialized using labels-only embeddings. $*$ = randomly initialized label embeddings.}
\label{table:combined_levelwise_performance}
\end{table}

\cref{table:combined_levelwise_performance} shows that the hierarchy-agnostic baseline is beaten by all models which use hierarchical information in one form or the other.

Embeddings that are completely different class of models and are seen to be widely used in the context of natural language but are relatively unexplored for image classification, are also able to outperform the baseline CNN classifier. The embeddings also perform better for most of the levels in the level-wise m-F1 score column.

\subsection{W's model capacity}
We also use a simple matrix $W$ that transforms \textit{fc7} image features to the embedding space. A more elaborate 4-layer feed-forward neural network was also used but we found it to be worse performing and hard to optimizer. Instead of using the CNN to extract features one could instead have all its parameters learn-able during the joint-embedding training procedure but in our experiments we observed it to drastically over-fit and have high performance on the \texttt{train} set and extremely low performance on the unseen \texttt{test} set.

\subsection{Sampling strategy}
To perform sampling of negative edges, we use the \texttt{pick-per-level} strategy with 5 $(u', v)$ + 5 $(u, v')$ negative edges for each positive edge. Different number of negative edges were tried with 1+1 and 50+50 but the 5+5 (=10:1 negative to positive edge ratio) worked best.

For joint-embedding the ETHEC dataset, since the images (around 50,000) outnumber the labels (723) we thought it might be useful to randomly sample negative edges such that the ratio of negative nodes have a proportion to be 50\%:50\% for images:label ratio however, the original strategy works better.

\subsection{Choice of Optimizer}
Initial experiments for the hyperbolic cones (HC) used the RSGD optimizer as it seemed to work for labels-only embeddings hyperbolic cones. When using the same to optimize over the labels for the joint-embedding model, we noticed that the label hierarchy moves towards the image labels and ends up collapsing from a very good initialization (taken from the labels-only embeddings). The collapse leads to entanglement between nodes from different labels and images, which leads it to a point of no return and the performance worsens due to the label-hierarchy becoming disarranged and its inability to recover. We believe that the reason for its inability to rearrange is due to there being a two different types of objects being embedded (and also being computed differently) and it compounded by using different optimizers.

In our experiments we obtain best results when using the Adam optimizer even if it means the update step for parameters living in hyperbolic space has to be performed in an approximate manner. Adam optimizer with an approximate update step works better in practice than RSGD with its mathematically more precise update step.

\subsection{Label initialization for joint-embeddings}
Using RSGD we observed that if the labels are not initialized with the labels-only embedding then the joint model finds it difficult to disentangle the label embeddings and eventually this effect is cascaded to the images causing the image classification performance to not improve.

With the RSGD replaced by the Adam optimizer, in experiments where we randomly initialized the label-embeddings, we observed them to disentangle and form entailment cones even with the images being involved and making the optimization more complex. The joint-model still works decently well with random label initialization and achieves an image classification m-F1 score of 0.7611 and even outperforms the hierarchy-agnostic CNN in the m-F1 for $L_1$ labels (see \cref{table:combined_levelwise_performance} for details).

\cite{ganea2018entailment_cones} recommends to use Poincar\'e embeddings \cite{nickel2017poincare} to initialize the hyperbolic cones model. The fact that the joint model as well as the labels-only hyperbolic cones have great performance without any special initialization scheme is interesting. We conjecture that this could be because of using an approximate yet better optimizer. 

\subsection{Inverted cosine embeddings and euclidean cones}

We notice the similarity between embedding label-hierarchy with Euclidean cones and the inverted cosine embeddings. We invert the cosine embeddings in order to have the most abstract concepts close to the origin and more specific concepts farther away. The cosine embeddings arrange themselves in an inverted manner due to them being more accurate and confident about labels from the upper levels in the hierarchy due to the dot product nature of a fully-connected layer in a neural network (=matrix multiplication and a non-linearity).

We invert the cosine embeddings by re-scaling them such that the one furthest are close to the origin and vice-versa.

\begin{equation}
\label{eq:invert_cosine_emb}
x_{\text{inverted}} = \frac{r * x * ||x_{\text{max}}||}{||x||}
\end{equation}

where, $x_{\text{max}}$ is the cosine embedding with the largest norm and $r \in \mathbb{R}$ is the minimum norm that any inverted embedding should have.

\begin{figure}[!htbp]
    \centering
    \includegraphics[width=0.9\textwidth]{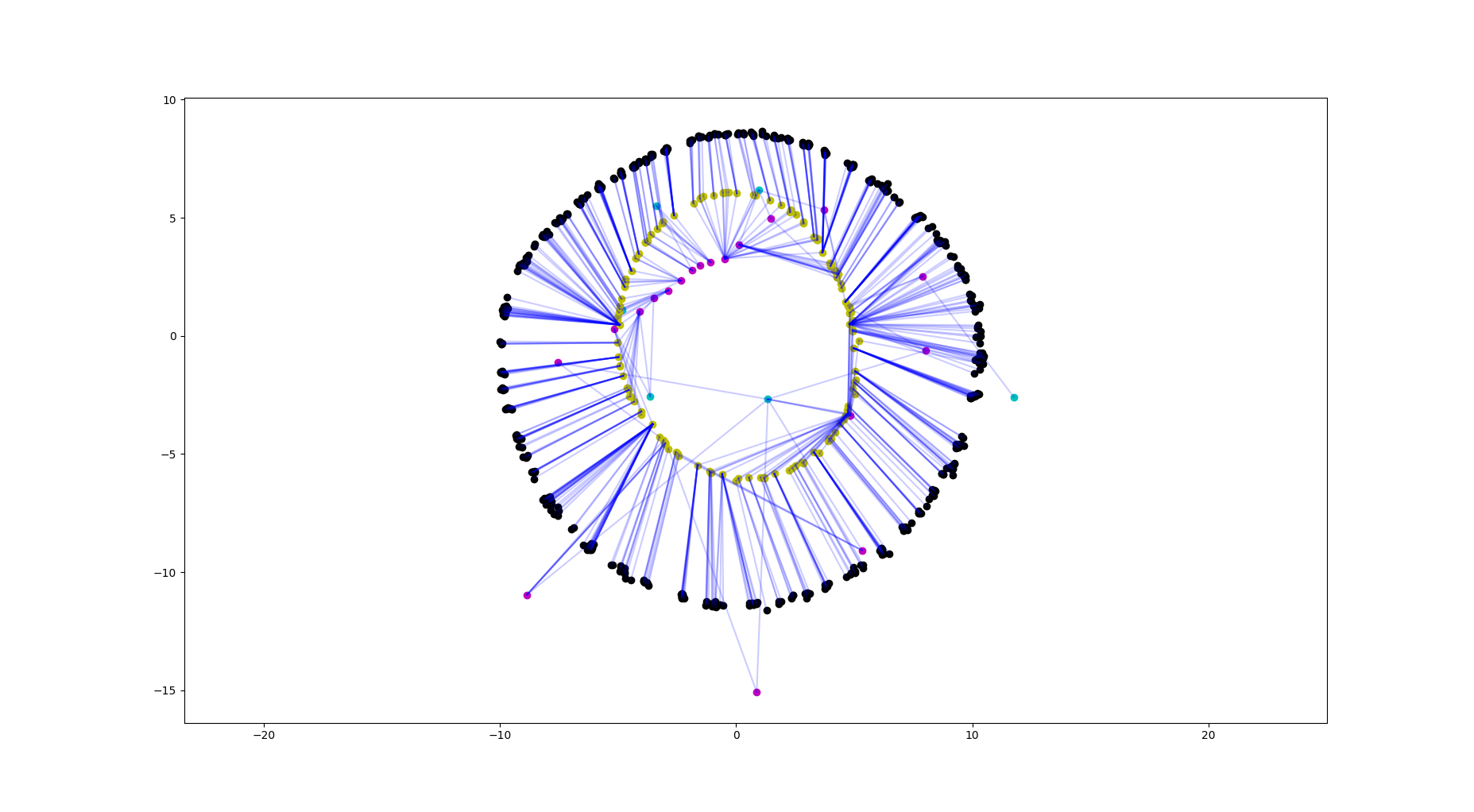}
    \caption{Visualization of inverted cosine embeddings (labels) in 2 dimensions. The cyan nodes represent \texttt{family}, the magenta nodes represent \texttt{sub-family}, the yellow nodes \texttt{genus} and black nodes \texttt{genus+species}. These closely resemble the euclidean cones.}
    \label{fig:inverted_cosine_emb}
\end{figure}

One can notice that the inverted cosine embeddings from \cref{fig:inverted_cosine_emb} (does not show the images) closely resemble the Euclidean cones of the label-embeddings from \cref{fig:ec_2d_labels_images}

\chapter{Conclusions}
\label{ch:conclusion}

\section{Future work}
In this section, we discuss possible future directions. We split them based on the domain these directions pertain to.
\begin{itemize}
    \item \textbf{Optimization.} During the experiments we noticed that using the Adam optimizer with an approximate update step for hyperbolic parameters worked better than a mathematically more accurate Riemannian SGD optimizer. From the point-of-view of optimizer, it is an interesting trade-off between strong but approximate and weak but accurate optimizers. We believe that this was observed because Adam is better-suited for optimizing such problems with highly non-convex landscape as the one we try to solve with embeddings-based models.
    
    In the results section we show that joint-embeddings when trained with labels-only embedding perform best. But randomly initialized labels also converge and have a decent performance to show for on the image classification task. In their experiments \cite{ganea2018entailment_cones} observed high-performing models when initialized with Poincar\'e embeddings from \cite{nickel2017poincare} as they conjecture that without such initialization the model does not perform as expected due the problem being difficult to optimize.
    
    
   
    \item \textbf{Datasets.} We show different methods that can be used to inject label-hierarchy knowledge to an arbitrary classifier. A direction would be to extend the proposed methods and the use of embeddings models for image classification to a variety of different datasets.
    
    \item \textbf{Label accuracy vs. label specificity trade-off.} On top of the proposed embedding and classification methods, one can think of adding an additional module/method to select a sweet-spot to trade-off between accuracy of the predicted labels and the specificity of the predicted labels such as the work proposed in \cite{deng2012hedging}. If the classifier is less confident about a prediction it can predict a more abstract label that is less informative but still correct.
    
    \item \textbf{Model complexity.} For embeddings-based models, one could use a more sophisticated model to map from \textit{fc7} image features to the embedding space for images. Obviously, this comes with possible issues like over-fitting and difficulty to optimize parameters living in non-euclidean space. The $W$ matrix used to map in our experiments lives in the Euclidean space but very recent work on hyperbolic neural networks \cite{ganea2018hyperbolicNN} could be used as a pointer to model feed-forward networks that replace the matrix $W$. This $W$ would then live as well as be optimized in hyperbolic space.
    
    \item \textbf{Applications.} We empirically show that a classifier that exploits its label-hierarchy outperforms a model that is hierarchy-agnostic. This provides a good base to improve existing image classification model to use this untapped information source.
    
    In addition to this, the learnt joint-embeddings can be used for downstream tasks such as image captioning, scene understanding and scene graph generation \cite{li2017scene, gu2019scene, xu2017scene} and visual question-answering (VQA) \cite{antol2015vqa}. Models that tackle these tasks lie at the intersection of images and natural language concepts. Our work moves in the direction to bridge the two different fields and treat them in a joint manner. Generally, for such tasks, the classifiers/CNN-backbones are used for visual feature extraction and object proposals while the semantics are obtained from a separate module (such as an LSTM \cite{li2017scene} or RNN \cite{gu2019scene, xu2017scene}) that models natural language and semantics. Because our proposed methods are aware of both visual similarity and semantic similarity (via the label-hierarchy information) this could improve performance by virtue of modeling visual features and semantics jointly.
\end{itemize}

\section{Summary}
\begin{itemize}
    \item We propose 4 different methods to pass label-hierarchy information to a classifier. The Marginalization model, Masked L-classifiers and the Hierarchical Softmax all perform better with a higher m-F1 score as compared to the methods. These models include information not exclusively about the number of levels in the hierarchy (like the per-level classifier) but also provide additional information information to the model (in different forms) about how labels are related to each other. Information about connections between labels is made available by (1) performing marginalization over child probabilities to obtain a probability distribution for upper levels in the hierarchy, (2) masking infeasible labels based on upper-level predictions to narrow down possible labels, and, (3) predict conditional distributions over the hierarchy and compute joint probabilities using the chain rule for probabilities.
    
    \item Order-preserving embeddings which have shown great promise for natural language related tasks are applied in the context of computer vision in this work. We break away from traditional softmax based classifiers and propose to use embeddings to perform image classification. These embeddings-based models outperform the hierarchy-agnostic classifier which has even been explicitly trained to perform multi-label image classification.
    
    This furthers the case that irrespective of the model paradigm (a classifier or an embedding) label-hierarchy information boosts performance on the image classification task. This shows promise for other tasks that encompass computer vision alone or computer vision in conjunction with natural language, where exploiting label-hierarchy could benefit the models.
    
    \item During the experiments with embeddings we realized that optimization is a relatively tricky process and sometimes methods are delicate and sensitive to the choice of initialization,  optimizers and hyper-parameters.
    
    \item We provide extensive experiments, empirical analysis, visualizations for the proposed methods and observe that the proposed label-hierarchy injection methods as well as the order-preserving joint-embeddings outperform the hierarchy-agnostic image classifier baseline on the introduced ETHEC dataset \cite{dhall_20.500.11850/365379} with 47,978 images and 723 labels spread across 4 hierarchical levels. The implementation for all the methods will be made publicly available.
\end{itemize}


\backmatter

\bibliographystyle{plain}
\bibliography{refs}


\end{document}